%% file: 0_MAIN.tex
\pdfoutput=1

\documentclass[11pt]{article}

\usepackage{acl}

\usepackage{times}
\usepackage{latexsym}
\usepackage{adjustbox}
\usepackage{amsmath}
\usepackage{amssymb}
\usepackage{graphicx}
\usepackage{tabularx}
\usepackage{algorithm}
\usepackage{multirow}
\usepackage{tabularx}
\usepackage{color, colortbl}
\usepackage{caption}
\usepackage{booktabs}
\usepackage{subcaption}
\usepackage{mwe}
\usepackage{soul}
\usepackage{url}
\usepackage{amsfonts}
\usepackage{bm}
\usepackage{euflag}
\usepackage[T1]{fontenc}
\newcommand\rurl[1]{%
  \href{http://#1}{\nolinkurl{#1}}%
}
\usepackage[utf8]{inputenc}
\definecolor{LightGray}{gray}{0.96}
\definecolor{LightCyan}{rgb}{0.92,0.968,0.968}
\usepackage{microtype}
\usepackage{inconsolata}
\usepackage[noend]{algpseudocode}
\newcommand{\rparagraph}[1]{\vspace{1.6mm}\noindent\textbf{#1.}}
\definecolor{Gray}{gray}{0.92}

\usepackage{todonotes}
\makeatletter
\newcommand*\iftodonotes{\if@todonotes@disabled\expandafter\@secondoftwo\else\expandafter\@firstoftwo\fi}
\makeatother

\newcommand{\argmin}{\operatornamewithlimits{argmin}}
\newcommand{\argmax}{\operatornamewithlimits{argmax}}

\title{Improving Word Translation via Two-Stage Contrastive Learning}

\author{Yaoyiran Li, Fangyu Liu, Nigel Collier, Anna Korhonen, \textnormal{and} Ivan Vuli\'{c} \\
  Language Technology Lab, TAL, University of Cambridge \\
  \texttt{\{yl711,fl399,nhc30,alk23,iv250\}@cam.ac.uk} \\} 

\begin{document}
\maketitle
\begin{abstract}
Word translation or bilingual lexicon induction (BLI) is a key cross-lingual task, aiming to bridge the lexical gap between different languages. In this work, we propose a robust and effective two-stage contrastive learning framework for the BLI task. At Stage C1, we propose to refine standard cross-lingual linear maps between static word embeddings (WEs) via a contrastive learning objective; we also show how to integrate it into the self-learning procedure for even more refined cross-lingual maps. In Stage C2, we conduct BLI-oriented contrastive fine-tuning of mBERT, unlocking its word translation capability. We also show that static WEs induced from the `C2-tuned' mBERT complement static WEs from Stage C1. Comprehensive experiments on standard BLI datasets for diverse languages and different experimental setups demonstrate substantial gains achieved by our framework. While the BLI method from Stage C1 already yields substantial gains over all state-of-the-art BLI methods in our comparison, even stronger improvements are met with the full two-stage framework: e.g., we report gains for $112/112$ BLI setups, spanning $28$ language pairs.
\end{abstract}

\section{Introduction and Motivation}
\label{s:introduction}
\input{1_introduction}
 
\section{Methodology}
\label{s:methodology}
\input{2_methodology}

\section{Experimental Setup}
\label{s:experimental}
\input{3_experimental}

\section{Results and Discussion}
\label{s:results}
\input{4_results}
 \section{Related Work}
\label{s:related_work}
\input{5_related_work}

\section{Conclusion}
\label{s:conclusion}
\input{6_conclusion}

\section*{Acknowledgements}
\label{s:acknowledgements}
\input{7_acknowledgements}

\section*{Ethics Statement}
\label{s:ethics}
\input{8_ethics}

\bibliography{anthology,custom}
\bibliographystyle{acl_natbib}

\clearpage
\input{x_appendix}

\end{document}

%% file: 1_introduction.tex
Bilingual lexicon induction (BLI) or word translation is one of the seminal and long-standing tasks in multilingual NLP \cite[\textit{inter alia}]{Rapp:1995acl,Gaussier:2004acl,Heyman:2017eacl,shi2021bilingual}. Its main goal is learning translation correspondences across languages, with applications of BLI ranging from language learning and acquisition \cite{Yuan:2020emnlp,Akyurek:2021acl} to machine translation \cite{Qi:2018naacl,Duan:2020acl,Chronopolou:2021naacl} and the development of language technology in low-resource languages and domains \cite{Irvine:2017cl,Heyman:2018bmc}. A large body of recent BLI work has focused on the so-called \textit{mapping-based} methods \cite{Mikolov:2013arxiv,artetxe2018robust,Ruder:2019survey}.\footnote{They are also referred to as \textit{projection-based} or \textit{alignment-based} methods \cite{glavas-etal-2019-properly,Ruder:2019survey}.} Such methods are particularly suitable for low-resource languages and weakly supervised learning setups: they support BLI with only as much as few thousand word
translation pairs (e.g., $1k$ or at most $5k$) as the only bilingual supervision \cite{Ruder:2019survey}.\footnote{In the extreme, \textit{fully unsupervised} mapping-based BLI methods can leverage monolingual data only without any bilingual supervision \cite[\textit{inter alia}]{conneau2017word,artetxe2018robust,hoshen2018nonadversarial,Mohiuddin2019revisiting,Ren:2020acl}. However, comparative empirical analyses \cite{vulic-etal-2019-really} show that, with all other components equal, using seed sets of only $500$-$1$$,$$000$ translation pairs, always outperforms fully unsupervised BLI methods. Therefore, in this work we focus on this more pragmatic (weakly) supervised BLI setup \cite{Artetxe:2020acl}; we assume the existence of at least $1$$,$$000$ seed translations per each language pair.}

\begin{figure}[t!]
    \centering
    \includegraphics[width=0.986\linewidth]{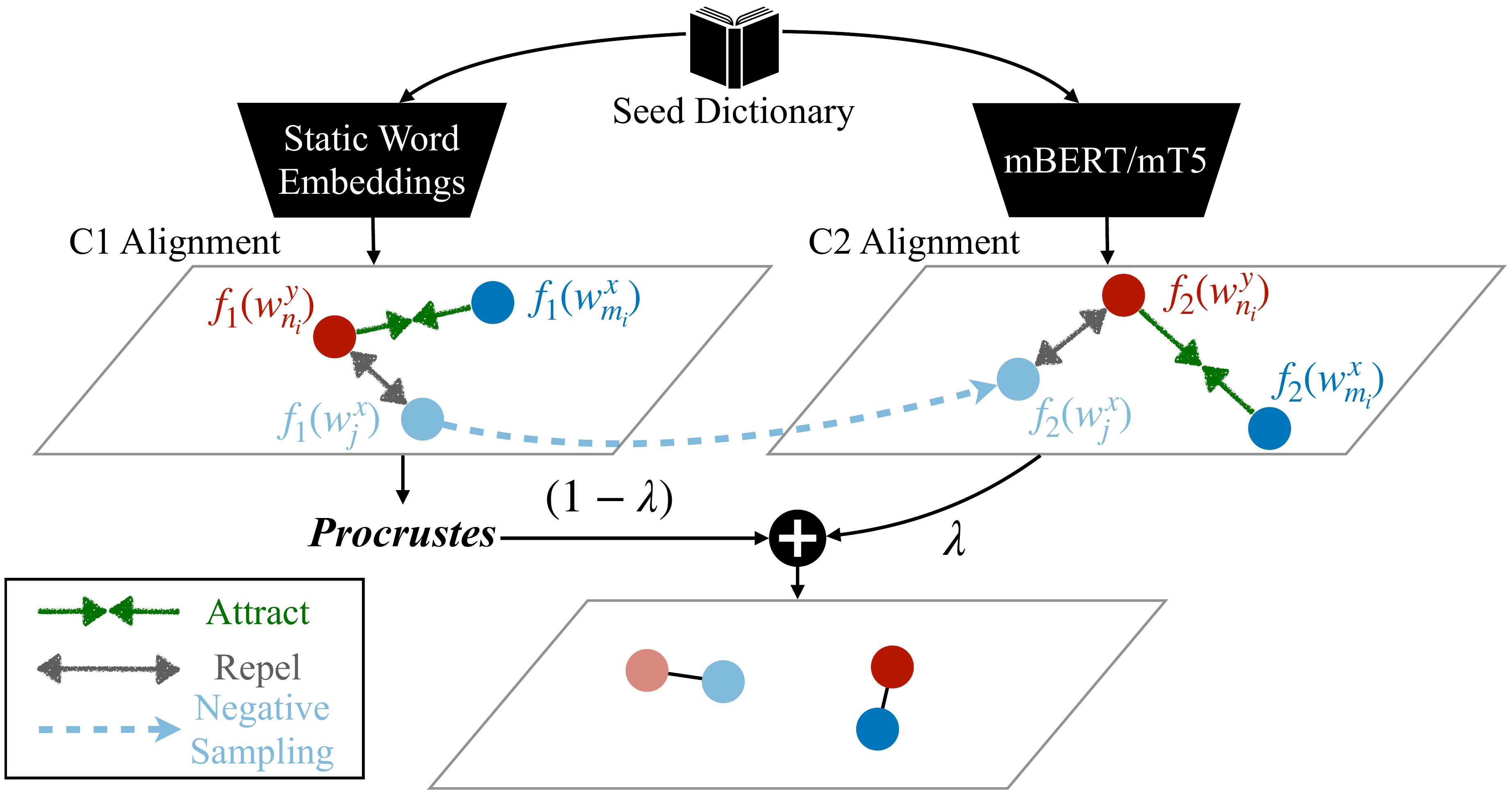}
    \caption{An illustration of the proposed two-stage BLI approach (see \S\ref{s:methodology}). It combines contrastive tuning on both static WEs (C1) and pretrained multilingual LMs (C2), where the static WEs are leveraged for selecting negative examples in contrastive tuning of the LM. The output of C1 and C2 is combined for the final BLI task.}
    \label{fig:front_pic}
    \vspace{-5mm}
\end{figure}

Unlike for many other tasks in multilingual NLP \cite{Doddapaneni:2021arxiv,Chau:2021ws,Ansell:2021arxiv}, state-of-the-art (SotA) BLI results are still achieved via static word embeddings (WEs) \cite{vulic-etal-2020-probing,liu2021fast}. A typical \textit{modus operandi} of mapping-based approaches is to first train monolingual WEs independently on monolingual corpora and then map them to a shared cross-lingual space via linear \cite{Mikolov:2013arxiv,glavas-etal-2019-properly} or non-linear mapping functions \cite{mohiuddin-etal-2020-lnmap}. In order to achieve even better results, many BLI methods also apply a self-learning loop where training dictionaries are iteratively (and gradually) refined, and improved mappings are then learned in each iteration \cite{artetxe2018robust,Karan:2020acl}. However, there is still ample room for improvement, especially for lower-resource languages and dissimilar language pairs \cite{vulic-etal-2019-really,Nasution:2021acm}.

On the other hand, another line of recent research has demonstrated that a wealth of lexical semantic information is encoded in large multilingual pretrained language models (LMs) such as mBERT \cite{devlin2018bert}, but \textbf{1)} it is not straightforward to transform the LMs into multilingual lexical encoders \cite{liu2021fast}, \textbf{2)} extract word-level information from them \cite{vulic-etal-2020-probing,vulic2021lexfit}, and \textbf{3)} word representations extracted from these LMs still cannot surpass static WEs in the BLI task \cite{vulic-etal-2020-probing,zhang-etal-2021-combining}. Motivated by these insights, in this work we investigate the following research questions:

\vspace{0.7mm}
\noindent \textbf{(RQ1)} Can we further improve (weakly supervised) mapping-based BLI methods based on static WEs?

\vspace{0.5mm}
\noindent \textbf{(RQ2)} How can we extract more useful cross-lingual word representations from pretrained multilingual LMs such as mBERT or mT5?

\vspace{0.5mm}
\noindent \textbf{(RQ3)} Is it possible to boost BLI by combining cross-lingual representations based on static WEs and the ones extracted from multilingual LMs?
\vspace{0.7mm}

Inspired by the wide success of contrastive learning techniques in \textit{sentence-level} representation learning \cite{Reimers:2019emnlp,carlsson2021semantic,gao2021simcse}, we propose a \textit{two-stage contrastive learning framework for effective word translation} in (weakly) supervised setups; it leverages and combines multilingual knowledge from static WEs and pretrained multilingual LMs. \textbf{Stage C1} operates solely on static WEs: in short, it is a mapping-based approach with self-learning, where in each step we additionally fine-tune linear maps with contrastive learning that operates on gradually refined positive examples (i.e., true translation pairs), and hard negative samples. \textbf{Stage C2} fine-tunes a pretrained multilingual LM (e.g., mBERT), again with a contrastive learning objective, using positive examples as well as negative examples extracted from the output of C1. Finally, we extract word representations from the multilingual LM fine-tuned in Stage C2, and combine them with static cross-lingual WEs from Stage C1; the combined representations are then used for BLI.


We run a comprehensive set of BLI experiments on the standard BLI benchmark \citep{glavas-etal-2019-properly}, comprising 8 diverse languages, in several setups. Our results indicate large gains over state-of-the-art BLI models: e.g., $\approx$+$8$ Precision@1 points on average, +$10$ points for many language pairs, gains for $107/112$ BLI setups already after Stage C1 (cf. RQ1), and for all $112/112$ BLI setups after Stage C2 (cf. RQ2 and RQ3). Moreover, our findings also extend to BLI for lower-resource languages from another BLI benchmark \cite{vulic-etal-2019-really}. Finally, as hinted in recent work \cite{zhang-etal-2021-combining}, our findings validate that multilingual lexical knowledge in LMs, when exposed and extracted as in our contrastive learning framework, can complement the knowledge in static cross-lingual WEs (RQ3), and benefit BLI. We release the code and share the data at \url{https://github.com/cambridgeltl/ContrastiveBLI}.




%% file: 2_methodology.tex
\noindent \textbf{Preliminaries and Task Formulation.}
In BLI, we assume two vocabularies $\mathcal{X}$$=$$\{w_{1}^{x},\ldots,w_{|\mathcal{X}|}^{x}\}$ and $\mathcal{Y}$$=$$\{w_{1}^{y},\ldots,w_{|\mathcal{Y}|}^y\}$ associated with two respective languages $L_x$ and $L_y$. We also assume that each vocabulary word is assigned its (static) type-level word embedding (WE); that is, the respective WE matrices for each vocabulary are $\bm{X}$$\in$$\mathbb{R}^{|\mathcal{X}|\times d}$, $\bm{Y}$$\in$$\mathbb{R}^{|\mathcal{Y}|\times d}$. Each WE is a $d$-dim row vector, with typical values $d$$=$$300$ for static WEs (e.g., fastText) \cite{bojanowski2017enriching}, and $d$$=$$768$ for mBERT.\footnote{We also tried XLM ($d$$=$$1$$,$$280$) and mT5$_{small}$ ($d$$=$$512$); mBERT is the best-performing pretrained LM in our preliminary investigation.} We also assume a set of \textit{seed} translation pairs $\mathcal{D}_0$$=$$\{(w_{m_{1}}^{x},w_{n_{1}}^{y}),...,(w_{m_{|\mathcal{D}_0|}}^{x},w_{n_{|\mathcal{D}_0|}}^{y})\}$ for training \cite{Mikolov:2013arxiv,glavas-etal-2019-properly}, where $1\leq m_{i}\leq |\mathcal{X}|, 1\leq n_{i}\leq |\mathcal{Y}|$. Typical values for the seed dictionary size $|\mathcal{D}_0|$ are $5k$ pairs and $1k$ pairs \cite{vulic-etal-2019-really}, often referred to as \textit{supervised} ($5k$) and \textit{semi-supervised} or \textit{weakly supervised} settings ($1k$) \cite{artetxe2018robust}. Given another \textit{test} lexicon $\mathcal{D}_T$$=$$\{(w_{t_{1}}^{x},w_{g_{1}}^{y}),...,(w_{t_{|\mathcal{D}_T|}}^{x},w_{g_{|\mathcal{D}_T|}}^{y})\}$, where $1\leq t_{i}\leq |\mathcal{X}|, 1\leq g_{i}\leq |\mathcal{Y}|$ and $\mathcal{D}_0 \cap \mathcal{D}_T = \emptyset$, for each $L_x$ test word $w_{t_{i}}^{x}$ in $\mathcal{D}_T$ the goal is to retrieve its correct translation from $L_y$'s vocabulary $\mathcal{Y}$, and evaluate the retrieved $L_y$ word against the gold $L_y$ translation $w_{g_{i}}^{y}$ from the pair $(w_{t_{i}}^{x},w_{g_{i}}^{y})\in\mathcal{D}_T$.\footnote{If $w_{t_{i}}^{x}$ has two or more translations in $\mathcal{D}_T$ (i.e, the same $L_x$ word appears in more than one word pairs in $\mathcal{D}_T$ with different gold $L_y$ words), a translation is then considered correct if it is any one of the ground-truth translations in $\mathcal{D}_T$.}

\rparagraph{Method in a Nutshell}
We propose a novel two-stage contrastive learning (CL) method, with both stages C1 and C2 realised via contrastive learning objectives (see Figure~\ref{fig:front_pic}). Stage C1 (\S\ref{s:c1}) operates solely on static WEs, and can be seen as a contrastive extension of mapping-based BLI approaches with static WEs. In practice, we blend contrastive learning with the standard SotA mapping-based framework with self-learning: VecMap \cite{artetxe2018robust}, with some modifications. Stage C1 operates solely on static WEs in exactly the same BLI setup as prior work, and thus it can be evaluated independently. In Stage C2 (\S\ref{s:c2}), we propose to leverage pretrained multilingual LMs for BLI: we contrastively fine-tune them for BLI and extract static `decontextualised' WEs from the tuned LMs. These LM-based WEs can be combined with WEs obtained in Stage C1 (\S\ref{s:combined}).

\subsection{Stage C1}
\label{s:c1}
Stage C1 is based on the VecMap framework \cite{artetxe2018robust} which features \textbf{1)} \textit{dual linear mapping}, where two separate linear transformation matrices map respective source and target WEs to a shared cross-lingual space; and \textbf{2)} a \textit{self-learning} procedure that, in each iteration $i$ refines the training dictionary and iteratively improves the mapping. We extend and refine VecMap's self-learning for supervised and semi-supervised settings via CL. 


\rparagraph{Initial Advanced Mapping} 
After $\ell_2$-normalising word embeddings,\footnote{Unlike VecMap, we do not mean-center WEs as this yielded slightly better results in our preliminary experiments.} the two mapping matrices, denoted as $\bm{W}_{x}$ for the source language $L_x$ and $\bm{W}_{y}$ for $L_y$, are computed via the Advanced Mapping (AM) procedure based on the training dictionary, as fully described in Appendix~\ref{appendix:am}; while VecMap leverages whitening, orthogonal mapping, re-weighting and de-whitening operations to derive mapped WEs, we compute $\bm{W}_{x}$ and $\bm{W}_{y}$ such that a one-off matrix multiplication produces the same result (see Appendix~\ref{appendix:am} for the details). 


\rparagraph{Contrastive Fine-Tuning}
At each iteration $i$, after the initial AM step, the two mapping matrices $\bm{W}_{x}$ and $\bm{W}_{y}$ are then further contrastively fine-tuned via the InfoNCE loss \cite{oord2018representation}, a standard and robust choice of a loss function in CL research \cite{Musgrave:2020eccv,liu2021mirrorwic,liu2021fast}. We denote the dictionary for CL in the current iteration as $\mathcal{D}_{\text{CL}}$. The core idea of our approach is to `attract' aligned WEs of positive (true) translation pairs from the dictionary $\mathcal{D}_{\text{CL}}$ together, and `repel' the aligned WEs of \textit{hard negative pairs} (i.e., words which are semantically similar but do not constitute a word translation pair) from each other.  

These hard negative pairs are extracted as follows. Let us suppose that $(w^{x}_{m_{i}}, w^{y}_{n_{i}})$ is a training (positive) pair in the current dictionary $\mathcal{D}_{\text{CL}}$, with its constituent words associated with static WEs $\mathbf{x}_{m_{i}}, \mathbf{y}_{n_{i}}$$\in$$ \mathbb{R}^{1\times d}$. We then retrieve the nearest neighbours of $\mathbf{y}_{n_{i}}\bm{W}_{y}$ from $\bm{X}\bm{W}_{x}$ and derive $\bar{w}^{x}_{m_{i}}\subset \mathcal{X}$ ($w^{x}_{m_{i}}$ excluded), a set of hard negative samples of size $N_{\text{neg}}$. In a similar (symmetric) manner, we also derive the set of negatives $\bar{w}^{y}_{n_{i}}\subset \mathcal{Y}$ ($w^{y}_{n_{i}}$ excluded). Then, for the training pair, the corresponding negative pairs are the union of $\{w^{x}_{m_{i}}\}\times\bar{w}^{y}_{n_{i}}$ and $\bar{w}^{x}_{m_{i}}\times \{w^{y}_{n_{i}}\}$, where `$\times$' denotes the Cartesian product. We use $\mathcal{\bar{D}}$ to denote the collection of all hard negative pairs over all training pairs in the current iteration $i$. We then fine-tune $\bm{W}_{x}$ and $\bm{W}_{y}$ by optimising the following contrastive objective:

\vspace{-2mm}
{\footnotesize
\begin{align}
& s_{i,j} = \exp(\cos(\mathbf{x}_{i}\bm{W}_{x} \ , \  \mathbf{y}_{j}\bm{W}_{y})/\tau), \\
& p_{i} = \frac{s_{m_{i},n_{i}}}{\underset{w^{y}_{j} \in \{w^{y}_{n_{i}}\}\bigcup  \bar{w}^{y}_{n_{i}}}{\sum}s_{m_{i},j} + \underset{w^{x}_{j} \in \bar{w}^{x}_{m_{i}}}{\sum}s_{j,n_{i}}}, \\
& \underset{\bm{W}_{x},\bm{W}_{y}}{\min}  -\mathbb{E}_{(w^{x}_{m_{i}},w^{y}_{n_{i}})\in \mathcal{D}_{\text{CL}}} 
\log (p_{i}).
\end{align}}%
\noindent $\tau$ denotes a standard temperature parameter.
The objective, formulated here for a single positive example, spans all positive examples from the current dictionary, along with the respective sets of negative examples computed as described above.

\begin{algorithm}[!t]
\caption{{\footnotesize Stage C1: Self-Learning}}
{\footnotesize
\begin{algorithmic}[1]
\State \textbf{Require:} $\bm{X}$,$\bm{Y}$,$\mathcal{D}_0$,$\mathcal{D}_{\text{add}}$ $\gets$ $\emptyset$
\For{$i$ $\gets$ $1$ \textbf{to} $N_{\text{iter}}$}
    \State $\bm{W}_{x},\bm{W}_{y}$ $\gets$ Initial AM using $\mathcal{D}_{i-1}$;
    \State $\mathcal{D}_{\text{CL}}$ $\gets$ $\mathcal{D}_0$ (supervised) or $\mathcal{D}_{i-1}$ (semi-supervised);
    \For {$j$ $\gets$ $1$ \textbf{to} $N_{\text{CL}}$}
        \State Retrieve $\mathcal{\bar{D}}$ for the pairs from $\mathcal{D}_{\text{CL}}$;
        \State $\bm{W}_{x},\bm{W}_{y}$ $\gets$ Optimise Contrastive Loss;
    \EndFor
    \vspace{-0.3mm}
    \State Compute new $\mathcal{D}_{\text{add}}$;
    \State Update $\mathcal{D}_{i}$ $\gets$ $\mathcal{D}_{0} \cup \mathcal{D}_{\text{add}}$;
\EndFor
\vspace{-0.2mm}
\State \Return{$\bm{W}_{x},\bm{W}_{y}$};
\end{algorithmic}
}%
\label{alg:c1}
\vspace{-0.5mm}
\end{algorithm}

\rparagraph{Self-Learning}
The application of (a) initial mapping via AM and (b) contrastive fine-tuning can be repeated iteratively. Such self-learning loops typically yield more robust and better-performing BLI methods \cite{artetxe2018robust,vulic-etal-2019-really}.  At each iteration $i$, a set of automatically extracted high-confidence translation pairs $\mathcal{D}_{\text{add}}$ is added to the seed dictionary $\mathcal{D}_0$, and this dictionary $\mathcal{D}_i$= $\mathcal{D}_0 \cup \mathcal{D}_{\text{add}}$ is then used in the next iteration $i+1$. 

Our dictionary augmentation method slightly deviates from the one used by VecMap. We leverage the most frequent $N_{\text{freq}}$ source and target vocabulary words, and conduct forward and backward dictionary induction \cite{artetxe2018robust}. Unlike VecMap, we do not add stochasticity to the process, and simply select the top $N_{\text{aug}}$ high-confidence word pairs from forward (i.e., source-to-target) induction and another $N_{\text{aug}}$ pairs from the backward induction. In practice, we retrieve the $2$$\times$$N_{\text{aug}}$ pairs with the highest Cross-domain Similarity Local Scaling (CSLS) scores \cite{conneau2017word},\footnote{Further details on the CSLS similarity and its relationship to cosine similarity are available in Appendix~\ref{appendix:csls}.} remove duplicate pairs and those that contradict with the ground truth in $\mathcal{D}_0$, and then add the rest into $\mathcal{D}_{\text{add}}$. 

For the initial AM step, we always use the augmented dictionary $\mathcal{D}_0 \cup \mathcal{D}_{\text{add}}$; the same augmented dictionary is used as $\mathcal{D}_{\text{CL}}$ for contrastive fine-tuning in weakly supervised setups.\footnote{When $|\mathcal{D}_0|$$=$$5k$, we leverage only $\mathcal{D}_0$ for contrastive fine-tuning ($\mathcal{D}_{\text{CL}}$$=$$\mathcal{D}_0$), as $\mathcal{D}_{\text{add}}$ might introduce noise into $\mathcal{D}_{\text{CL}}$ and thus deteriorate the quality of learned maps.} We repeat the self-learning loop for $N_{\text{iter}}$ times: in each iteration, we optimise the contrastive loss $N_{\text{CL}}$ times; that is, we go $N_{\text{CL}}$ times over all the positive pairs from the training dictionary (at this iteration). $N_{\text{iter}}$ and $N_{\text{CL}}$ are tunable hyper-parameters. Self-learning in Stage C1 is summarised in Algorithm~\ref{alg:c1}.


\subsection{Stage C2}
\label{s:c2}

 Previous work tried to prompt off-the-shelf multilingual LMs for word translation knowledge via masked natural language templates \cite{gonen-etal-2020-greek}, averaging over their contextual encodings in a large corpus \cite{vulic-etal-2020-probing,zhang-etal-2021-combining}, or extracting type-level WEs from the LMs directly without context \cite{vulic2020multi,vulic2021lexfit}. However, even sophisticated templates and WE extraction strategies still lag behind fastText-based methods in BLI performance \cite{vulic2021lexfit}. 
 
 \rparagraph{(BLI-Oriented) Contrastive Fine-Tuning}
Here, we propose to fine-tune off-the-shelf multilingual LMs relying on the supervised BLI signal: the aim is to expose type-level word translation knowledge directly from the LM, without any external corpora. In practice, we first prepare a dictionary of positive examples for contrastive fine-tuning: (a) $\mathcal{D}_{\text{CL}}$$=$$\mathcal{D}_0$ when $|\mathcal{D}_0|$ spans $5k$ pairs, or (b) when $|\mathcal{D}_0|$$=$$1k$, we add the $N_{\text{aug}}$$=$$4k$ automatically extracted highest-confidence pairs from Stage C1 (based on their CSLS scores, not present in $\mathcal{D}_0$) to $\mathcal{D}_0$ (i.e., $\mathcal{D}_{\text{CL}}$ spans $1k$ + $4k$ word pairs). We then extract $N_{\text{neg}}$ hard negatives in the same way as in \S\ref{s:c1}, relying on the shared cross-lingual space derived as the output of Stage C1. Our hypothesis is that a difficult task of discerning between true translation pairs and highly similar non-translations as hard negatives, formulated within a contrastive learning objective, will enable mBERT to expose its word translation knowledge, and complement the knowledge already available after Stage C1.



Throughout this work, we assume the use of pretrained mBERT$_{\text{base}}$ model with $12$ Transformer layers and $768$-dim embeddings. Each raw word input $w$ is tokenised, via mBERT's dedicated tokeniser, into the following sequence: $[CLS] [sw_1]\ldots[sw_M][SEP]$, $M\geq1$, where $[sw_1]\ldots[sw_M]$ refers to the sequence of $M$ constituent subwords/WordPieces of $w$, and $[CLS]$ and $[SEP]$ are special tokens \cite{vulic-etal-2020-probing}.


The sequence is then passed through mBERT as the encoder, its encoding function denoted as $\textit{f}_{\theta}(\cdot)$: it extracts the representation of the [CLS] token in the last Transformer layer as the representation of the input word $w$. The full set of mBERT's parameters $\theta$ then gets contrastively fine-tuned in Stage C2, again relying on the InfoNCE CL loss:

\vspace{-1mm}
{\footnotesize
\begin{align}
& s'_{i,j} = \exp(\cos(\textit{f}_{\theta}(w^{x}_{i}),\textit{f}_{\theta}(w^{y}_{j}))/\tau), \\
& p'_{i} = \frac{s'_{m_{i},n_{i}}}{\underset{w^{y}_{j} \in \{w^{y}_{n_{i}}\}\bigcup \bar{w}^{y}_{n_{i}}}{\sum}s'_{m_{i},j} + \underset{w^{x}_{j} \in \bar{w}^{x}_{m_{i}}}{\sum}s'_{j,n_{i}}}, \\
& \underset{\theta}{\min}  -\mathbb{E}_{(w^{x}_{m_{i}},w^{y}_{n_{i}})\in \mathcal{D}_{\text{CL}}} \log (p'_{i}).
\end{align}}%
\noindent Type-level WE for each input word $w$ is then obtained simply as $f_{\theta'}(w)$, where $\theta'$ refers to the parameters of the `BLI-tuned' mBERT model.

\subsection{Combining the Output of C1 and C2}
\label{s:combined}
In order to combine the output WEs from Stage C1 and the mBERT-based WEs from Stage C2, we also need to map them into a `shared' space: in other words, for each word $w$, its C1 WE and its C2 WE can be seen as two different views of the same data point. We thus learn an additional linear orthogonal mapping from the C1-induced cross-lingual WE space into the C2-induced cross-lingual WE space. It transforms $\ell_2$-normed $300$-dim C1-induced cross-lingual WEs into $768$-dim cross-lingual WEs. Learning of the linear map $\bm{W}$$\in$$\mathbb{R}^{d_{1}\times d_{2}}$, where in our case $d_1$$=$$300$ and $d_2$$=$$768$, is formulated as a Generalised Procrustes problem \cite{schonemann1966generalized,viklands2006algorithms} operating on all (i.e., both $L_x$ and $L_y$) words from $\mathcal{D}_{\text{CL}}$.\footnote{Technical details of the learning procedure are described in Appendix~\ref{appendix:procrustes}. It is important to note that in this case we do not use word translation pairs $(w_{m_{i}}^{x},w_{n_{i}}^{y})$ directly to learn the mapping, but rather each word $w_{m_{i}}^{x}$ and $w_{n_{i}}^{y}$ is duplicated to create training pairs $(w_{m_{i}}^{x},w_{m_{i}}^{x})$ and $(w_{n_{i}}^{y},w_{n_{i}}^{y})$, where the left word/item in each pair is assigned its WE from C1, and the right word/item is assigned its WE after C2.} 

Unless noted otherwise, a final representation of an input word $w$ is then a linear combination of (a) its C1-based vector $\mathbf{v}_w$ mapped to a $768$-dim representation via $\bm{W}$, and (b) its $768$-dim encoding $f_{\theta'}(w)$ from BLI-tuned mBERT:

\vspace{-1.5mm}
{\footnotesize
\begin{align}
(1-\lambda)\frac{\mathbf{v}_w\bm{W}}{\left\|\mathbf{v}_w\bm{W}\right\|_{2}} + \lambda \frac{f_{\theta'}(w)}{\left\|f_{\theta'}(w)\right\|_{2}},
\label{formula:CL2Loss04}
\end{align}}%
where $\lambda$ is a tunable interpolation hyper-parameter.

%% file: 3_experimental.tex
\noindent \textbf{Monolingual WEs and BLI Setup.} We largely follow the standard BLI setup from prior work \cite[\textit{inter alia}]{artetxe2018robust,joulin-etal-2018-loss,glavas-etal-2019-properly,Karan:2020acl}. The main evaluation is based on the standard BLI dataset from \newcite{glavas-etal-2019-properly}: it comprises 28 language pairs with a good balance of typologically similar and distant languages (Croatian: \textsc{hr}, English: \textsc{en}, Finnish: \textsc{fi}, French: \textsc{fr}, German: \textsc{de}, Italian: \textsc{it}, Russian: \textsc{ru}, Turkish: \textsc{tr}). Again following prior work, we rely on monolingual fastText vectors trained on Wikipedia data for each language \cite{bojanowski2017enriching}, where vocabularies in each language are trimmed to the $200k$ most frequent words (i.e., $|\mathcal{X}|$$=$$200k$ and $|\mathcal{Y}|$$=$$200k$). The same fastText WEs are used for our Stage C1 and in all baseline BLI models. mBERT in Stage C2 operates over the same vocabularies spanning $200k$ word types in each language.

We use $1k$ translation pairs (semi-supervised BLI setup) or $5k$ pairs (supervised) as seed dictionary $\mathcal{D}_0$; test sets span $2k$ pairs \cite{glavas-etal-2019-properly}. With $56$ BLI directions in total,\footnote{For any two languages $L_i$ and $L_j$, we run experiments both for $L_i\rightarrow L_j$ and $L_j\rightarrow L_i$ directions.} this yields a total of $112$ BLI setups for each model in our comparison. The standard \textit{Precision@1} (P@1) BLI measure is reported, and we rely on CSLS ($k$$=$$10$) to score word similarity \cite{conneau2017word}.\footnote{The same trends in results are observed with Mean Reciprocal Rank (MRR) as another BLI evaluation measure \cite{glavas-etal-2019-properly}; we omit MRR scores for clarity. Moreover, similar relative trends, but with slightly lower absolute BLI scores, are observed when replacing CSLS with the simpler cosine similarity measure: the results are available in the Appendix.}

\rparagraph{Training Setup and Hyperparameters} Since standard BLI datasets typically lack a validation set \cite{Ruder:2019survey}, following prior work \cite{glavas-etal-2019-properly,Karan:2020acl} we conduct hyperparameter tuning on a \textit{single, randomly selected} language pair EN$\to$TR, and apply those hyperparameter values in all other BLI runs. 


In Stage C1, when $|\mathcal{D}_0|$$=$$5k$, the hyperparameter values are $N_{\text{iter}}$$=$$2$, $N_{\text{CL}}$$=$$200$, $N_{\text{neg}}$$=$$150$, $N_{\text{freq}}$$=$$60k$, $N_{\text{aug}}$$=$$10k$. SGD optimiser is used, with a learning rate of $1.5$ and $\gamma$$=$$0.99$. When $|\mathcal{D}_0|$$=$$1k$, the values are $N_{\text{iter}}$$=$$3$, $N_{\text{CL}}$$=$$50$, $N_{\text{neg}}$$=$$60$, $N_{\text{freq}}$$=$$20k$, and $N_{\text{aug}}$$=$$6k$; SGD with a learning rate of $2.0$, $\gamma$$=$$1.0$. $\tau$$=$$1.0$ and dropout is $0$ in both cases, and the batch size for contrastive learning is always equal to the size of the current dictionary $|\mathcal{D}_{\text{CL}}|$ (i.e., $|\mathcal{D}_0|$ ($5k$ case), or $|\mathcal{D}_0 \cup \mathcal{D}_{\text{add}}|$ which varies over iterations ($1k$ case); see \S\ref{s:c1}). In Stage C2, $N_{\text{neg}}$$=$$28$ and the maximum sequence length is $6$. We use AdamW \cite{Loschilov:2018iclr} with learning rate of $2e-5$ and weight decay of $0.01$. We fine-tune mBERT for $5$ epochs, with a batch size of $100$; dropout rate is $0.1$ and $\tau$$=$$0.1$. Unless noted otherwise, $\lambda$ is fixed to $0.2$.

\rparagraph{Baseline Models}
Our BLI method is evaluated against four strong SotA BLI models from recent literature, all of them with publicly available implementations. Here, we provide brief summaries:\footnote{For further technical details and descriptions of each BLI model, we refer to their respective publications. We used publicly available implementations of all the baseline models.}

\vspace{0.5mm}
\noindent \textbf{RCSLS} \cite{joulin-etal-2018-loss} optimises a relaxed CSLS loss, learns a non-orthogonal mapping, and has been established as a strong BLI model in empirical comparative analyses as its objective function is directly `BLI-oriented' \cite{glavas-etal-2019-properly}.


\vspace{0.5mm}
\noindent \textbf{VecMap}'s core components \cite{artetxe2018robust} have been outlined in \S\ref{s:c1}.

\vspace{0.5mm}
\noindent \textbf{LNMap} \cite{mohiuddin-etal-2020-lnmap} non-linearly maps the original static WEs into two latent semantic spaces learned via non-linear autoencoders,\footnote{This step is directed towards mitigating non-isomorphism \cite{sogaard2018limitations,Dubossarsky:2020emnlp} between the original WE spaces, which should facilitate their alignment.} and then learns two other non-linear mappings between the latent autoencoder-based spaces. 


\vspace{0.5mm}
\noindent \textbf{FIPP} \cite{sachidananda2021filtered}, in brief, first finds common (i.e., isomorphic) geometric structures in monolingual WE spaces of both languages, and then aligns the Gram matrices of the WEs found in those common structures.

\vspace{0.8mm}
For all baselines, we have verified that the hyperparameter values suggested in their respective repositories yield (near-)optimal BLI performance. Unless noted otherwise, we run VecMap, LNMap, and FIPP with their own self-learning procedures.\footnote{RCSLS is packaged without self-learning; extending it to support self-learning is non-trivial and goes beyond the scope of this work.}

\rparagraph{Model Variants} We denote the full two-stage BLI model as \textbf{C2 (Mod)}, where \textbf{Mod} refers to the actual model/method used to derive the shared cross-lingual space used by Stage C2. For instance, \textbf{C2 (C1)} refers to the model variant which relies on our Stage C1, while \textbf{C2 (RCSLS)} relies on RCSLS as the base method. We also evaluate BLI performance of our Stage \textbf{C1} BLI method alone.

\rparagraph{Multilingual LMs} We adopt mBERT as the default pretrained multilingual LM in Stage C2. Our supplementary experiments also cover the $1280$-dim XLM model\footnote{We pick the XLM large model pretrained on $100$ languages with masked language modeling (MLM) objective.} \cite{lample2019cross} and $512$-dim mT5$_{small}$ \cite{xue-etal-2021-mt5}.\footnote{We also tested XLM-R$_{\text{base}}$, but in our preliminary experiments it shows inferior BLI performance.} For clarity, we use \textbf{C2 [LM]} to denote \textbf{C2 (C1)} obtained from different LMs; when \textbf{[LM]} is not specified, mBERT is used. We adopt a smaller batch size of $50$ for \textbf{C2 [XLM]} considering the limit of GPU memory, and train \textbf{C2 [mT5]} with a larger learning rate of $6e-4$ for $6$ epochs, since we found it much harder to train than \textbf{C2 [mBERT]}.

%% file: 4_results.tex
The main results are provided in Table~\ref{table:main}, while the full results per each individual language pair, and also with cosine similarity as the word retrieval function, are provided in Appendix~\ref{appendix:full}. The main findings are discussed in what follows.


\begin{table}[!t]
\begin{center}
\resizebox{1.0\linewidth}{!}{%
\def\arraystretch{0.9}
\begin{tabular}{l cccccc}
\toprule 

\rowcolor{Gray}
\multicolumn{1}{c}{[5k] \bf Pairs}  &\multicolumn{1}{c}{\bf RCSLS$^{+}$}  &\multicolumn{1}{c}{\bf VecMap$^{x}$} &\multicolumn{1}{c}{\bf LNMap} &\multicolumn{1}{c}{\bf FIPP}
&\multicolumn{1}{c}{\bf C1}
&\multicolumn{1}{c}{\bf C2 (C1)}
\\ \cmidrule(lr){2-5} \cmidrule(lr){6-7}

\multicolumn{1}{c}{DE$\to$$*$}  &\multicolumn{1}{c}{43.77}  &\multicolumn{1}{c}{40.49}  &\multicolumn{1}{c}{40.35}  &\multicolumn{1}{c}{40.95}  &\multicolumn{1}{c}{\underline{46.14}}  &\multicolumn{1}{c}{\textbf{48.86}}\\

\multicolumn{1}{c}{$*$$\to$DE}  &\multicolumn{1}{c}{44.74}  &\multicolumn{1}{c}{42.18}  &\multicolumn{1}{c}{39.55}  &\multicolumn{1}{c}{41.66}  &\multicolumn{1}{c}{\underline{46.39}}  &\multicolumn{1}{c}{\textbf{50.12}}\\

\multicolumn{1}{c}{EN$\to$$*$}  &\multicolumn{1}{c}{50.94}  &\multicolumn{1}{c}{45.43}  &\multicolumn{1}{c}{44.74}  &\multicolumn{1}{c}{45.76}  &\multicolumn{1}{c}{\underline{51.31}}  &\multicolumn{1}{c}{\textbf{54.31}}\\

\multicolumn{1}{c}{$*$$\to$EN}  &\multicolumn{1}{c}{49.17}  &\multicolumn{1}{c}{50.19}  &\multicolumn{1}{c}{44.32}  &\multicolumn{1}{c}{47.96}  &\multicolumn{1}{c}{\underline{52.61}}  &\multicolumn{1}{c}{\textbf{55.47}}\\

\multicolumn{1}{c}{FI$\to$$*$}  &\multicolumn{1}{c}{35.11}  &\multicolumn{1}{c}{36.29}  &\multicolumn{1}{c}{33.18}  &\multicolumn{1}{c}{34.83}  &\multicolumn{1}{c}{\underline{39.80}}  &\multicolumn{1}{c}{\textbf{43.44}}\\

\multicolumn{1}{c}{$*$$\to$FI}  &\multicolumn{1}{c}{33.49}  &\multicolumn{1}{c}{33.40}  &\multicolumn{1}{c}{34.15}  &\multicolumn{1}{c}{33.00}  &\multicolumn{1}{c}{\underline{38.82}}  &\multicolumn{1}{c}{\textbf{41.97}}\\

\multicolumn{1}{c}{FR$\to$$*$}  &\multicolumn{1}{c}{47.02}  &\multicolumn{1}{c}{44.67}  &\multicolumn{1}{c}{42.80}  &\multicolumn{1}{c}{44.03}  &\multicolumn{1}{c}{\underline{49.12}}  &\multicolumn{1}{c}{\textbf{51.91}}\\

\multicolumn{1}{c}{$*$$\to$FR}  &\multicolumn{1}{c}{49.42}  &\multicolumn{1}{c}{48.86}  &\multicolumn{1}{c}{46.25}  &\multicolumn{1}{c}{48.08}  &\multicolumn{1}{c}{\underline{51.84}}  &\multicolumn{1}{c}{\textbf{54.53}}\\

\multicolumn{1}{c}{HR$\to$$*$}  &\multicolumn{1}{c}{34.06}  &\multicolumn{1}{c}{36.26}  &\multicolumn{1}{c}{33.41}  &\multicolumn{1}{c}{33.52}  &\multicolumn{1}{c}{\underline{40.22}}  &\multicolumn{1}{c}{\textbf{45.53}}\\

\multicolumn{1}{c}{$*$$\to$HR}  &\multicolumn{1}{c}{32.80}  &\multicolumn{1}{c}{32.96}  &\multicolumn{1}{c}{31.34}  &\multicolumn{1}{c}{31.52}  &\multicolumn{1}{c}{\underline{37.82}}  &\multicolumn{1}{c}{\textbf{42.65}}\\

\multicolumn{1}{c}{IT$\to$$*$}  &\multicolumn{1}{c}{46.59}  &\multicolumn{1}{c}{44.77}  &\multicolumn{1}{c}{43.23}  &\multicolumn{1}{c}{44.11}  &\multicolumn{1}{c}{\underline{48.92}}  &\multicolumn{1}{c}{\textbf{51.91}}\\

\multicolumn{1}{c}{$*$$\to$IT}  &\multicolumn{1}{c}{48.41}  &\multicolumn{1}{c}{47.85}  &\multicolumn{1}{c}{45.53}  &\multicolumn{1}{c}{46.64}  &\multicolumn{1}{c}{\underline{50.99}}  &\multicolumn{1}{c}{\textbf{53.85}}\\

\multicolumn{1}{c}{RU$\to$$*$}  &\multicolumn{1}{c}{40.99}  &\multicolumn{1}{c}{41.01}  &\multicolumn{1}{c}{37.94}  &\multicolumn{1}{c}{39.72}  &\multicolumn{1}{c}{\underline{44.17}}  &\multicolumn{1}{c}{\textbf{47.24}}\\

\multicolumn{1}{c}{$*$$\to$RU}  &\multicolumn{1}{c}{40.10}  &\multicolumn{1}{c}{35.62}  &\multicolumn{1}{c}{35.66}  &\multicolumn{1}{c}{36.03}  &\multicolumn{1}{c}{\underline{42.15}}  &\multicolumn{1}{c}{\textbf{45.20}}\\

\multicolumn{1}{c}{TR$\to$$*$}  &\multicolumn{1}{c}{31.29}  &\multicolumn{1}{c}{31.54}  &\multicolumn{1}{c}{30.14}  &\multicolumn{1}{c}{30.34}  &\multicolumn{1}{c}{\underline{36.61}}  &\multicolumn{1}{c}{\textbf{39.86}}\\

\multicolumn{1}{c}{$*$$\to$TR}  &\multicolumn{1}{c}{31.66}  &\multicolumn{1}{c}{29.42}  &\multicolumn{1}{c}{28.99}  &\multicolumn{1}{c}{28.37}  &\multicolumn{1}{c}{\underline{35.67}}  &\multicolumn{1}{c}{\textbf{39.26}}\\

\multicolumn{1}{c}{Avg.}  &\multicolumn{1}{c}{41.22}  &\multicolumn{1}{c}{40.06}  &\multicolumn{1}{c}{38.22}  &\multicolumn{1}{c}{39.16}  &\multicolumn{1}{c}{\underline{44.54}}  &\multicolumn{1}{c}{\textbf{47.88}}\\


\toprule 

\rowcolor{Gray}
\multicolumn{1}{c}{[1k] \bf Pairs}  &\multicolumn{1}{c}{\bf RCSLS$^+$}  &\multicolumn{1}{c}{\bf VecMap$^{x}$} &\multicolumn{1}{c}{\bf LNMap} &\multicolumn{1}{c}{\bf FIPP}
&\multicolumn{1}{c}{\bf C1}
&\multicolumn{1}{c}{\bf C2 (C1)}
\\ \cmidrule(lr){2-5} \cmidrule(lr){6-7}

\multicolumn{1}{c}{DE$\to$$*$}  &\multicolumn{1}{c}{33.43}  &\multicolumn{1}{c}{36.69}  &\multicolumn{1}{c}{37.28}  &\multicolumn{1}{c}{37.70}  &\multicolumn{1}{c}{\underline{43.94}}  &\multicolumn{1}{c}{\textbf{46.61}}\\

\multicolumn{1}{c}{$*$$\to$DE}  &\multicolumn{1}{c}{32.23}  &\multicolumn{1}{c}{38.63}  &\multicolumn{1}{c}{36.74}  &\multicolumn{1}{c}{39.47}  &\multicolumn{1}{c}{\underline{43.15}}  &\multicolumn{1}{c}{\textbf{46.01}}\\

\multicolumn{1}{c}{EN$\to$$*$}  &\multicolumn{1}{c}{38.16}  &\multicolumn{1}{c}{38.63}  &\multicolumn{1}{c}{40.44}  &\multicolumn{1}{c}{42.26}  &\multicolumn{1}{c}{\underline{47.16}}  &\multicolumn{1}{c}{\textbf{49.84}}\\

\multicolumn{1}{c}{$*$$\to$EN}  &\multicolumn{1}{c}{38.57}  &\multicolumn{1}{c}{48.39}  &\multicolumn{1}{c}{43.61}  &\multicolumn{1}{c}{46.68}  &\multicolumn{1}{c}{\underline{51.59}}  &\multicolumn{1}{c}{\textbf{54.03}}\\

\multicolumn{1}{c}{FI$\to$$*$}  &\multicolumn{1}{c}{22.49}  &\multicolumn{1}{c}{33.08}  &\multicolumn{1}{c}{30.00}  &\multicolumn{1}{c}{32.11}  &\multicolumn{1}{c}{\underline{36.81}}  &\multicolumn{1}{c}{\textbf{40.28}}\\

\multicolumn{1}{c}{$*$$\to$FI}  &\multicolumn{1}{c}{22.29}  &\multicolumn{1}{c}{27.40}  &\multicolumn{1}{c}{29.95}  &\multicolumn{1}{c}{29.88}  &\multicolumn{1}{c}{\underline{36.61}}  &\multicolumn{1}{c}{\textbf{39.63}}\\

\multicolumn{1}{c}{FR$\to$$*$}  &\multicolumn{1}{c}{34.98}  &\multicolumn{1}{c}{38.65}  &\multicolumn{1}{c}{39.77}  &\multicolumn{1}{c}{41.08}  &\multicolumn{1}{c}{\underline{46.23}}  &\multicolumn{1}{c}{\textbf{48.57}}\\

\multicolumn{1}{c}{$*$$\to$FR}  &\multicolumn{1}{c}{36.83}  &\multicolumn{1}{c}{46.61}  &\multicolumn{1}{c}{43.81}  &\multicolumn{1}{c}{46.26}  &\multicolumn{1}{c}{\underline{49.75}}  &\multicolumn{1}{c}{\textbf{52.17}}\\

\multicolumn{1}{c}{HR$\to$$*$}  &\multicolumn{1}{c}{21.59}  &\multicolumn{1}{c}{33.22}  &\multicolumn{1}{c}{30.05}  &\multicolumn{1}{c}{30.93}  &\multicolumn{1}{c}{\underline{37.28}}  &\multicolumn{1}{c}{\textbf{42.16}}\\

\multicolumn{1}{c}{$*$$\to$HR}  &\multicolumn{1}{c}{20.87}  &\multicolumn{1}{c}{28.15}  &\multicolumn{1}{c}{27.67}  &\multicolumn{1}{c}{28.15}  &\multicolumn{1}{c}{\underline{34.00}}  &\multicolumn{1}{c}{\textbf{38.77}}\\

\multicolumn{1}{c}{IT$\to$$*$}  &\multicolumn{1}{c}{36.67}  &\multicolumn{1}{c}{39.45}  &\multicolumn{1}{c}{39.93}  &\multicolumn{1}{c}{42.20}  &\multicolumn{1}{c}{\underline{46.55}}  &\multicolumn{1}{c}{\textbf{49.22}}\\

\multicolumn{1}{c}{$*$$\to$IT}  &\multicolumn{1}{c}{38.33}  &\multicolumn{1}{c}{45.49}  &\multicolumn{1}{c}{43.47}  &\multicolumn{1}{c}{45.17}  &\multicolumn{1}{c}{\underline{48.50}}  &\multicolumn{1}{c}{\textbf{50.94}}\\

\multicolumn{1}{c}{RU$\to$$*$}  &\multicolumn{1}{c}{28.45}  &\multicolumn{1}{c}{37.75}  &\multicolumn{1}{c}{35.13}  &\multicolumn{1}{c}{38.24}  &\multicolumn{1}{c}{\underline{42.21}}  &\multicolumn{1}{c}{\textbf{44.61}}\\

\multicolumn{1}{c}{$*$$\to$RU}  &\multicolumn{1}{c}{27.78}  &\multicolumn{1}{c}{26.16}  &\multicolumn{1}{c}{29.71}  &\multicolumn{1}{c}{31.28}  &\multicolumn{1}{c}{\underline{38.02}}  &\multicolumn{1}{c}{\textbf{41.04}}\\

\multicolumn{1}{c}{TR$\to$$*$}  &\multicolumn{1}{c}{18.72}  &\multicolumn{1}{c}{26.97}  &\multicolumn{1}{c}{26.63}  &\multicolumn{1}{c}{27.05}  &\multicolumn{1}{c}{\underline{33.77}}  &\multicolumn{1}{c}{\textbf{36.89}}\\

\multicolumn{1}{c}{$*$$\to$TR}  &\multicolumn{1}{c}{17.59}  &\multicolumn{1}{c}{23.63}  &\multicolumn{1}{c}{24.26}  &\multicolumn{1}{c}{24.68}  &\multicolumn{1}{c}{\underline{32.34}}  &\multicolumn{1}{c}{\textbf{35.57}}\\

\multicolumn{1}{c}{Avg.}  &\multicolumn{1}{c}{29.31}  &\multicolumn{1}{c}{35.56}  &\multicolumn{1}{c}{34.90}  &\multicolumn{1}{c}{36.45}  &\multicolumn{1}{c}{\underline{41.74}}  &\multicolumn{1}{c}{\textbf{44.77}}\\

\bottomrule

\end{tabular}
}%

\caption{P@1 scores on the BLI benchmark of \newcite{glavas-etal-2019-properly} with bilingual supervision (i.e., $\mathcal{D}_0$ size) of $5k$ (upper half) and $1k$ translation pairs (bottom half). $L\to$$*$ and $*\to$$L$ denote the average BLI scores of BLI setups where $L$ is the source and the target language, respectively. The word similarity measure is CSLS (see \S\ref{s:experimental}). \underline{Underlined} scores are the peak scores among methods that rely solely on static fastText WEs; \textbf{Bold} scores denote the highest scores overall (i.e., the use of word translation knowledge exposed from mBERT is allowed). $^+$RCSLS is always used without self learning (see the footnote in \ref{s:experimental}); $^{x}$We report VecMap with self-learning in the $1k$-pairs scenario, and its variant without self-learning when using supervision of $5k$ pairs as it performs better than the variant with self-learning.}
\label{table:main}
\end{center}
\vspace{-1mm}
\end{table}

\rparagraph{Stage C1 versus Baselines}
First, we note that there is not a single strongest baseline among the four SotA BLI methods. For instance, RCSLS and VecMap are slightly better than LNMap and FIPP with $5k$ supervision pairs, while FIPP and VecMap come forth as the stronger baselines with $1k$ supervision. There are some score fluctuations over individual language pairs, but the average performance of all baseline models is within a relatively narrow interval: the average performance of all four baselines is within $3$ P@1 points with $5k$ pairs (i.e., ranging from $38.22$ to $41.22$), and VecMap, FIPP, and LNMap are within $2$ points with $1k$ pairs. 

Strikingly, contrastive learning in Stage C1 already yields substantial gains over all four SotA BLI models, which is typically much higher than the detected variations between the baselines. We mark that C1 improves over all baselines in $51/56$ BLI setups (in the $5k$ case), and in all $56/56$ BLI setups when $\mathcal{D}_0$ spans $1k$ pairs. The average gains with the C1 variant are $\approx$$5$ P@1 points over the SotA baselines with $5k$ pairs, and $\approx$$6$ P@1 points with $1k$ pairs (ignoring RCSLS in the $1k$ scenario). Note that all the models in comparison, each currently considered SotA in the BLI task, use exactly the same monolingual WEs and leverage exactly the same amount of bilingual supervision. The gains achieved with our Stage C1 thus strongly indicate the potential and usefulness of word-level contrastive fine-tuning when learning linear cross-lingual maps with static WEs (see RQ1 from \S\ref{s:introduction}).

\rparagraph{Stage C1 + Stage C2}
The scores improve further with the full two-stage procedure. The \textit{C2 (C1)} variant increases the average P@1 by another $3.3$ ($5k$) and $3$ P@1 points ($1k$), and we observe gains for all language pairs in both translation directions, rendering Stage C2 universally useful. These gains indicate that mBERT does contain word translation knowledge in its parameters. However, the model must be fine-tuned (i.e., transformed) to `unlock' the knowledge from its parameters: this is done through a BLI-guided contrastive fine-tuning procedure (see \S\ref{s:c2}). Our findings thus further confirm the `rewiring hypothesis' from prior work \cite{vulic2021lexfit,liu2021fast,gao2021simcse}, here validated for the BLI task (see RQ2 from \S\ref{s:introduction}), which states that task-relevant knowledge at sentence- and word-level can be `rewired'/exposed from the off-the-shelf LMs, even when leveraging very limited task supervision, e.g., with only $1k$ or $5k$ word translation pairs as in our experiments.

\rparagraph{Performance over Languages}
The absolute BLI scores naturally depend on the actual source and target languages: e.g., the lowest absolute performance is observed for morphologically rich (\textsc{hr}, \textsc{ru}, \textsc{fi}, \textsc{tr}) and  non-Indo-European languages (\textsc{fi}, \textsc{tr}). However, both C1 and C2 (C1) model variants offer wide and substantial gains in performance for \textit{all} language pairs, irrespective of the absolute score. This result further suggests wide applicability and robustness of our BLI method.

\begin{figure*}[!t]
    \centering
    \begin{subfigure}[!ht]{0.312\linewidth}
        \centering
        \includegraphics[width=0.96\linewidth]{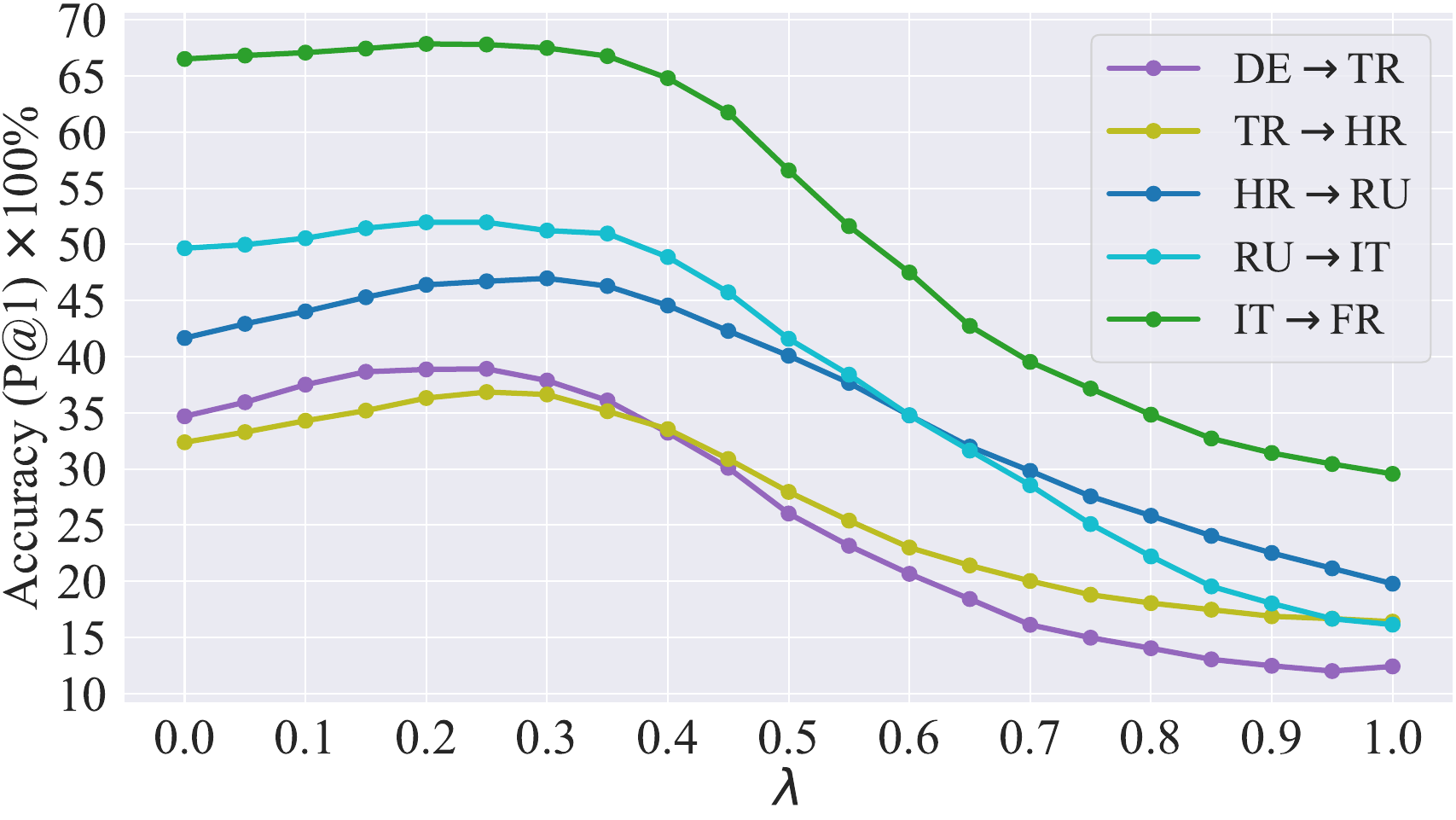}
        \label{fig:lambda5k}
    \end{subfigure}
    \begin{subfigure}[!ht]{0.312\textwidth}
        \centering
        \includegraphics[width=0.96\linewidth]{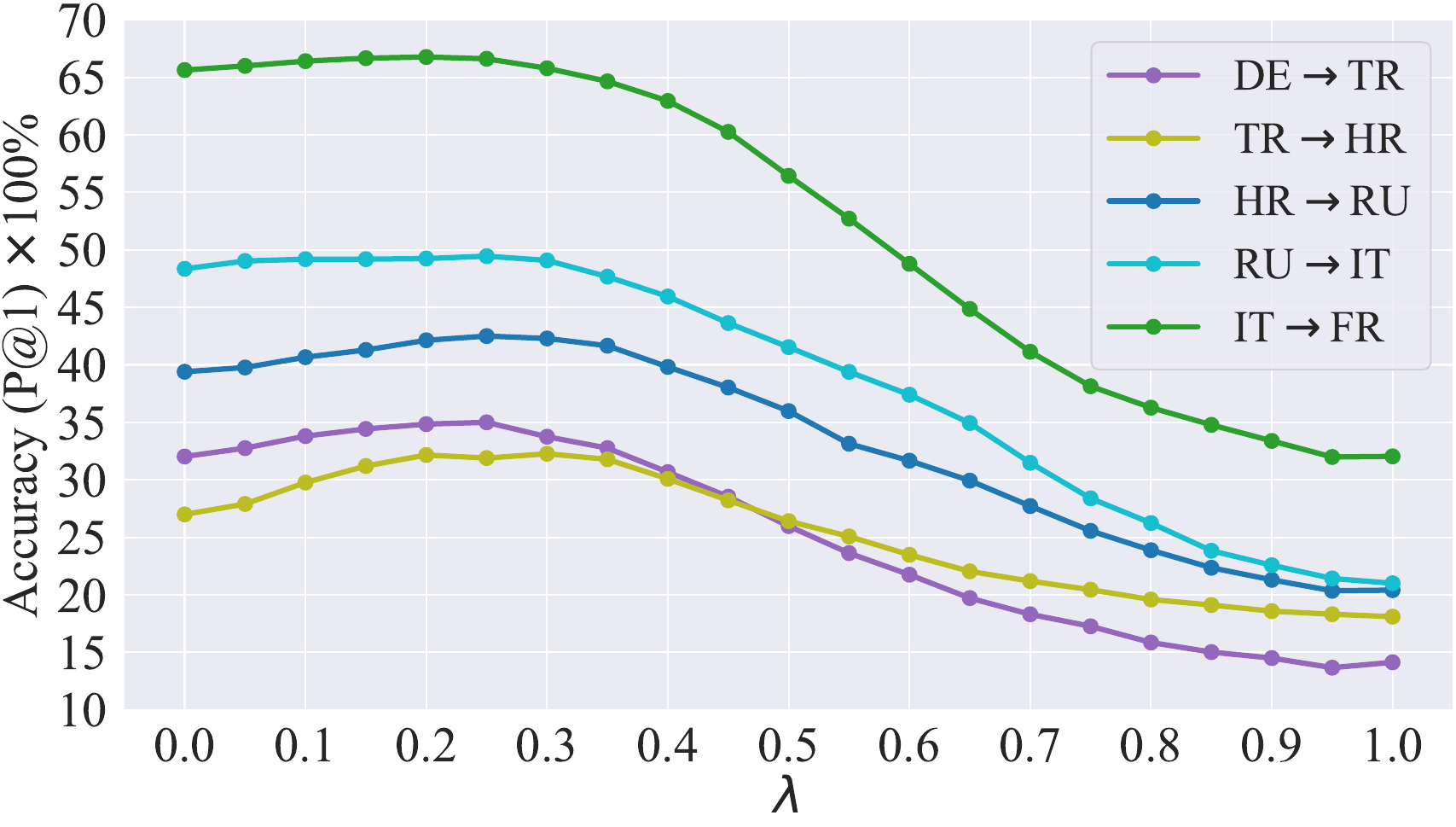}
        \label{fig:lambda1k}
    \end{subfigure}
    \begin{subfigure}[!ht]{0.312\textwidth}
        \centering
        \includegraphics[width=0.96\linewidth]{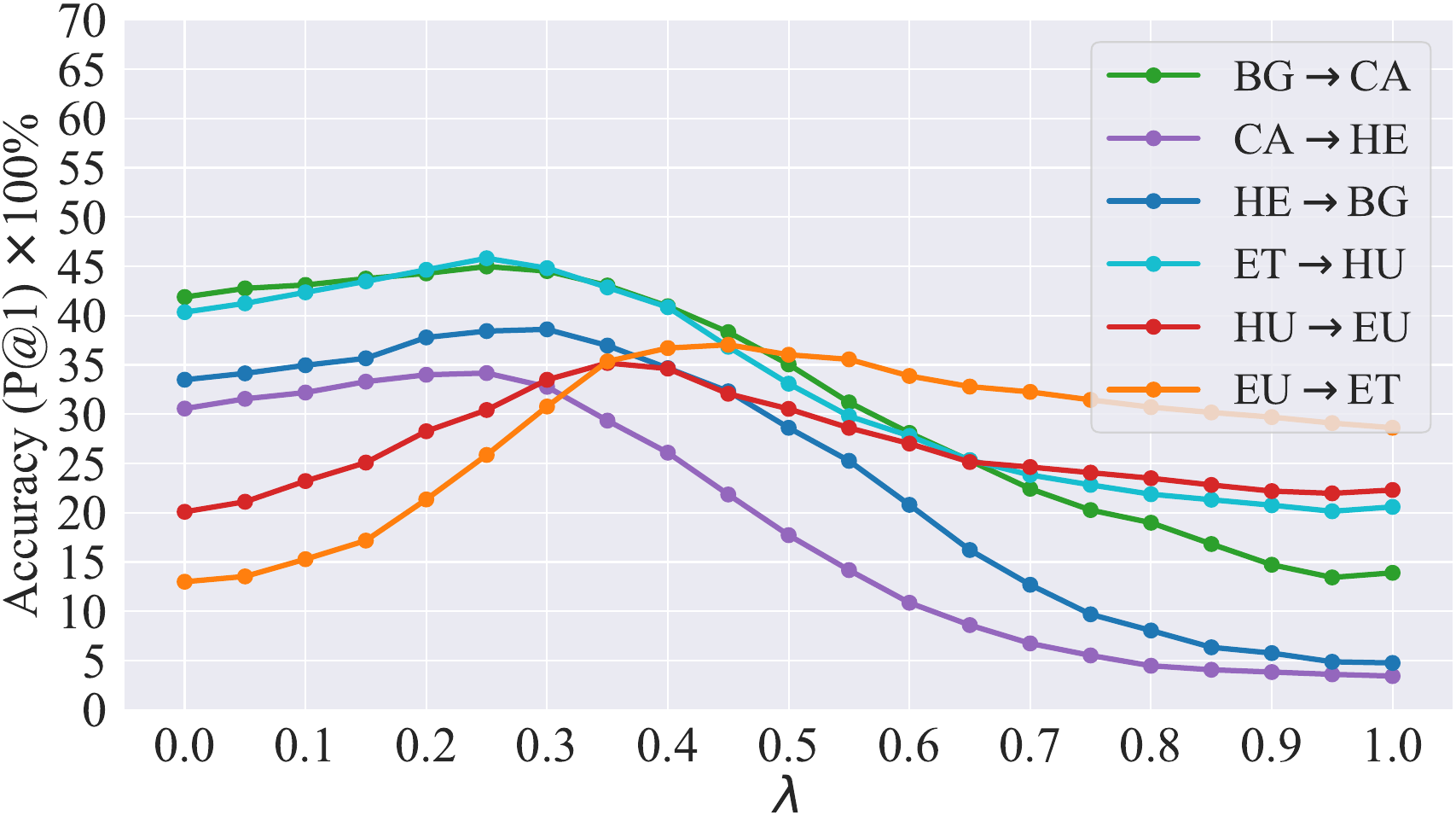}
        \label{fig:lambda1k_lowresource}
    \end{subfigure}
    \caption{BLI scores with different $\lambda$ values: (left) $|\mathcal{D}_0|$=$5k$; (middle) $|\mathcal{D}_0|$=$1k$; (right) PanLex-BLI, $|\mathcal{D}_0|$=$1k$.} 
    \vspace{-3mm}
\label{fig:lambda}
\end{figure*}

\subsection{Further Discussion}
\label{s:further}

\begin{table}[!t]
\begin{center}
\def\arraystretch{0.85}
\resizebox{0.38\textwidth}{!}{%
\begin{tabular}{llll}
\toprule 

\rowcolor{Gray}
\multicolumn{1}{c}{[1k] \textbf{Pairs}}  &\multicolumn{1}{c}{\bf BG$\to$CA}  &\multicolumn{1}{c}{\bf CA$\to$HE} &\multicolumn{1}{c}{\bf HE$\to$BG} 

\\ \cmidrule(lr){2-4}

\multicolumn{1}{c}{VecMap}  &\multicolumn{1}{c}{39.43}  &\multicolumn{1}{c}{24.64}  &\multicolumn{1}{c}{31.55}  \\

\multicolumn{1}{c}{FIPP}  &\multicolumn{1}{c}{34.29}  &\multicolumn{1}{c}{20.63}  &\multicolumn{1}{c}{26.38}  \\

\multicolumn{1}{c}{C1}  &\multicolumn{1}{c}{\underline{41.88}}  &\multicolumn{1}{c}{\underline{30.56}}  &\multicolumn{1}{c}{\underline{33.49}}  \\

\multicolumn{1}{c}{mBERT}  &\multicolumn{1}{c}{1.64}  &\multicolumn{1}{c}{1.28}  &\multicolumn{1}{c}{0.88}  \\

\multicolumn{1}{c}{mBERT (tuned)}  &\multicolumn{1}{c}{13.90}  &\multicolumn{1}{c}{ 3.43}  &\multicolumn{1}{c}{4.76}  \\

\multicolumn{1}{c}{C2 (C1)}  &\multicolumn{1}{c}{\textbf{44.28}}  &\multicolumn{1}{c}{\textbf{33.99}}  &\multicolumn{1}{c}{\textbf{37.78}}  \\

\toprule 

\rowcolor{Gray}
\multicolumn{1}{c}{[1k] \textbf{Pairs}}  &\multicolumn{1}{c}{\bf ET$\to$HU}  &\multicolumn{1}{c}{\bf HU$\to$EU} &\multicolumn{1}{c}{\bf EU$\to$ET} 

\\ \cmidrule(lr){2-4}

\multicolumn{1}{c}{VecMap}  &\multicolumn{1}{c}{35.55}  &\multicolumn{1}{c}{20.03}  &\multicolumn{1}{c}{9.83}  \\

\multicolumn{1}{c}{FIPP}  &\multicolumn{1}{c}{30.30}  &\multicolumn{1}{c}{11.58}  &\multicolumn{1}{c}{8.22}  \\

\multicolumn{1}{c}{C1}  &\multicolumn{1}{c}{\underline{40.35}}  &\multicolumn{1}{c}{\underline{20.09}}  &\multicolumn{1}{c}{\underline{13.00}}  \\

\multicolumn{1}{c}{mBERT}  &\multicolumn{1}{c}{15.40}  &\multicolumn{1}{c}{16.97}  &\multicolumn{1}{c}{23.70}  \\

\multicolumn{1}{c}{mBERT (tuned)}  &\multicolumn{1}{c}{20.59}  &\multicolumn{1}{c}{ 22.30}  &\multicolumn{1}{c}{28.62}  \\

\multicolumn{1}{c}{C2 (C1)}  &\multicolumn{1}{c}{\textbf{44.64}}  &\multicolumn{1}{c}{28.26}  &\multicolumn{1}{c}{21.35}  \\

\multicolumn{1}{c}{C2 (C1, $\lambda$=$0.4$)}  &\multicolumn{1}{c}{-}  &\multicolumn{1}{c}{\textbf{34.62}}  &\multicolumn{1}{c}{\textbf{36.70}}  \\

\bottomrule
\end{tabular}
}

\caption{BLI scores on the Panlex-BLI sets.}
\label{table:lowresource}
\end{center}
\vspace{-1mm}
\end{table}

\noindent \textbf{Evaluation on Lower-Resource Languages.}
The robustness of our BLI method is further tested on another BLI evaluation set: PanLex-BLI \cite{vulic-etal-2019-really}, which focuses on BLI evaluation for lower-resource language; $1k$ training pairs and $2k$ test pairs are derived from PanLex \cite{Kamholz2014panlex} for each BLI direction. The results for a subset of six languages (Basque: \textsc{eu}, Bulgarian: \textsc{bg}, Catalan: \textsc{ca}, Estonian: \textsc{et}, Hebrew: \textsc{he}, Hungarian: \textsc{hu}) are presented in Table~\ref{table:lowresource}. Overall, the results further confirm the efficacy of the \textit{C2 (C1)}, with gains observed even with typologically distant language pairs (e.g., \textsc{he}$\rightarrow$\textsc{bg} and \textsc{eu}$\rightarrow$\textsc{et}).

 
\begin{table}[!t]
\begin{center}
\def\arraystretch{0.85}
\resizebox{0.38\textwidth}{!}{%
\begin{tabular}{l lll}
\toprule 

\rowcolor{Gray}
\multicolumn{1}{c}{[5k] \textbf{Pairs}}  &\multicolumn{1}{c}{\bf DE$\to$TR}  &\multicolumn{1}{c}{\bf TR$\to$HR} &\multicolumn{1}{c}{\bf HR$\to$RU} 

\\ \cmidrule(lr){2-4}

\multicolumn{1}{c}{RCSLS}  &\multicolumn{1}{c}{30.99}  &\multicolumn{1}{c}{24.60}  &\multicolumn{1}{c}{37.19}  \\

\multicolumn{1}{c}{C2 (RCSLS)}  &\multicolumn{1}{c}{36.52}  &\multicolumn{1}{c}{33.17}  &\multicolumn{1}{c}{44.77}  \\

\multicolumn{1}{c}{VecMap}  &\multicolumn{1}{c}{27.18}  &\multicolumn{1}{c}{25.99}  &\multicolumn{1}{c}{37.98}  \\

\multicolumn{1}{c}{C2 (VecMap)}  &\multicolumn{1}{c}{34.95}  &\multicolumn{1}{c}{34.29}  &\multicolumn{1}{c}{44.98}  \\

\multicolumn{1}{c}{C1}  &\multicolumn{1}{c}{\underline{34.69}}  &\multicolumn{1}{c}{\underline{32.37}}  &\multicolumn{1}{c}{\underline{41.66}}  \\

\multicolumn{1}{c}{C2 (C1)}  &\multicolumn{1}{c}{\textbf{38.86}}  &\multicolumn{1}{c}{\textbf{36.32}}  &\multicolumn{1}{c}{\textbf{46.40}}  \\

\toprule 

\rowcolor{Gray}
\multicolumn{1}{c}{[1k] \textbf{Pairs}}  &\multicolumn{1}{c}{\bf DE$\to$TR}  &\multicolumn{1}{c}{\bf TR$\to$HR} &\multicolumn{1}{c}{\bf HR$\to$RU} 

\\ \cmidrule(lr){2-4}

\multicolumn{1}{c}{RCSLS}  &\multicolumn{1}{c}{18.21}  &\multicolumn{1}{c}{13.84}  &\multicolumn{1}{c}{24.72}  \\

\multicolumn{1}{c}{C2 (RCSLS)}  &\multicolumn{1}{c}{25.40}  &\multicolumn{1}{c}{22.52}  &\multicolumn{1}{c}{33.88}  \\

\multicolumn{1}{c}{VecMap}  &\multicolumn{1}{c}{23.37}  &\multicolumn{1}{c}{20.50}  &\multicolumn{1}{c}{36.09}  \\

\multicolumn{1}{c}{C2 (VecMap)}  &\multicolumn{1}{c}{27.91}  &\multicolumn{1}{c}{26.84}  &\multicolumn{1}{c}{40.45}  \\

\multicolumn{1}{c}{C1}  &\multicolumn{1}{c}{\underline{32.03}}  &\multicolumn{1}{c}{\underline{27.00}}  &\multicolumn{1}{c}{\underline{39.40}}  \\

\multicolumn{1}{c}{C2 (C1)}  &\multicolumn{1}{c}{\textbf{34.85}}  &\multicolumn{1}{c}{\textbf{32.16}}  &\multicolumn{1}{c}{\textbf{42.14}}  \\

 \bottomrule

\end{tabular}
}

\caption{Stage C2 with different `support' methods: RCSLS, VecMap, and C1. P@1$\times100\%$ scores.}
\label{table:C2onother}
\end{center}
\vspace{-1mm}
\end{table}


\begin{table}[!t]
\begin{center}
\def\arraystretch{0.85}
\resizebox{0.45\textwidth}{!}{%
\begin{tabular}{lllll}
\toprule 

\rowcolor{Gray}
\multicolumn{1}{c}{[5k] \bf Pairs}  &\multicolumn{1}{c}{\bf C1}  &\multicolumn{1}{c}{\bf C2 [mBERT]} &\multicolumn{1}{c}{\bf C2 [XLM]} &\multicolumn{1}{c}{\bf C2 [mT5]}

\\ \cmidrule(lr){2-2} \cmidrule(lr){3-5} 

\multicolumn{1}{c}{DE$\to$TR}  &\multicolumn{1}{c}{34.69}  &\multicolumn{1}{c}{\textbf{38.86}}  &\multicolumn{1}{c}{38.08}  &\multicolumn{1}{c}{37.19}\\

\multicolumn{1}{c}{EN$\to$IT}  &\multicolumn{1}{c}{63.45}  &\multicolumn{1}{c}{\textbf{65.60}}  &\multicolumn{1}{c}{65.45}  &\multicolumn{1}{c}{64.15}\\

\multicolumn{1}{c}{EN$\to$HR}  &\multicolumn{1}{c}{40.70}  &\multicolumn{1}{c}{\textbf{47.20}}  &\multicolumn{1}{c}{45.20}  &\multicolumn{1}{c}{43.00}\\

\multicolumn{1}{c}{FI$\to$RU}  &\multicolumn{1}{c}{37.73}  &\multicolumn{1}{c}{\textbf{40.99}}  &\multicolumn{1}{c}{37.94}  &\multicolumn{1}{c}{38.36}\\

\multicolumn{1}{c}{HR$\to$RU}  &\multicolumn{1}{c}{41.66}  &\multicolumn{1}{c}{\textbf{46.40}}  &\multicolumn{1}{c}{46.29}  &\multicolumn{1}{c}{43.87}\\

\multicolumn{1}{c}{IT$\to$FR}  &\multicolumn{1}{c}{66.51}  &\multicolumn{1}{c}{\textbf{67.86}}  &\multicolumn{1}{c}{66.61}  &\multicolumn{1}{c}{67.34}\\

\multicolumn{1}{c}{RU$\to$IT}  &\multicolumn{1}{c}{49.66}  &\multicolumn{1}{c}{51.96}  &\multicolumn{1}{c}{\textbf{52.33}}  &\multicolumn{1}{c}{50.39}\\

\multicolumn{1}{c}{TR$\to$HR}  &\multicolumn{1}{c}{32.37}  &\multicolumn{1}{c}{\textbf{36.32}}  &\multicolumn{1}{c}{32.22}  &\multicolumn{1}{c}{34.56}\\

\toprule 

\rowcolor{Gray}
\multicolumn{1}{c}{[1k] \bf Pairs}  &\multicolumn{1}{c}{\bf C1}  &\multicolumn{1}{c}{\bf C2 [mBERT]} &\multicolumn{1}{c}{\bf C2 [XLM]} &\multicolumn{1}{c}{\bf C2 [mT5]}
\\ \cmidrule(lr){2-2} \cmidrule(lr){3-5} 

\multicolumn{1}{c}{DE$\to$TR}  &\multicolumn{1}{c}{32.03}  &\multicolumn{1}{c}{\textbf{34.85}}  &\multicolumn{1}{c}{31.66}  &\multicolumn{1}{c}{34.43}\\

\multicolumn{1}{c}{EN$\to$IT}  &\multicolumn{1}{c}{59.60}  &\multicolumn{1}{c}{61.05}  &\multicolumn{1}{c}{\textbf{61.80}}  &\multicolumn{1}{c}{60.05}\\

\multicolumn{1}{c}{EN$\to$HR}  &\multicolumn{1}{c}{35.65}  &\multicolumn{1}{c}{\textbf{42.35}}  &\multicolumn{1}{c}{41.75}  &\multicolumn{1}{c}{39.40}\\

\multicolumn{1}{c}{FI$\to$RU}  &\multicolumn{1}{c}{33.89}  &\multicolumn{1}{c}{37.15}  &\multicolumn{1}{c}{\textbf{38.36}}  &\multicolumn{1}{c}{36.00}\\

\multicolumn{1}{c}{HR$\to$RU}  &\multicolumn{1}{c}{39.40}  &\multicolumn{1}{c}{42.14}  &\multicolumn{1}{c}{\textbf{43.35}}  &\multicolumn{1}{c}{41.45}\\

\multicolumn{1}{c}{IT$\to$FR}  &\multicolumn{1}{c}{65.63}  &\multicolumn{1}{c}{\textbf{66.77}}  &\multicolumn{1}{c}{66.51}  &\multicolumn{1}{c}{66.15}\\

\multicolumn{1}{c}{RU$\to$IT}  &\multicolumn{1}{c}{48.35}  &\multicolumn{1}{c}{49.24}  &\multicolumn{1}{c}{\textbf{50.86}}  &\multicolumn{1}{c}{49.24}\\

\multicolumn{1}{c}{TR$\to$HR}  &\multicolumn{1}{c}{27.00}  &\multicolumn{1}{c}{\textbf{32.16}}  &\multicolumn{1}{c}{27.05}  &\multicolumn{1}{c}{30.35}\\

\bottomrule
\end{tabular}
}

\caption{Stage C2 with different pretrained LMs: mBERT, XLM, and mT5. P@1$\times100\%$ scores.}
\label{table:morePLMs}
\end{center}
\end{table}

\vspace{0.5mm}
\noindent \textbf{Usefulness of Stage C2?} The results in Table~\ref{table:main} have confirmed the effectiveness of our two-stage \textit{C2 (C1)} BLI method (see RQ3 in \S\ref{s:introduction}). However, Stage C2 is in fact independent of our Stage C1, and thus can also be combined with other standard BLI methods. Therefore, we seek to validate whether combining exposed mBERT-based translation knowledge can also aid other BLI methods. In other words, instead of drawing positive and negative samples from Stage C1 (\S\ref{s:c2}) and combining C2 WEs with WEs from C1 (\S\ref{s:combined}), we replace C1 with our baseline models. The results of these \textit{C2 (RCSLS)} and \textit{C2 (VecMap)} variants for a selection of language pairs are provided in Table~\ref{table:C2onother}.

The gains achieved with all \textit{C2 ($\cdot$)} variants clearly indicate that Stage C2 produces WEs which aid all BLI methods. In fact, combining it with RCSLS and VecMap yields even larger relative gains over the base models than combining it with our Stage C1. However, since Stage C1 (as the base model) performs better than RCSLS and VecMap, the final absolute scores with \textit{C2 (C1)} still outperform \textit{C2 (RCSLS)} and \textit{C2 (VecMap)}.

\vspace{0.5mm}
\noindent \textbf{Different Multilingual LMs?} Results on eight language pairs, shown in Table~\ref{table:morePLMs}, indicate that \textit{C2 (C1)} is also compatible with different LMs. The overall trend is that all three \textit{C2 [LM]} variants derive some gains when compared to C1. \textit{C2 [mBERT]} is the best-performing model and derives gains in all $112/112$ BLI setups (also see Appendix~\ref{appendix:full}); \textit{C2 [mT5]} outperforms C1 in all $16/16$ cases, and the gains are observed for $14/16$ cases with \textit{C2 [XLM]}. It is also worth noticing that \textit{C2 [XLM]} can surpass \textit{C2 [mBERT]} on several pairs.

\vspace{0.5mm}
\noindent \textbf{Combining C1 and C2?}
The usefulness of combining the representations from two stages is measured through varying the value of $\lambda$ for several BLI setups. The plots are shown in Figure~\ref{fig:lambda}, and indicate that Stage C1 is more beneficial to the performance, with slight gains achieved when allowing the `influx' of mBERT knowledge (e.g., $\lambda$ in the $[0.0,0.3]$ interval). While mBERT-based WEs are not sufficient as standalone representations for BLI, they seem to be even more useful in the combined model for lower-resource languages on PanLex-BLI, with steeper increase in performance, and peak scores achieved with larger $\lambda$ values.

\begin{table}[!t]
\begin{center}
\def\arraystretch{0.85}
\resizebox{0.38\textwidth}{!}{%
\begin{tabular}{l lll}

\toprule 
\rowcolor{Gray}
\multicolumn{1}{c}{[5k] \textbf{Pairs}}  &\multicolumn{1}{c}{\bf EN$\to$$*$}  &\multicolumn{1}{c}{\bf DE$\to$$*$} &\multicolumn{1}{c}{\bf IT$\to$$*$} 

\\ \cmidrule(lr){2-4}

\multicolumn{1}{c}{C1 w/o CL}  &\multicolumn{1}{c}{41.58}  &\multicolumn{1}{c}{39.30} &\multicolumn{1}{c}{42.67} \\

\multicolumn{1}{c}{C1 w/o SL}  &\multicolumn{1}{c}{50.99}  &\multicolumn{1}{c}{45.07} &\multicolumn{1}{c}{48.39} \\

\multicolumn{1}{c}{C1}  &\multicolumn{1}{c}{\underline{51.31}}  &\multicolumn{1}{c}{\underline{46.14}} &\multicolumn{1}{c}{\underline{48.92}} \\

\multicolumn{1}{c}{mBERT}  &\multicolumn{1}{c}{9.55}  &\multicolumn{1}{c}{9.39} &\multicolumn{1}{c}{8.13} \\

\multicolumn{1}{c}{mBERT (tuned)}  &\multicolumn{1}{c}{15.87}  &\multicolumn{1}{c}{18.66} &\multicolumn{1}{c}{20.18} \\

\multicolumn{1}{c}{C1 + mBERT}  &\multicolumn{1}{c}{51.55}  &\multicolumn{1}{c}{46.25} &\multicolumn{1}{c}{48.91} \\

\multicolumn{1}{c}{C2 (C1)}  &\multicolumn{1}{c}{\textbf{54.31}}  &\multicolumn{1}{c}{\textbf{48.86}} &\multicolumn{1}{c}{\textbf{51.91}} \\


\toprule 
\rowcolor{Gray}
\multicolumn{1}{c}{[1k] \textbf{Pairs}}  &\multicolumn{1}{c}{\bf EN$\to$$*$}  &\multicolumn{1}{c}{\bf DE$\to$$*$} &\multicolumn{1}{c}{\bf IT$\to$$*$} 

\\ \cmidrule(lr){2-4}

\multicolumn{1}{c}{C1 w/o CL}  &\multicolumn{1}{c}{39.46}  &\multicolumn{1}{c}{37.54} &\multicolumn{1}{c}{40.37} \\

\multicolumn{1}{c}{C1 w/o SL}  &\multicolumn{1}{c}{39.31}  &\multicolumn{1}{c}{32.59} &\multicolumn{1}{c}{36.45} \\

\multicolumn{1}{c}{C1}  &\multicolumn{1}{c}{\underline{47.16}}  &\multicolumn{1}{c}{\underline{43.94}} &\multicolumn{1}{c}{\underline{46.55}} \\

\multicolumn{1}{c}{mBERT}  &\multicolumn{1}{c}{9.55}  &\multicolumn{1}{c}{9.39} &\multicolumn{1}{c}{8.13} \\

\multicolumn{1}{c}{mBERT (tuned)}  &\multicolumn{1}{c}{17.29}  &\multicolumn{1}{c}{20.92} &\multicolumn{1}{c}{23.29} \\

\multicolumn{1}{c}{C1 + mBERT}  &\multicolumn{1}{c}{47.56}  &\multicolumn{1}{c}{44.08} &\multicolumn{1}{c}{46.74} \\

\multicolumn{1}{c}{C2 (C1)}  &\multicolumn{1}{c}{\textbf{49.84}}  &\multicolumn{1}{c}{\textbf{46.61}} &\multicolumn{1}{c}{\textbf{49.22}} \\

\bottomrule

\end{tabular}
}

\vspace{-1mm}
\caption{Ablation study. CL = Contrastive Learning; SL = Self-Learning. `mBERT' and `mBERT (tuned)' refer to using word encodings from mBERT directly for BLI, before and after fine-tuning in Stage C2. Very similar trends are observed for all other language pairs (available in Appendix~\ref{appendix:ablation}).}
\label{table:ablationmain}
\end{center}
\vspace{-2mm}
\end{table}

\vspace{0.5mm}
\noindent \textbf{Ablation Study}, with results summarised in Table~\ref{table:ablationmain}, displays several interesting trends. First, both CL and self-learning are key components in the $1k$-setups: removing any of them yields substantial drops. In $5k$-setups, self-learning becomes less important, and removing it yields only negligible drops, while CL remains a crucial component (see also Appendix~\ref{appendix:ablation}). Further, Table~\ref{table:ablationmain} complements the results from Figure~\ref{fig:lambda} and again indicates that, while Stage C2 indeed boosts word translation capacity of mBERT, using mBERT features alone is still not sufficient to achieve competitive BLI scores. After all, pretrained LMs are contextualised encoders designed for (long) sequences rather than individual words or tokens. Finally, Table~\ref{table:ablationmain} shows the importance of fine-tuning mBERT before combining it with C1-based WEs (\S\ref{s:combined}): directly adding WEs extracted from the off-the-shelf mBERT does not yield any benefits (see the scores for the \textit{C1+mBERT} variant, where $\lambda$ is also $0.2$). 

\begin{figure}[!t]
    \centering
    \begin{subfigure}[!ht]{0.49\linewidth}
        \centering
        \includegraphics[width=0.995\linewidth]{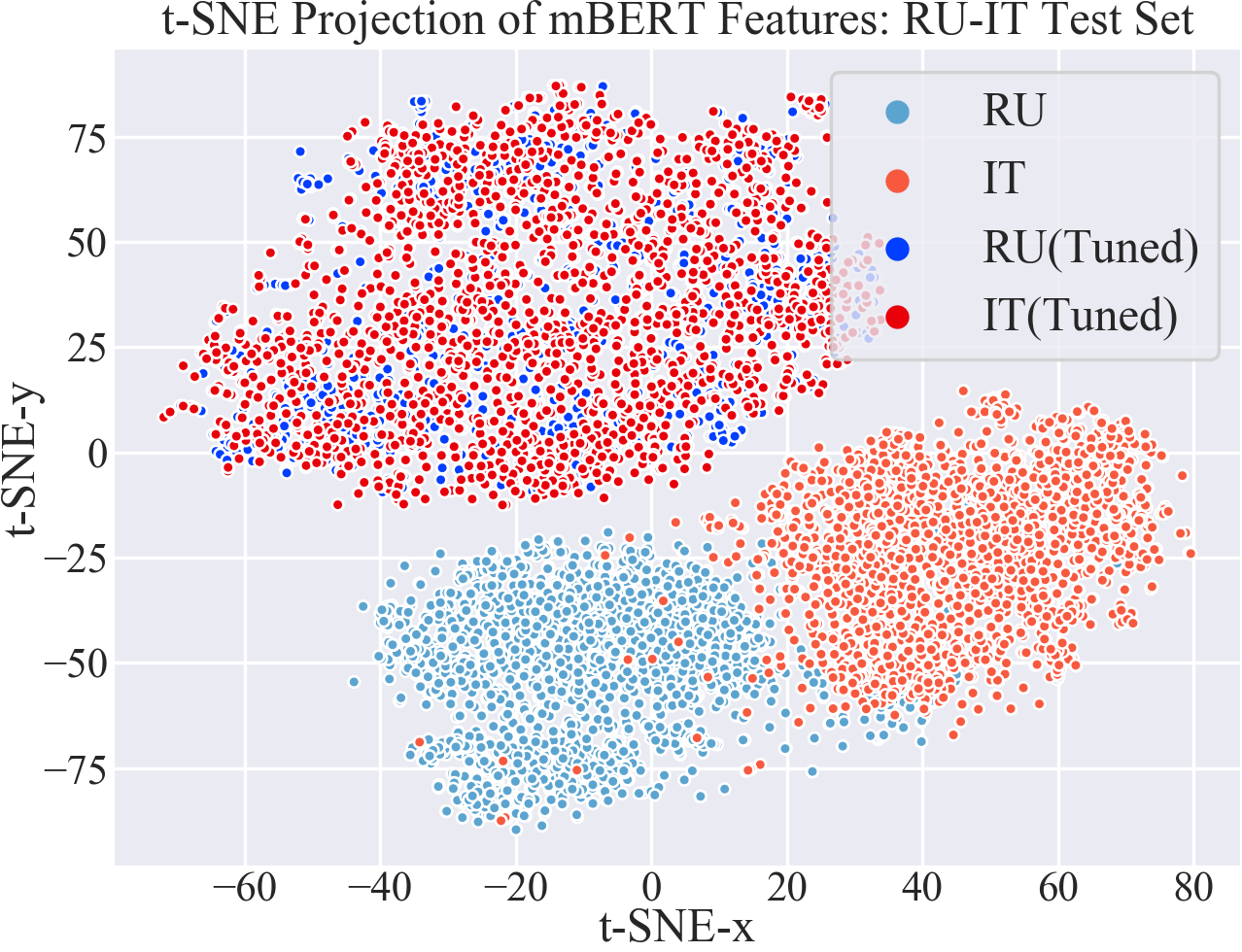}
        \label{fig:ruit}
    \end{subfigure}
    \begin{subfigure}[!ht]{0.49\linewidth}
        \centering
        \includegraphics[width=0.995\linewidth]{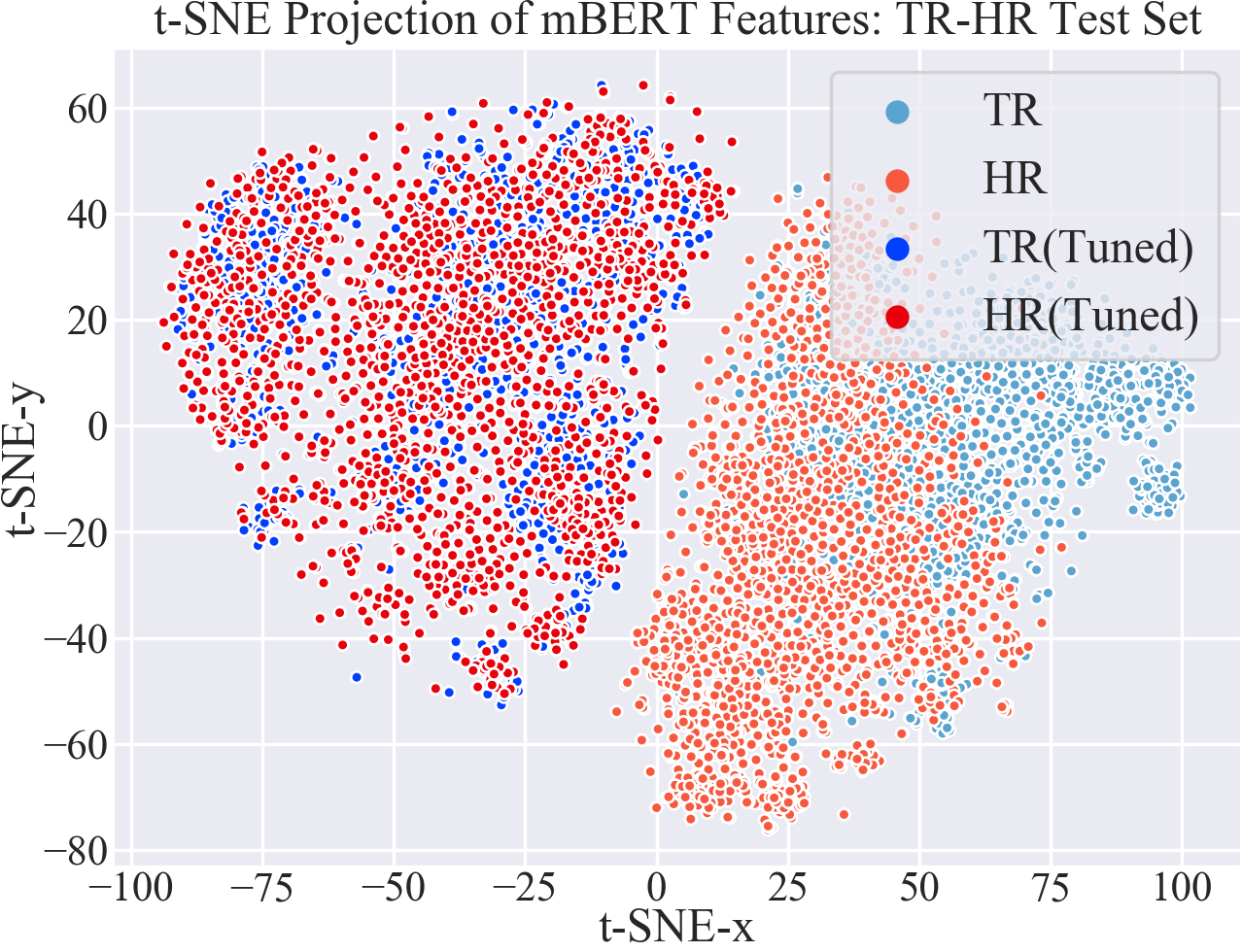}
        \label{fig:trhr}
    \end{subfigure}
    \caption{A t-SNE visualisation \cite{tsne:2012} of mBERT encodings of words from BLI test sets for \textsc{ru}-\textsc{it} (left) and \textsc{tr}-\textsc{hr} (right). Similar plots for more language pairs are in Appendix~\ref{appendix:tsne}.} 
    \vspace{-1mm}
\label{fig:tsne-main}
\end{figure}

\begin{figure}[!t]
    \centering
    \begin{subfigure}[!ht]{0.49\linewidth}
        \centering
        \includegraphics[width=0.995\linewidth]{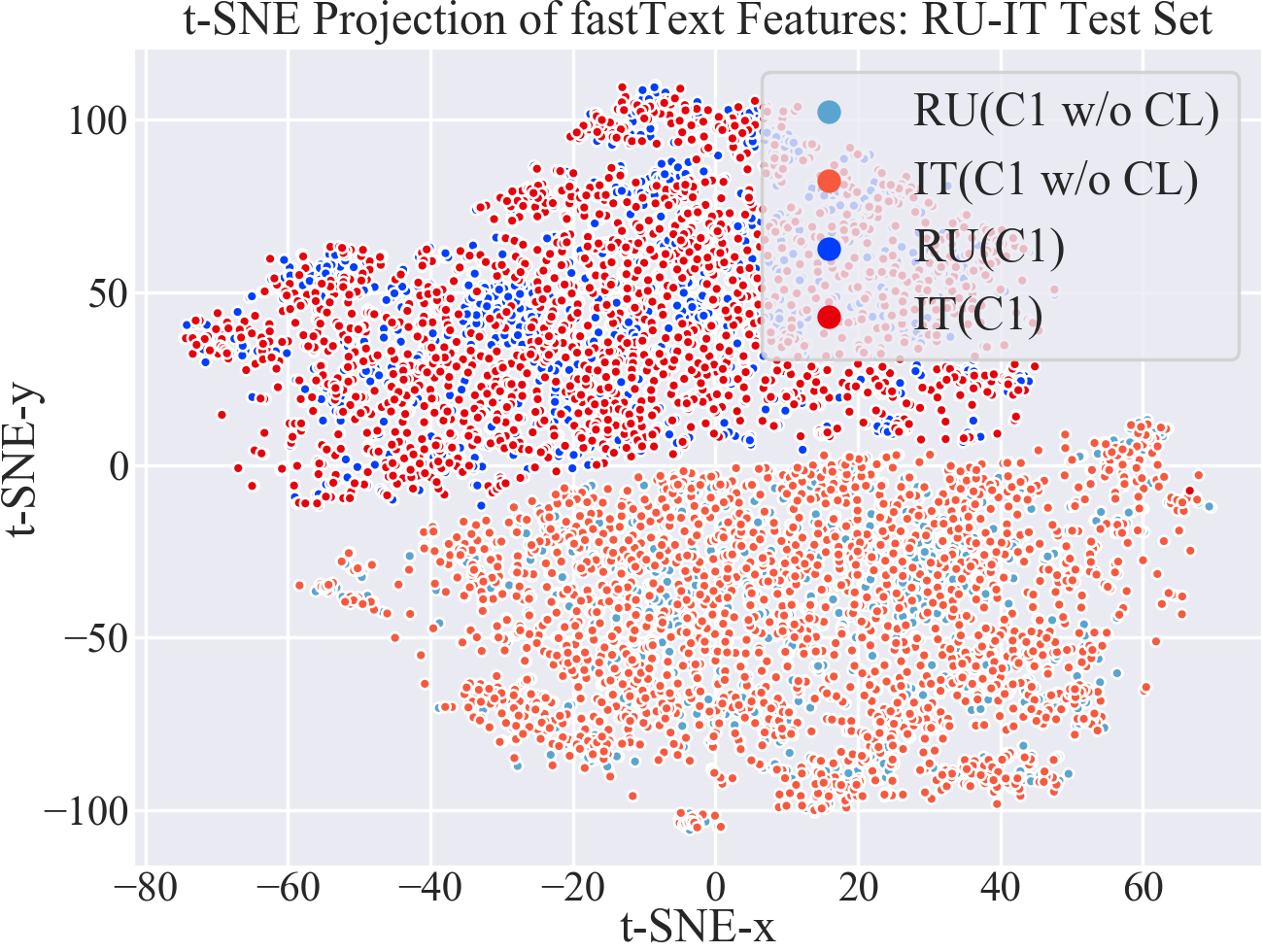}
        \label{fig:ruit-ft}
    \end{subfigure}
    \begin{subfigure}[!ht]{0.49\linewidth}
        \centering
        \includegraphics[width=0.995\linewidth]{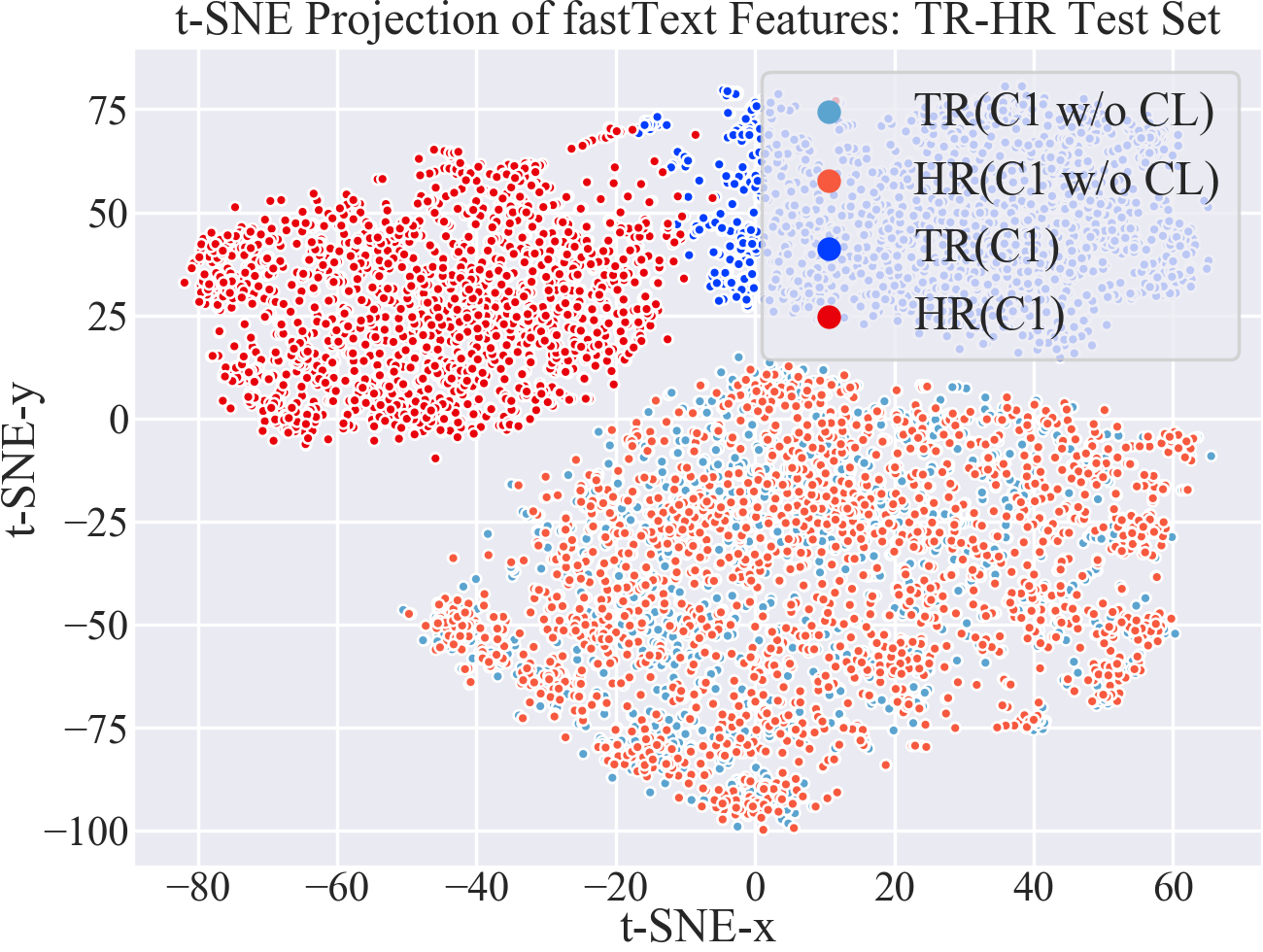}
        \label{fig:trhr-ft}
    \end{subfigure}
    \vspace{-2mm}
    \caption{A t-SNE visualisation \cite{tsne:2012} of mapped fastText WEs of words from BLI test sets for \textsc{ru}-\textsc{it} (left) and \textsc{tr}-\textsc{hr} (right). Similar plots for more language pairs are in Appendix~\ref{appendix:tsne_fastText}.} 
    \vspace{-2mm}
\label{fig:tsne-ft}
\end{figure}

\vspace{0.5mm}
\noindent \textbf{The Impact of Contrastive Fine-Tuning} on mBERT's representation space for two language pairs is illustrated by a t-SNE plot in Figure~\ref{fig:tsne-main}. The semantic space of off-the-shelf mBERT displays a clear separation of language-specific subspaces \cite{Libovicky:2020emnlp,Dufter:2020emnlp}, which makes it unsuitable for the BLI task. On the other hand, contrastive fine-tuning reshapes the subspaces towards a shared (cross-lingual) space, the effects of which are then also reflected in mBERT's improved BLI capability (see Table~\ref{table:ablationmain} again). 

To understand the role of CL in Stage C1, we visualise static WEs mapped by C1 without CL (i.e., AM+SL, see \S\ref{s:c1}) and also from the complete Stage C1, respectively. Figure~\ref{fig:tsne-ft} shows that C1 without CL already learns a sensible cross-lingual space. However, we note that advanced mapping (AM) in C1 without CL learns a (near-)orthogonal map, which might result in mismatches, especially with dissimilar language pairs. With TR-HR, the plot reveals that there exists a gap between C1-aligned WE spaces although the final BLI performance still gets improved: this might be due to `repelling' negatives from each other during CL. 

Finally, we direct interested readers to Appendix~\ref{appendix:transexamples} where we present some qualitative translation examples.

%% file: 5_related_work.tex
This work is related to three topics, each with a large body of work;  we can thus provide only a condensed summary of the most relevant research.

\rparagraph{Mapping-Based BLI} 
These BLI methods are highly popular due to reduced bilingual supervision requirements; consequently, they are applicable to low-resource languages and domains, learning linear \cite{conneau2017word,artetxe2018robust,joulin-etal-2018-loss,patra-etal-2019-bilingual,10.1162/tacl_a_00257,sachidananda2021filtered} and non-linear maps \cite{mohiuddin-etal-2020-lnmap,glavas-vulic-2020-non,ganesan2021learning}, typically using self-learning in weakly supervised setups. 


\noindent \textbf{Contrastive Learning in NLP}
aims to learn a semantic space where embeddings of similar text inputs are close to each other, while those of dissimilar ones are `repelled' apart. It has shown remarkable success in training generic sentence encoders \citep{giorgi-etal-2021-declutr,carlsson2021semantic,liu2021self,gao2021simcse} and promising performance on downstream tasks like summarisation \citep{liu2021simcls} or NER \citep{das2021container}.

\rparagraph{Exposing Lexical Knowledge from Pretrained LMs} 
Extracting lexical features from off-the-shelf multilingual LMs typically yields subpar performance in lexical tasks \cite{vulic-etal-2020-probing}. To unlock the lexical knowledge encoded in PLMs, \citet{liu2021self} and \citet{vulic2021lexfit} fine-tune LMs via contrastive learning with manually curated or automatically extracted phrase/word pairs to transform it into effective text encoders. \citet{wang-etal-2021-phrase} and \citet{liu2021mirrorwic} apply similar techniques for phrase and word-in-context representation learning, respectively. The success of these methods suggests that LMs store a wealth of lexical knowledge: yet, as we confirm here for BLI, fine-tuning is typically needed to expose it.


%% file: 6_conclusion.tex
We have proposed a simple yet extremely effective and robust two-stage contrastive learning framework for improving bilingual lexicon induction (BLI). In Stage C1, we tune cross-lingual linear mappings between static word embeddings with a contrastive objective and achieve substantial gains in $107$ out of $112$ BLI setups on the standard BLI benchmark. In Stage C2, we further propose a contrastive fine-tuning procedure to harvest cross-lingual lexical knowledge from multilingual pretrained language models. The representations from this process, when combined with Stage C1 embeddings, have resulted in further boosts in BLI performance, with large gains in all $112$ setups. We have also conducted a series of finer-grained evaluations, analyses and ablation studies. 

%% file: 7_acknowledgements.tex
{\scriptsize\euflag} We thank the anonymous reviewers for their valuable feedback. This work is supported by the ERC PoC Grant MultiConvAI (no. 957356) and a research donation from Huawei. YL and FL are supported by Grace $\&$ Thomas C. H. Chan Cambridge International Scholarship.

%% file: 8_ethics.tex
Our research aims to benefit the efforts in delivering truly multilingual language technology also to under-resourced languages and cultures via bridging the lexical gap between languages, groups and cultures.  As a key task in cross-lingual NLP, bilingual lexicon induction or word translation has broad applications in, e.g., machine translation, language acquisition and potentially protecting endangered languages. Furthermore, compared with many previous studies, we stress the importance of diversity in the sense that our experiments cover various language families and include six lower-resource languages from the PanLex-BLI dataset. Hoping that our work can contribute to extending modern NLP techniques to lower-resource and under-represented languages, we focus on semi-supervised settings and achieve significant improvements with self-learning techniques.

The two BLI datasets we use are both publicly available. To our best knowledge, the data (i.e., word translation pairs) do not contain any sensitive information and have no foreseeable risk.

%% file: x_appendix.tex
\appendix

\section{Technical Details and Further Clarifications}
\subsection{Advanced Mapping (AM) in Stage C1}
\label{appendix:am}
Suppose $\bm{X}_{\mathcal{D}}, \bm{Y}_{\mathcal{D}} \in \mathbb{R}^{|\mathcal{D}|\times d}$ are source and target embedding matrices corresponding to the training dictionary $\mathcal{D}$. Then, whitening transformation is applied on them respectively, and we derive $\bm{X}_{\mathcal{D}}^{'}$ and $\bm{Y}_{\mathcal{D}}^{'}$. After that, we compute the singular value decomposition (SVD) of $\bm{X}_{\mathcal{D}}^{'^{T}}\bm{Y}_{\mathcal{D}}^{'}$:

\begin{equation}
\begin{aligned}
\bm{X}_{\mathcal{D}}^{'} = \bm{X}_{\mathcal{D}} (\bm{X}_{\mathcal{D}}^{T}\bm{X}_{\mathcal{D}})^{-\frac{1}{2}},
\end{aligned}
\label{formula:a101}
\end{equation}
\begin{equation}
\begin{aligned}
\bm{Y}_{\mathcal{D}}^{'} = \bm{Y}_{\mathcal{D}} (\bm{Y}_{\mathcal{D}}^{T}\bm{Y}_{\mathcal{D}})^{-\frac{1}{2}}, \\
\end{aligned}
\label{formula:a102}
\end{equation}
\begin{equation}
\begin{aligned}
\bm{U}\bm{S}\bm{V}^{T} = \bm{X}_{\mathcal{D}}^{'^{T}}\bm{Y}_{\mathcal{D}}^{'}.
\end{aligned}
\label{formula:a103}
\end{equation}
$\bm{W}_{x}$ and $\bm{W}_{y}$ are then derived after re-weighting and de-whitening as follows:

\begin{equation}
\begin{aligned}
\bm{W}_{x}\!=\!(\bm{X}_{\mathcal{D}}^{T}\!\bm{X}_{\mathcal{D}})^{-\frac{1}{2}}\bm{U}\bm{S}^{\frac{1}{2}}\bm{U}^{T}(\bm{X}_{\mathcal{D}}^{T}\!\bm{X}_{\mathcal{D}})^{\frac{1}{2}}\bm{U},
\end{aligned}
\label{formula:a104}
\end{equation}

\begin{equation}
\begin{aligned}
\bm{W}_{y}=(\bm{Y}_{\mathcal{D}}^{T}\bm{Y}_{\mathcal{D}})^{-\frac{1}{2}}\bm{V}\bm{S}^{\frac{1}{2}}\bm{V}^{T}(\bm{Y}_{\mathcal{D}}^{T}\bm{Y}_{\mathcal{D}})^{\frac{1}{2}}\bm{V}.
\end{aligned}
\label{formula:a105}
\end{equation}






\subsection{Word Similarity/Retrieval Measures}
\label{appendix:csls}
Here, we denote the already aligned word embedding matrices of the source language $L_x$ and the target language $L_y$ as $\bm{X}$ and $\bm{Y}$ respectively. Given two word embeddings $\mathbf{x}$ and $\mathbf{y}$ from the already aligned cross-lingual WE space (they are row vectors of $\bm{X}$ and $\bm{Y}$ respectively), their similarity can be defined as their cosine similarity $m(\mathbf{x},\mathbf{y})=\text{cosine}(\mathbf{x},\mathbf{y})$. In the FIPP model, we calculate dot product between $\mathbf{x}$ and $\mathbf{y}$ instead without normalisation, as with FIPP this produces better BLI scores in general.\footnote{\url{https://github.com/vinsachi/FIPPCLE/blob/main/xling-bli/code/eval.py}}

For the simple Nearest Neighbor (NN) BLI with cosine (or dot product), we retrieve the word from the entire target language vocabulary of size $200k$ with the highest similarity score and mark it as the translation of the input/query word in the source language.

For the Cross-domain Similarity Local Scaling (CSLS) measure, a CSLS score is defined as $\text{CSLS}(\mathbf{x},\mathbf{y}) = 2m(\mathbf{x},\mathbf{y})-r_{\bm{X}}(\mathbf{y})-r_{\bm{Y}}(\mathbf{x}) $. $r_{\bm{X}}(y)$ is the average $m(\cdot, \cdot)$ score of $\mathbf{y}$ and its k-NNs ($k=10$) in $\bm{X}$;  $r_{\bm{Y}}(\mathbf{x})$ is the average $m(\cdot, \cdot)$ scores of $x$ and its k-NNs ($k=10$) in $\bm{Y}$. Note that when using CSLS scores to retrieve the translation of $\mathbf{x}$ in $\bm{Y}$, the term $r_{\bm{Y}}(\mathbf{x})$ can be ignored, as it is a constant for all $\mathbf{y}$, and we can similarly ignore $r_{\bm{X}}(\mathbf{y})$ when doing BLI in the opposite direction. 

\subsection{Generalised Procrustes in Stage C2}
\label{appendix:procrustes}
We consider the following Procrustes problem:

\begin{equation}
\begin{aligned}
\underset{\bm{W}}{\argmin} \left\|\bm{X}\bm{W}-\bm{Y}\right\|^{2}_{F}, \bm{W}\bm{W}^{T}=\bm{I},
\end{aligned}
\label{formula:a301}
\end{equation}
\noindent where $\bm{X}\in \mathbb{R}^{n\times d_{1}}$ is an embedding matrix and the set of its row vectors in the C1-induced cross-lingual space spans all source and target words in the training set $\mathcal{D}$ of size $n$, the row vectors of $\bm{Y}\in \mathbb{R}^{n\times d_{2}}$ are mBERT-encoded embeddings in the C2-induced space corresponding to the same words from $\bm{X}$, and $\bm{W}\in \mathbb{R}^{d_{1}\times d_{2}}$, $d_{1}\leq d_{2}$. A classical Orthogonal Procrustes Problem assumes that $d_{1}=d_{2}$ and $\bm{W}$ is an orthogonal matrix (i.e., it should be a square matrix), where its optimal solution is given by $\bm{U}\bm{V}^{T}$; here, $\bm{U}\bm{S}\bm{V}^{T}$ is the full singular value decomposition (SVD) of $\bm{X}^{T}\bm{Y}$. In our experiments, we need to address the case $d_{1}<d_{2}$ when mapping $300$-dimensional static fastText WEs to the $768$-dimensional space of mBERT-based WEs. It is easy to show that when $d_{1}<d_{2}$, $\bm{U}[\bm{S},\bm{0}]\bm{V}^{T}$$=$$\bm{X}^{T}\bm{Y}$ (again the full SVD decomposition), the optimal $\bm{W}$ is then $\bm{U}[\bm{I},\bm{0}]\bm{V}^{T}$ (it degrades to the Orthogonal Procrustes Problem when $d_{1}=d_{2}$). Below, we provide a simple proof.

Let $\bm{\Omega}=\bm{U}^{T}\bm{W}\bm{V}$, then $\bm{\Omega} \bm{\Omega}^{T} = \bm{I}$. Therefore, each of its element $-1\leq \bm{\Omega}_{i,j}\leq 1$.


\begin{equation}
\begin{aligned}
&\underset{\bm{W}}{\argmin} \left\|\bm{X}\bm{W}-\bm{Y}\right\|^{2}_{F}\\
= &\underset{\bm{W}}{\argmin} \langle \bm{X}\bm{W}-\bm{Y},\bm{X}\bm{W}-\bm{Y}\rangle_{F}\\
= &\underset{\bm{W}}{\argmin} \left\|\bm{X}\bm{W}\right\|^{2}_{F}+\left\|\bm{Y}\right\|^{2}_{F}-2\langle \bm{X}\bm{W},\bm{Y}\rangle_{F}\\
=&\underset{\bm{W}}{\argmax}\langle \bm{X}\bm{W},\bm{Y}\rangle_{F}\\
=&\underset{\bm{W}}{\argmax}\langle \bm{W},\bm{X}^{T}\bm{Y}\rangle_{F}\\
=&\underset{\bm{W}}{\argmax}\langle \bm{W},\bm{U}[\bm{S},\bm{0}]\bm{V}^{T}\rangle_{F}\\
=&\underset{\bm{W}}{\argmax}\langle [\bm{S},\bm{0}], \bm{U}^{T}\bm{W}\bm{V}\rangle_{F}\\
=&\underset{\bm{W}}{\argmax}\langle [\bm{S},\bm{0}], \bm{\Omega}\rangle_{F}\\
\end{aligned}
\label{formula:a302}
\end{equation}

In the formula above, $\left\|\cdot\right\|_{F}$ and $\langle \cdot,\cdot\rangle_{F}$ are Frobenius norm and Frobenius inner product, and we leverage their properties throughout the proof. Note that $S$ is a diagonal matrix with non-negative elements and thus the maximum is achieved when $\bm{\Omega}=[\bm{I},\bm{0}]$ and $\bm{W}=\bm{U}[\bm{I},\bm{0}]\bm{V}^{T}$.

Note that the Procrustes mapping over word embedding matrices keeps word similarities on both sides intact. Since $\bm{W}\bm{W}^{T}$$=$$\bm{I}$, $\cos(\mathbf{x}_{i}\bm{W}, \mathbf{x}_{j}\bm{W})=\cos(\mathbf{x}_{i}, \mathbf{x}_{j})$.

We would also like to add an additional note, although irrelevant to our own experiments, that the above derivation cannot address $d_{1}>d_{2}$ scenarios: in that case $\bm{W}\bm{W}^{T}$ cannot be a full-rank matrix and thus $\bm{W}\bm{W}^{T}\neq\bm{I}$.

\subsection{Languages in BLI Evaluation}

\begin{table}[ht!]
\begin{center}
\resizebox{0.4\textwidth}{!}{%
\begin{tabular}{llll}
\toprule 
\rowcolor{Gray}
\multicolumn{1}{c}{}  &\multicolumn{1}{c}{\bf Language}  &\multicolumn{1}{c}{\bf Family} &\multicolumn{1}{c}{\bf Code} 

\\  \cmidrule(lr){2-4}

\multirow{8}{*}{ \rotatebox{90}{\small XLING}}
 &\multicolumn{1}{c}{Croatian}  &\multicolumn{1}{c}{Slavic}  &\multicolumn{1}{c}{HR}  \\

 &\multicolumn{1}{c}{English}  &\multicolumn{1}{c}{Germanic}  &\multicolumn{1}{c}{EN}  \\
 
 &\multicolumn{1}{c}{Finnish}  &\multicolumn{1}{c}{Uralic}  &\multicolumn{1}{c}{FI}  \\ 

 &\multicolumn{1}{c}{French}  &\multicolumn{1}{c}{Romance}  &\multicolumn{1}{c}{FR}  \\
 
 &\multicolumn{1}{c}{German}  &\multicolumn{1}{c}{Germanic}  &\multicolumn{1}{c}{DE}  \\
  
 &\multicolumn{1}{c}{Italian}  &\multicolumn{1}{c}{Romance}  &\multicolumn{1}{c}{IT}  \\ 

 &\multicolumn{1}{c}{Russian}  &\multicolumn{1}{c}{Slavic}  &\multicolumn{1}{c}{RU}  \\
 
 &\multicolumn{1}{c}{Turkish}  &\multicolumn{1}{c}{Turkic}  &\multicolumn{1}{c}{TR}  \\
 
 \cmidrule(lr){2-4}
 
\multirow{6}{*}{ \rotatebox{90}{\small PanLex-BLI \ \ }}
 &\multicolumn{1}{c}{Basque}  &\multicolumn{1}{c}{–(isolate)}  &\multicolumn{1}{c}{EU}  \\

 &\multicolumn{1}{c}{Bulgarian}  &\multicolumn{1}{c}{Slavic}  &\multicolumn{1}{c}{BG}  \\

 &\multicolumn{1}{c}{Catalan}  &\multicolumn{1}{c}{Romance}  &\multicolumn{1}{c}{CA}  \\
 
 &\multicolumn{1}{c}{Estonian}  &\multicolumn{1}{c}{Uralic}  &\multicolumn{1}{c}{ET}  \\ 

 &\multicolumn{1}{c}{Hebrew}  &\multicolumn{1}{c}{Afro-Asiatic}  &\multicolumn{1}{c}{HE}  \\
  
 &\multicolumn{1}{c}{Hungarian}  &\multicolumn{1}{c}{Uralic}  &\multicolumn{1}{c}{HU}  \\ 
 
\bottomrule
\end{tabular}
}

\caption{A list of languages in our experiments along with their language family and ISO 639-1 code.}
\label{table:languages}
\end{center}
\end{table}

\section{Reproducibility Checklist}
\begin{itemize}
    \item \textbf{BLI Data}: The two BLI datasets are publicly available.\footnote{\url{https://github.com/codogogo/xling-eval}} \footnote{\url{https://github.com/cambridgeltl/panlex-bli}}
    \item \textbf{Static WEs}: We use the preprocessed fastText WEs provided by \newcite{glavas-etal-2019-properly}. For PanLex-BLI, we follow the original paper's setup \cite{vulic-etal-2019-really} and adopt fastText WEs pretrained on both Common Crawl and Wikipedia \cite{bojanowski2017enriching}.\footnote{\url{https://fasttext.cc/docs/en/crawl-vectors}}
    Following prior work, all static WEs are trimmed to contain vectors for the top $200k$ most frequent words in each language. 
    \item \textbf{Pretrained LMs}: The model variants used in our work are `bert-base-multilingual-uncased' for mBERT, `xlm-mlm-100-1280' for XLM and `google/mt5-small' for mT5, all retrieved from the \rurl{huggingface.co} model repository.    
    \item \textbf{Baseline BLI Models}: All models are accessible online as publicly available github repositories.
    \item \textbf{Source Code}: Our code is available online at \url{https://github.com/cambridgeltl/ContrastiveBLI}.
    \item \textbf{Computing Infrastructure}: We run our main experiments on a machine with a 4.00GHz 4-core i7-6700K CPU, 64GB RAM and two 12GB NVIDIA TITAN X GPUs. For the experiments with XLM and mT5 only, we leverage a cluster where we have access to two 24GB RTX 3090 GPUs.
    \item \textbf{Software}: We rely on Python 3.6.10, PyTorch 1.7.0 and \rurl{huggingface.co} Transformers 4.4.2. Automatic Mixed Precision (AMP)\footnote{\url{https://pytorch.org/docs/stable/amp}} is leveraged during Stage C2 training. 
    
    \item \textbf{Runtime}: The training process (excluding data loading and evaluation) typically takes 650 seconds for Stage C1 (seed dictionary of $5k$ pairs, $2$ self-learning iterations) and $200$ seconds for C1 ($1k$ pairs, $3$ self-learning iterations) on a single GPU. Stage C2 runs for $\approx$ $500$ seconds on two GPUs (TITAN X). 
    \item \textbf{Robustness and Randomness}: Our improvement is robust since both C1 and C2 outperform existing SotA methods in $112$ BLI setups by a considerable margin. We regard our C1 as a deterministic algorithm because we adopt 0 dropout and a batch size equal to the size of the whole training dictionary (no randomness from shuffling). In C2, considering its robustness, we fix the random seed to $33$ over all runs and setups.       
\end{itemize}


\section{Visualisation of mBERT-Based Word Representations}
\label{appendix:tsne}
To illustrate the impact of the proposed BLI-oriented fine-tuning of mBERT in Stage C2 on its representation space, we visualise the $768$-dimensional mBERT word representations (i.e., mBERT-encoded word features alone, without the infusion of C1-aligned static WEs). We encode BLI test sets (i.e., these sets include $2k$ source-target word pairs unseen during C2 fine-tuning), before and after fine-tuning, relying on $1k$ training samples as the seed dictionary $\mathcal{D}_0$. 

Here, we provide comparative t-SNE visualisations between source and target word mBERT-based decontextualised word representations (see \S\ref{s:c2}) for six language pairs from the BLI dataset of \newcite{glavas-etal-2019-properly}: EN-IT, FI-RU, EN-HR, HR-RU, DE-TR, and IT-FR, while two additional visualisations are available in the main paper (for RU-IT and TR-HR, see Figure~\ref{fig:tsne-main} in \S\ref{s:further}). As visible in Figures~\ref{fig:tsneenit} to \ref{fig:tsneitfr} below, before BLI-oriented fine-tuning in Stage C2, there is an obvious separation between mBERT's representation subspaces in the two languages. This undesired property gets mitigated, to a considerable extent, by the fine-tuning procedure in Stage C2. 

\section{Visualisation of fastText-based Word Representations}
\label{appendix:tsne_fastText}

\begin{figure}[!ht]
\hspace*{-0.45cm}           
\centering
\includegraphics[width=6.8cm]{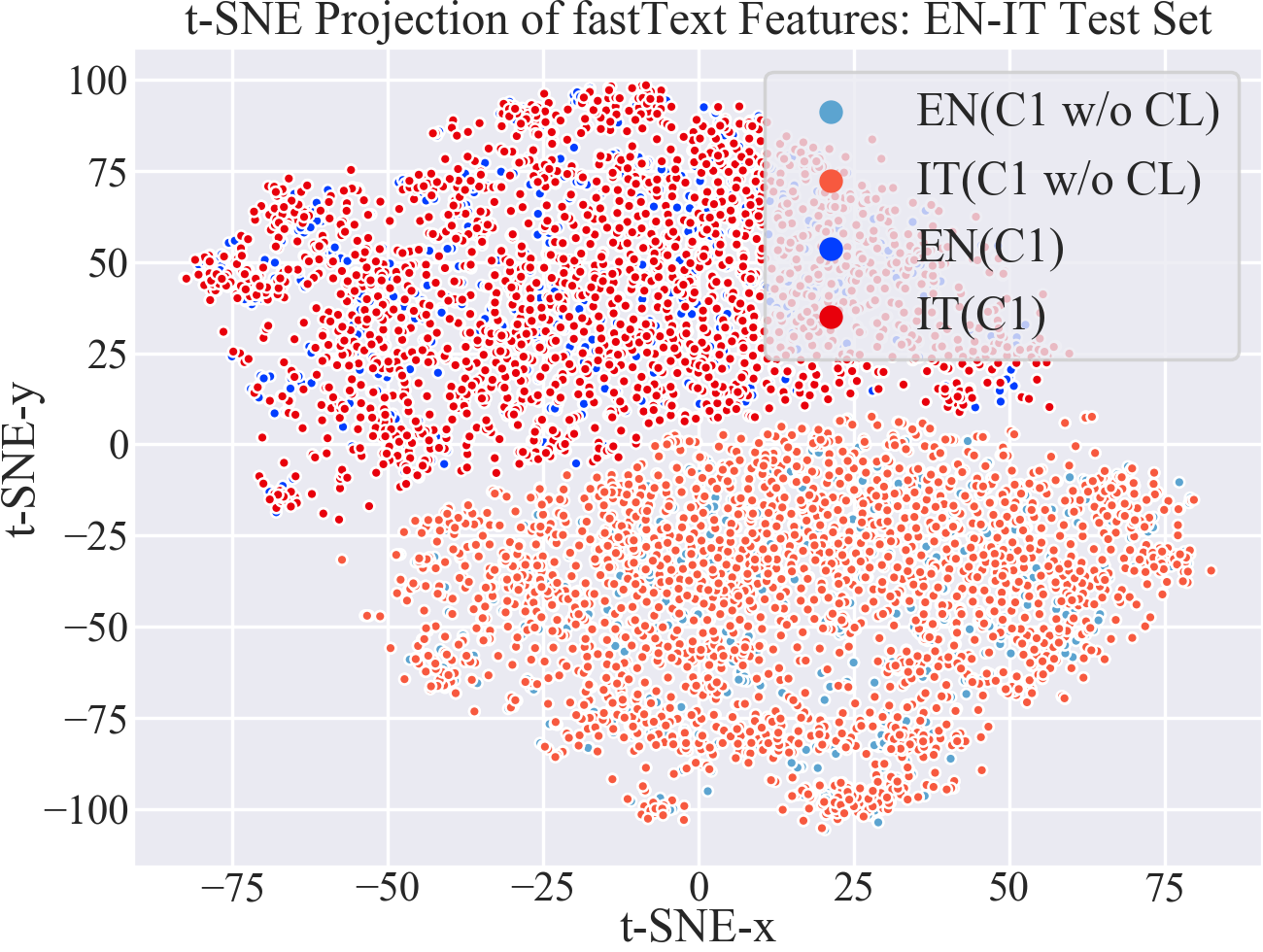}
\caption{A t-SNE visualisation of mapped fastText WEs of words from the EN-IT BLI test set. The representations derived from C1 w/o CL are plotted in muted blue and red, and the whole C1 alignment in bright colours.}
\label{fig:tsneenit_ft}
\end{figure}

\begin{figure}[!ht]
\hspace*{-0.45cm}           
\centering
\includegraphics[width=6.8cm]{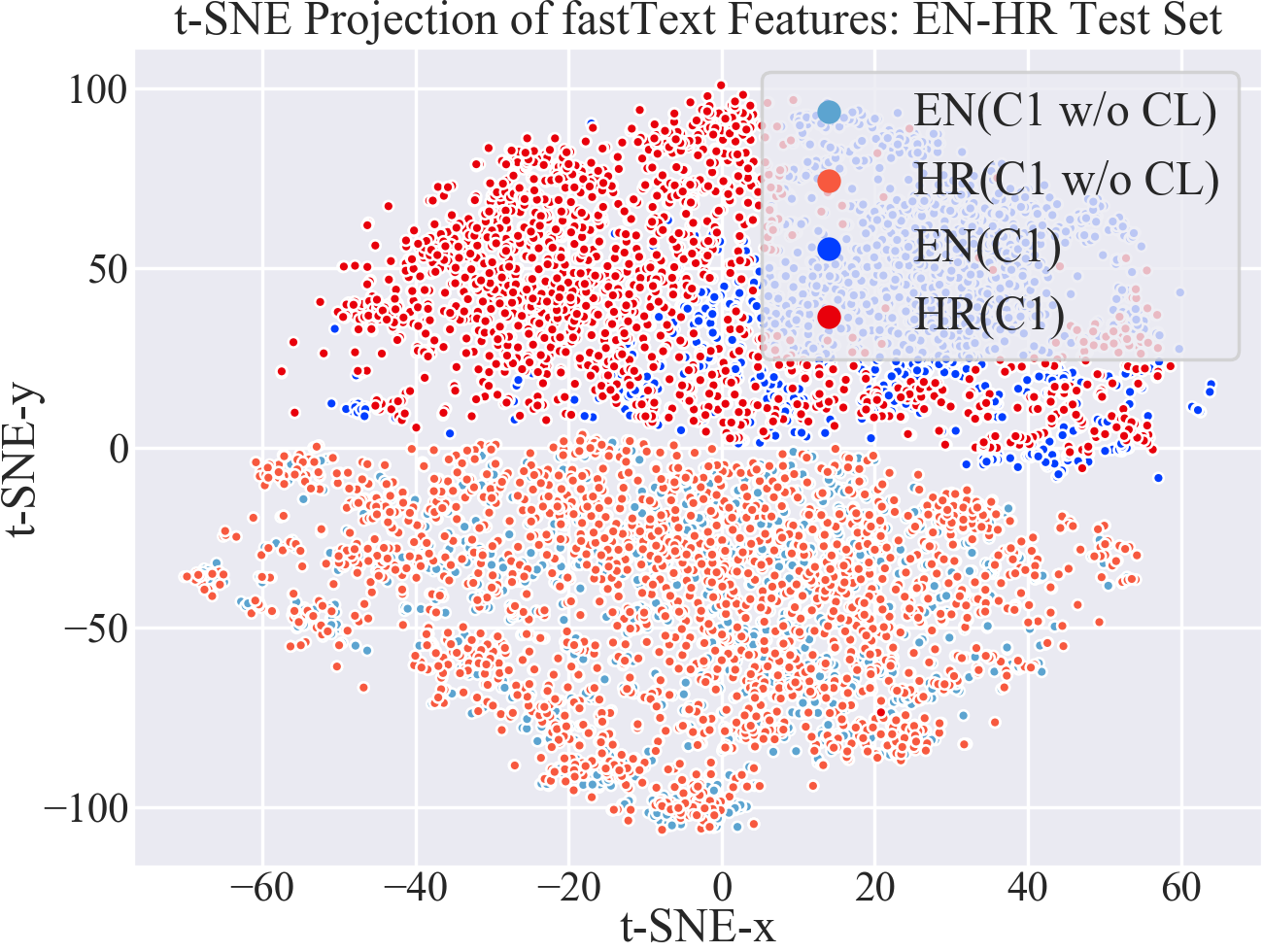}
\caption{A t-SNE visualisation of mapped fastText WEs of words from the EN-HR BLI test set. The representations derived from C1 w/o CL are plotted in muted blue and red, and the whole C1 alignment in bright colours.}
\label{fig:tsneenhr_ft}
\end{figure}

\begin{figure}[!ht]
\hspace*{-0.45cm}           
\centering
\includegraphics[width=6.8cm]{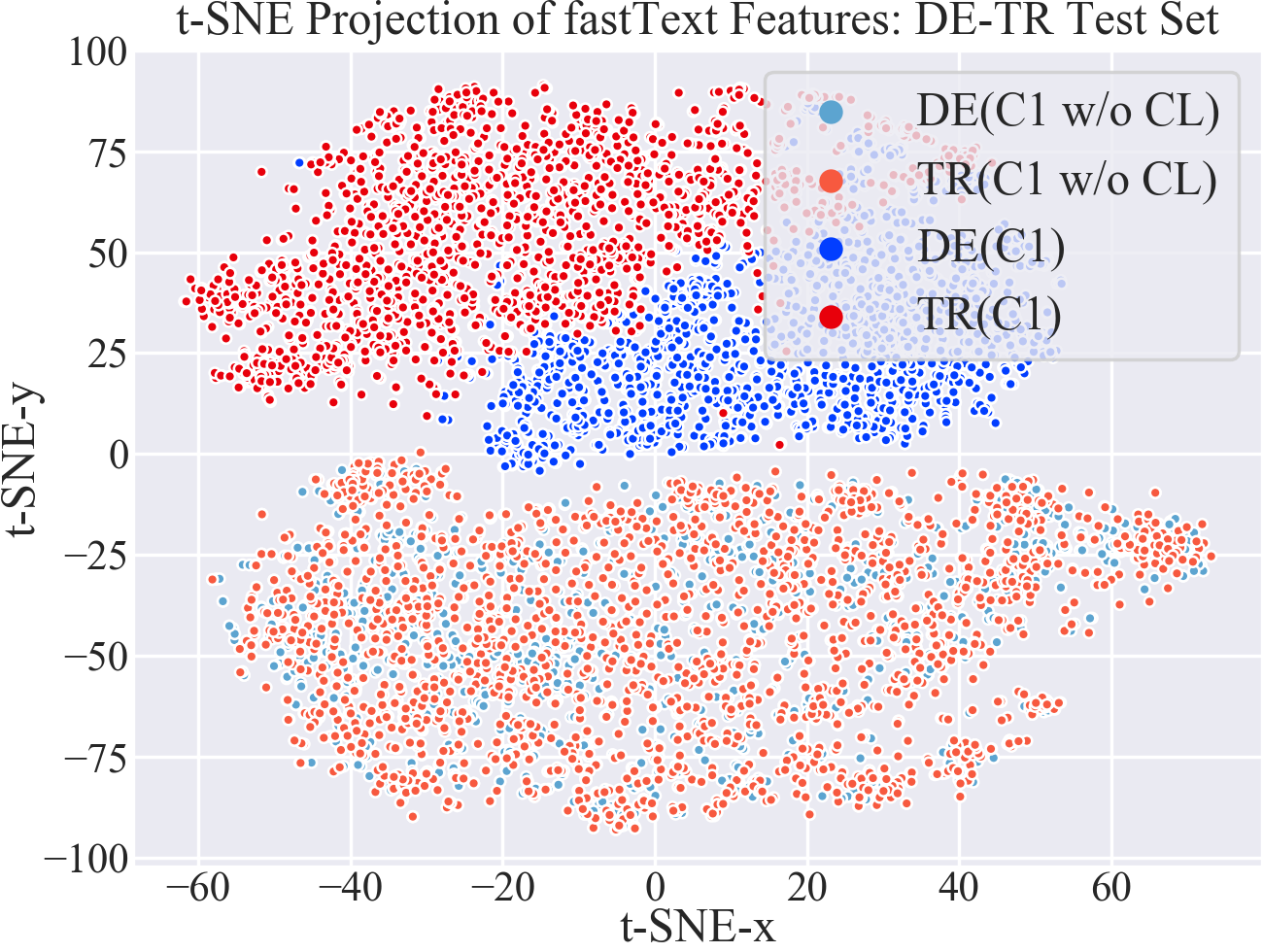}
\caption{A t-SNE visualisation of mapped fastText WEs of words from the DE-TR BLI test set. The representations derived from C1 w/o CL are plotted in muted blue and red, and the whole C1 alignment in bright colours.}
\label{fig:tsnedetr_ft}
\end{figure}

\begin{figure}[!ht]
\hspace*{-0.45cm}           
\centering
\includegraphics[width=6.8cm]{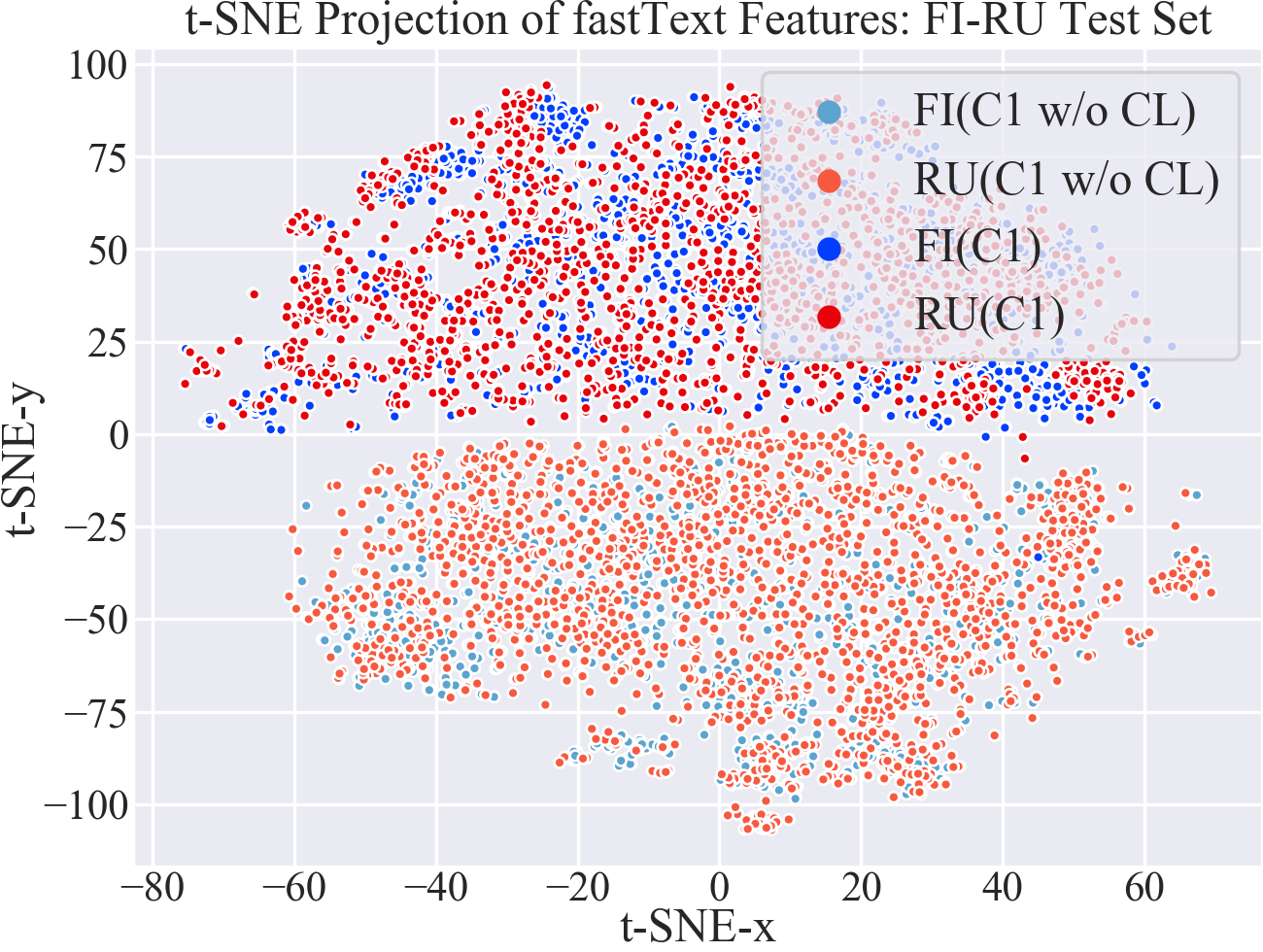}
\caption{A t-SNE visualisation of mapped fastText WEs of words from the FI-RU BLI test set. The representations derived from C1 w/o CL are plotted in muted blue and red, and the whole C1 alignment in bright colours.}
\label{fig:tsnefiru_ft}
\end{figure}

\begin{figure}[!ht]
\hspace*{-0.45cm}           
\centering
\includegraphics[width=6.8cm]{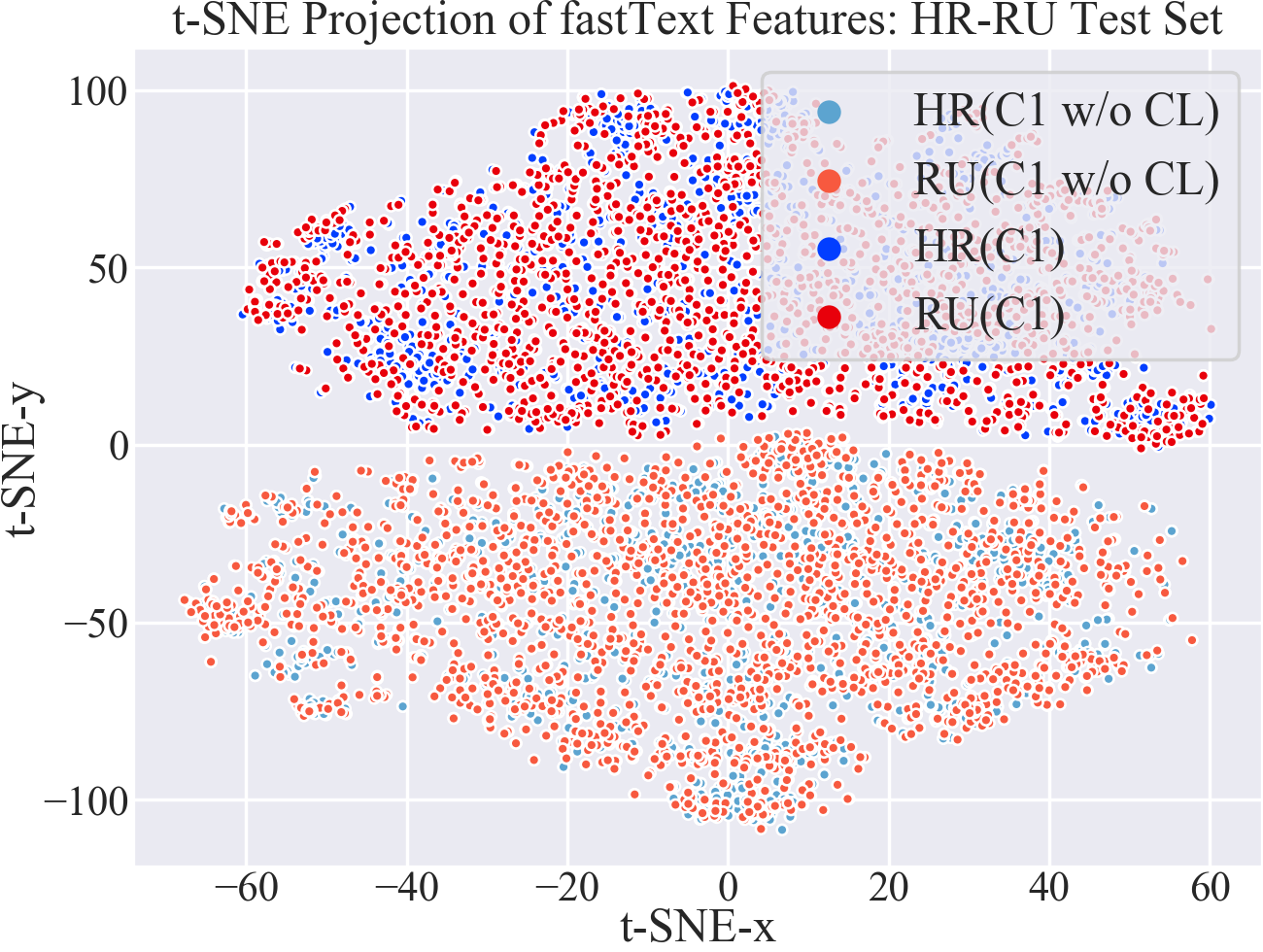}
\caption{A t-SNE visualisation of mapped fastText WEs of words from the HR-RU BLI test set. The representations derived from C1 w/o CL are plotted in muted blue and red, and the whole C1 alignment in bright colours.}
\label{fig:tsnehrru_ft}
\end{figure}

\begin{figure}[!ht]
\hspace*{-0.45cm}           
\centering
\includegraphics[width=6.8cm]{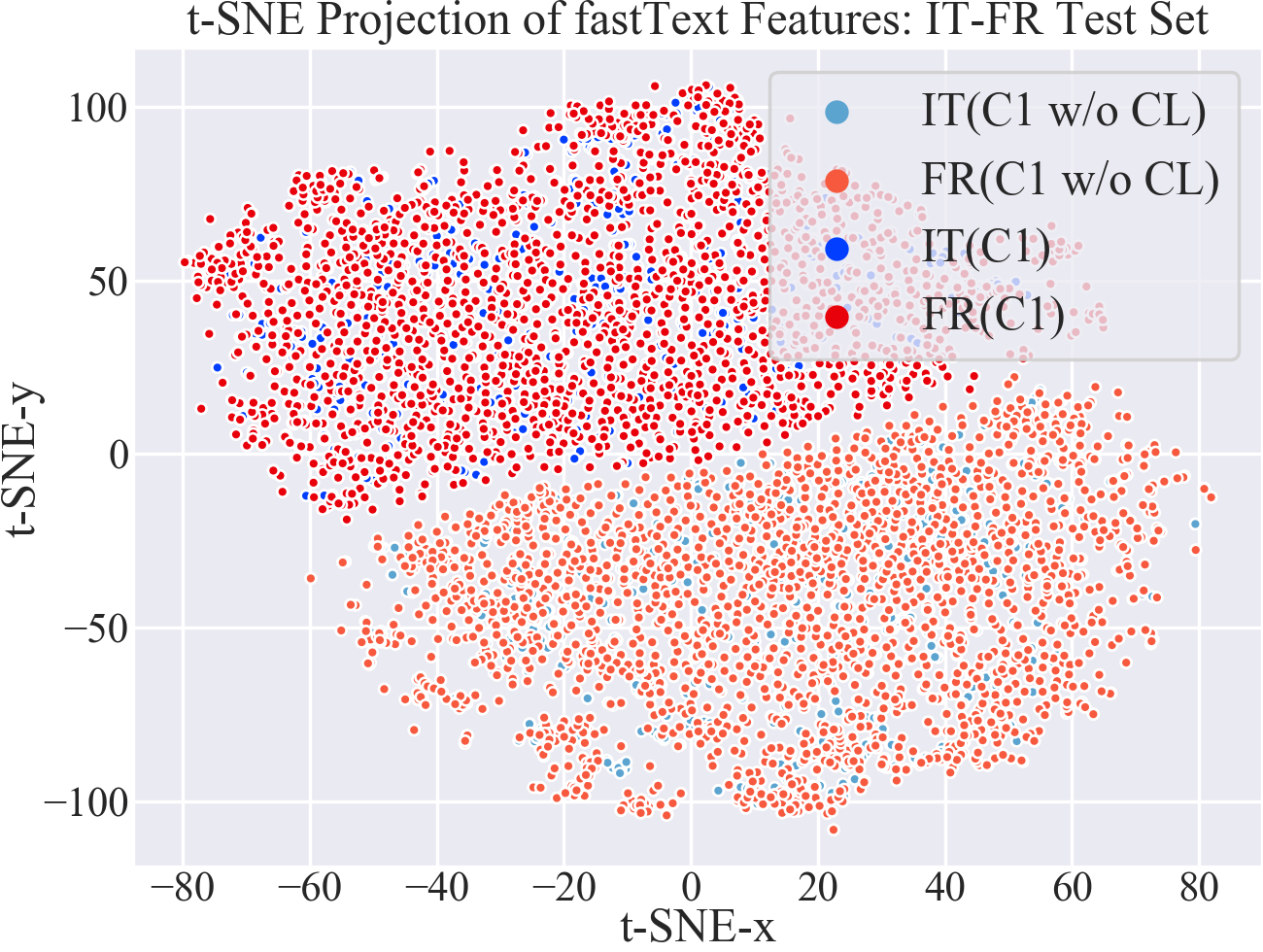}
\caption{A t-SNE visualisation of mapped fastText WEs of words from the IT-FR BLI test set. The representations derived from C1 w/o CL are plotted in muted blue and red, and the whole C1 alignment in bright colours.}
\label{fig:tsneitfr_ft}
\end{figure}

To show the impact of contrastive tuning in Stage C1, in Figures~\ref{fig:tsneenit_ft} to \ref{fig:tsneitfr_ft} we provide t-SNE plots of $300$-dimensional C1-aligned fastText embeddings with and without contrastive tuning (see \S\ref{s:c1}) respectively for the same six language pairs as in Appendix~\ref{appendix:tsne}. The C1 w/o CL alignment consists of advanced mapping and self-learning loops, which has already been discussed in our ablation study (see \S\ref{s:further}). Like in Appendix~\ref{appendix:tsne}, the linear maps are learned on $1k$ seed translation pairs and our plots only cover the BLI test sets.

\begin{figure}[!p]
\hspace*{-0.45cm}           
\centering
\includegraphics[width=6.8cm]{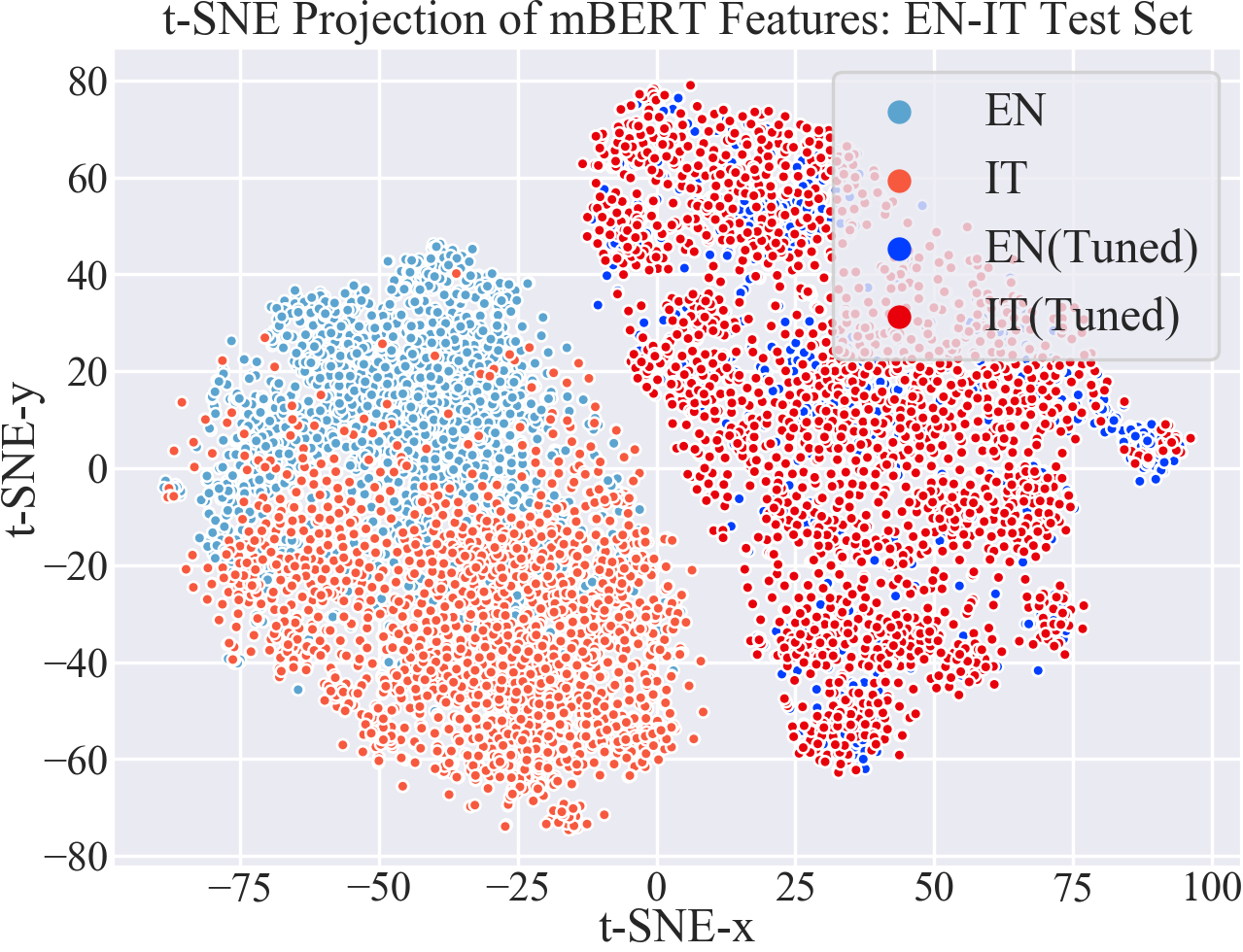}
\caption{A t-SNE visualisation of mBERT-encoded representations of words from the EN-IT BLI test set. The representations before BLI-oriented fine-tuning of mBERT in Stage C2 are plotted in muted blue and red, and after fine-tuning in bright colours.}
\label{fig:tsneenit}
\end{figure}

\begin{figure}[!ht]
\hspace*{-0.45cm}     
\centering
\includegraphics[width=6.8cm]{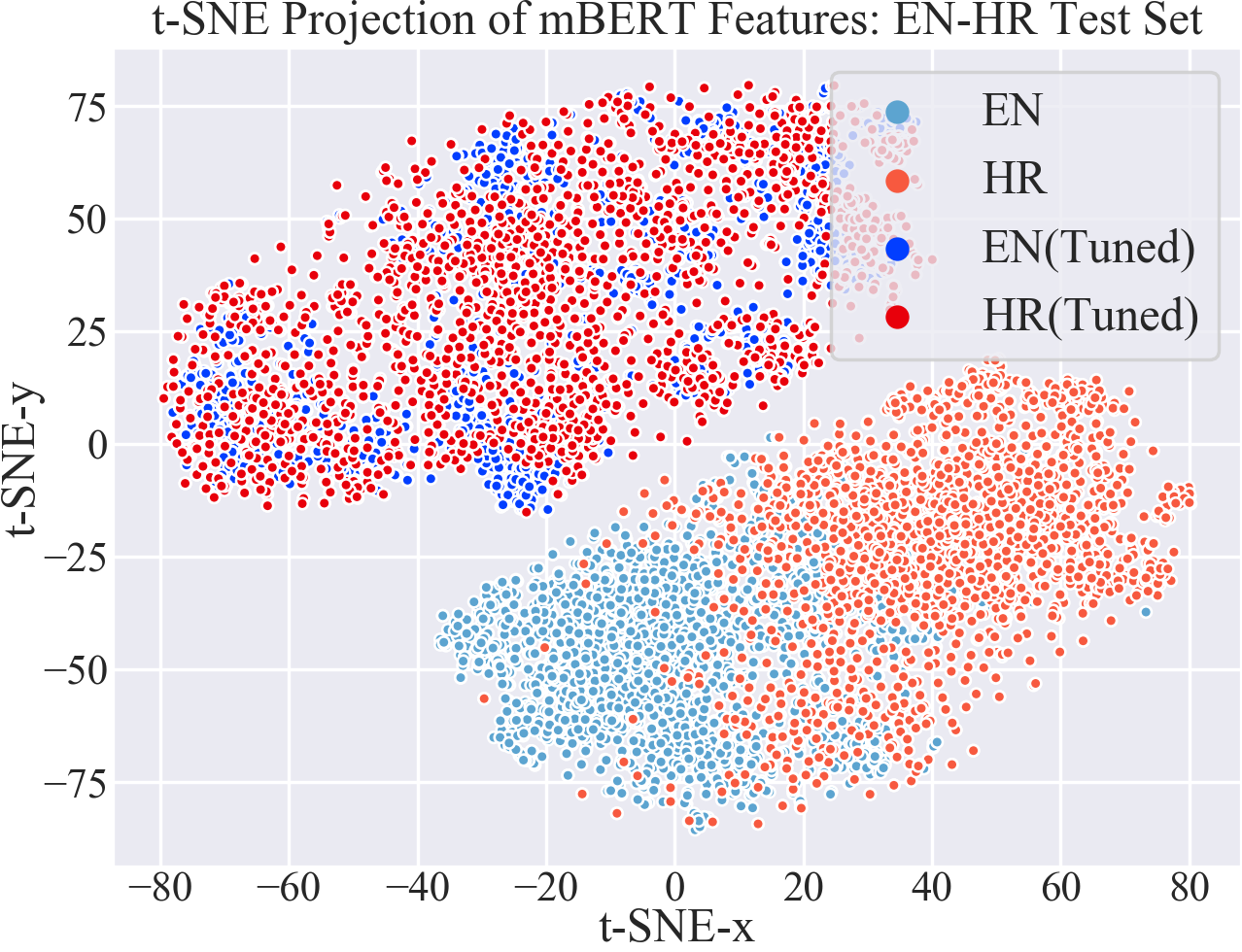}
\caption{A t-SNE visualisation of mBERT-encoded representations of words from the EN-HR BLI test set. The representations before BLI-oriented fine-tuning of mBERT in Stage C2 are plotted in muted blue and red, and after fine-tuning in bright colours.}
\label{fig:tsneenhr}
\end{figure}

\begin{figure}[!ht]
\hspace*{-0.45cm}     
\centering
\includegraphics[width=6.8cm]{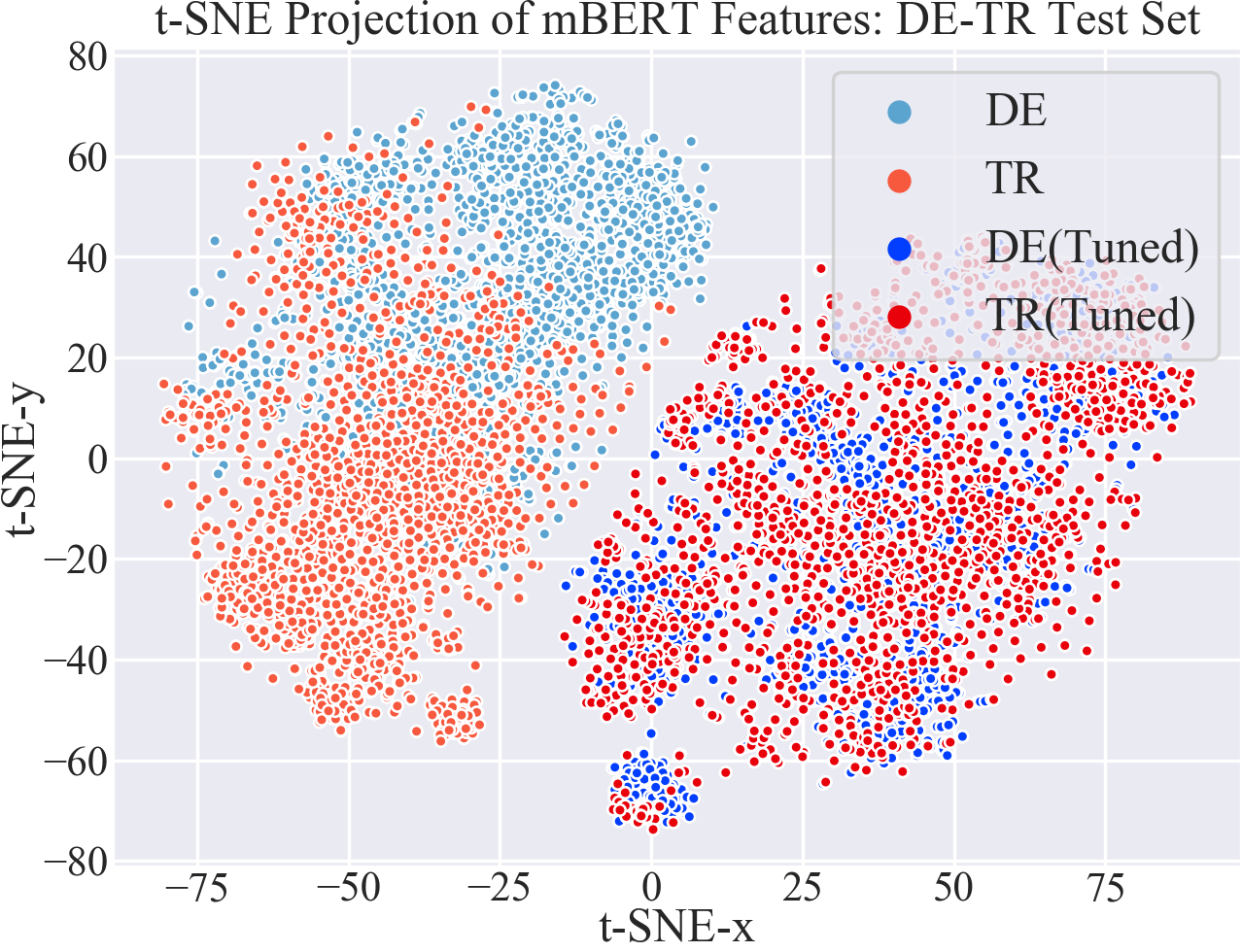}
\caption{A t-SNE visualisation of mBERT-encoded representations of words from the DE-TR BLI test set. The representations before BLI-oriented fine-tuning of mBERT in Stage C2 are plotted in muted blue and red, and after fine-tuning in bright colours.}
\label{fig:tsnedetr}
\end{figure}

\begin{figure}[!ht]
\hspace*{-0.45cm}     
\centering
\includegraphics[width=6.8cm]{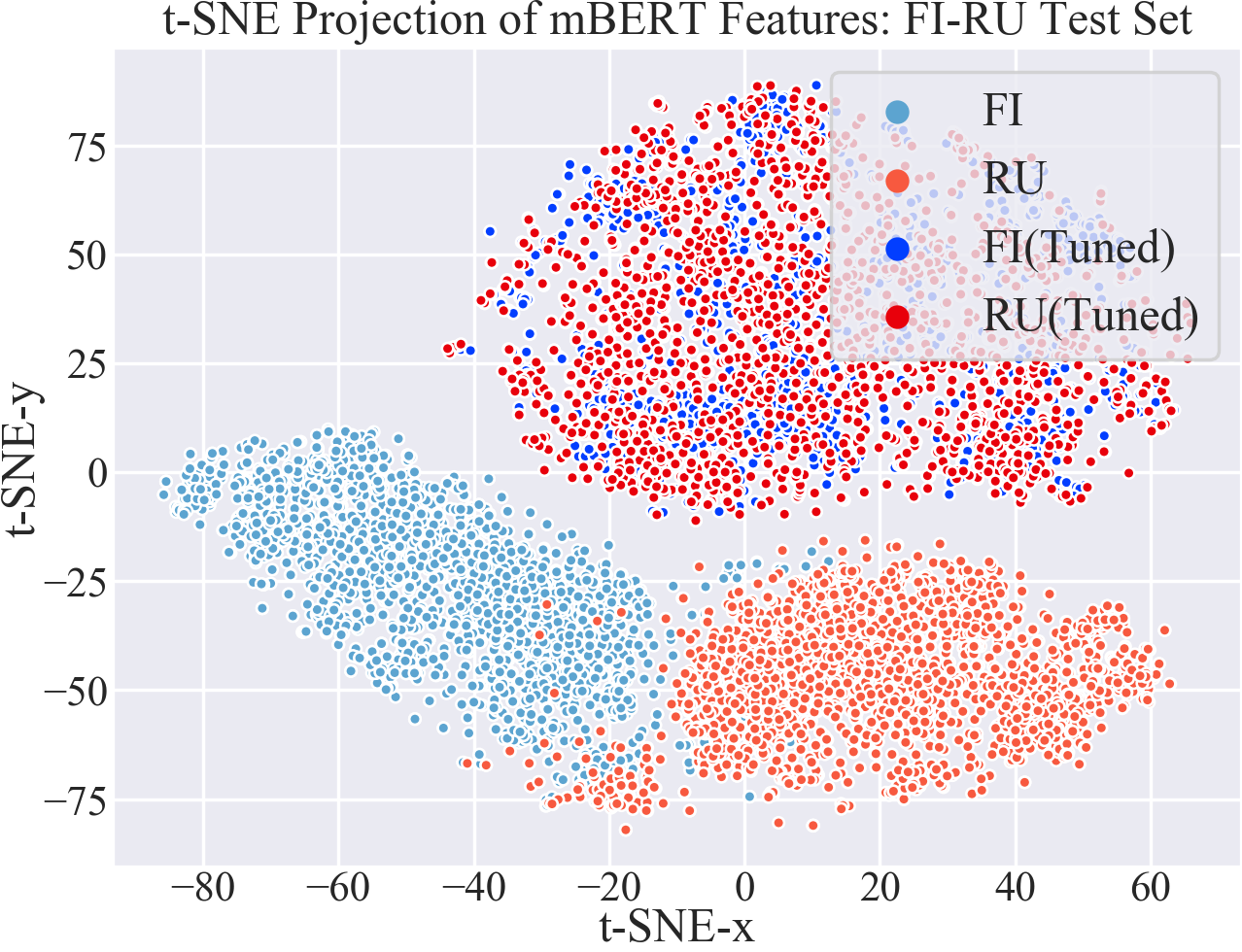}
\caption{A t-SNE visualisation of mBERT-encoded representations of words from the FI-RU BLI test set. The representations before BLI-oriented fine-tuning of mBERT in Stage C2 are plotted in muted blue and red, and after fine-tuning in bright colours.}
\label{fig:tsnefiru}
\end{figure}

\begin{figure}[!ht]
\hspace*{-0.45cm}     
\centering
\includegraphics[width=6.8cm]{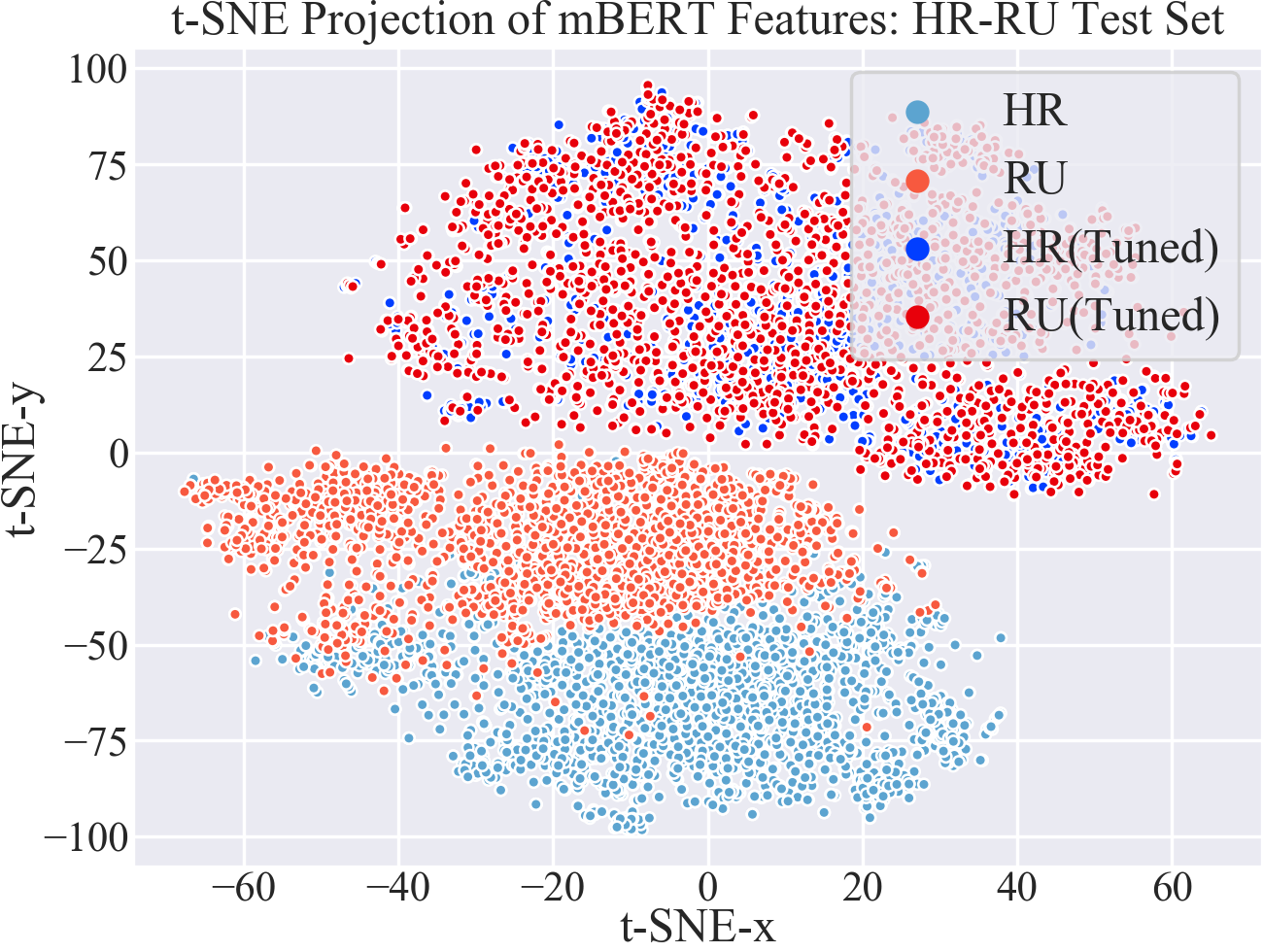}
\caption{A t-SNE visualisation of mBERT-encoded representations of words from the HR-RU BLI test set. The representations before BLI-oriented fine-tuning of mBERT in Stage C2 are plotted in muted blue and red, and after fine-tuning in bright colours.}
\label{fig:tsnehrru}
\end{figure}

\begin{figure}[!ht]
\hspace*{-0.45cm}     
\centering
\includegraphics[width=6.8cm]{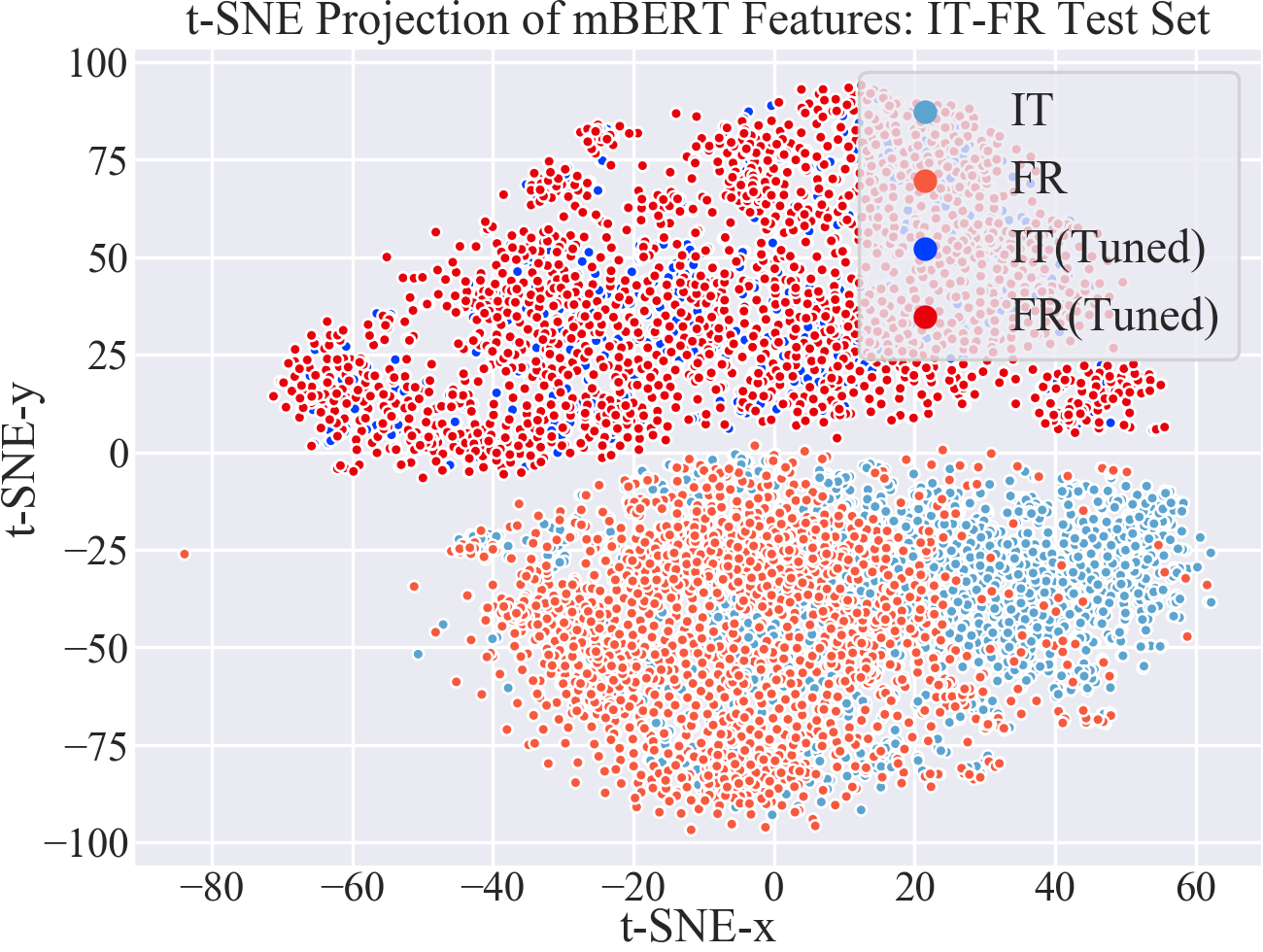}
\caption{A t-SNE visualisation of mBERT-encoded representations of words from the IT-FR BLI test set. The representations before BLI-oriented fine-tuning of mBERT in Stage C2 are plotted in muted blue and red, and after fine-tuning in bright colours.}
\label{fig:tsneitfr}
\end{figure}

\section{Full BLI Results}
Complete results on the BLI dataset of \newcite{glavas-etal-2019-properly}, per each language pair and also including NN-based BLI scores, are provided in Tables~\ref{table:appendix-full} and \ref{table:appendix2}. It can be seen as an expanded variant of the main Table~\ref{table:main} presented in the main paper.

\label{appendix:full}

\begin{table*}[ht]
\begin{center}
\resizebox{0.79\textwidth}{!}{%
\begin{tabular}{lllllll}
\toprule 

\rowcolor{Gray}
\multicolumn{1}{c}{[5k] \bf Pairs}  &\multicolumn{1}{c}{\bf RCSLS}  &\multicolumn{1}{c}{\bf VecMap-Sup} &\multicolumn{1}{c}{\bf LNMap} &\multicolumn{1}{c}{\bf FIPP}
&\multicolumn{1}{c}{\bf C1}
&\multicolumn{1}{c}{\bf C2 (C1)}
\\ \cmidrule(lr){2-5} \cmidrule(lr){6-7}

\multicolumn{1}{c}{DE$\to$FI}  &\multicolumn{1}{c}{30.62/37.35}  &\multicolumn{1}{c}{29.21/33.59}  &\multicolumn{1}{c}{31.35/36.10}  &\multicolumn{1}{c}{30.93/35.37}  &\multicolumn{1}{c}{\underline{38.97}/\underline{42.10}}  &\multicolumn{1}{c}{\textbf{41.47}/\textbf{44.65}}\\

\multicolumn{1}{c}{FI$\to$DE}  &\multicolumn{1}{c}{32.48/39.36}  &\multicolumn{1}{c}{35.42/38.73}  &\multicolumn{1}{c}{31.32/36.73}  &\multicolumn{1}{c}{36.05/39.41}  &\multicolumn{1}{c}{\underline{39.83}/\underline{42.46}}  &\multicolumn{1}{c}{\textbf{44.30}/\textbf{47.03}}\\

\multicolumn{1}{c}{DE$\to$FR}  &\multicolumn{1}{c}{47.63/52.74}  &\multicolumn{1}{c}{46.64/50.44}  &\multicolumn{1}{c}{44.91/48.46}  &\multicolumn{1}{c}{47.89/50.44}  &\multicolumn{1}{c}{\underline{51.49}/\underline{53.78}}  &\multicolumn{1}{c}{\textbf{54.09}/\textbf{55.56}}\\

\multicolumn{1}{c}{FR$\to$DE}  &\multicolumn{1}{c}{47.23/51.22}  &\multicolumn{1}{c}{45.37/47.75}  &\multicolumn{1}{c}{41.65/44.80}  &\multicolumn{1}{c}{45.73/47.85}  &\multicolumn{1}{c}{\underline{50.13}/\underline{51.37}}  &\multicolumn{1}{c}{\textbf{53.23}/\textbf{53.29}}\\

\multicolumn{1}{c}{DE$\to$HR}  &\multicolumn{1}{c}{29.26/33.75}  &\multicolumn{1}{c}{27.07/32.08}  &\multicolumn{1}{c}{27.65/32.34}  &\multicolumn{1}{c}{27.65/31.09}  &\multicolumn{1}{c}{\underline{34.17}/\underline{37.66}}  &\multicolumn{1}{c}{\textbf{39.07}/\textbf{42.41}}\\

\multicolumn{1}{c}{HR$\to$DE}  &\multicolumn{1}{c}{30.30/36.35}  &\multicolumn{1}{c}{32.98/37.24}  &\multicolumn{1}{c}{28.98/33.72}  &\multicolumn{1}{c}{31.51/34.30}  &\multicolumn{1}{c}{\underline{39.14}/\underline{41.35}}  &\multicolumn{1}{c}{\textbf{45.03}/\textbf{48.29}}\\

\multicolumn{1}{c}{DE$\to$IT}  &\multicolumn{1}{c}{47.68/52.63}  &\multicolumn{1}{c}{47.78/50.55}  &\multicolumn{1}{c}{44.91/47.94}  &\multicolumn{1}{c}{46.90/49.97}  &\multicolumn{1}{c}{\underline{50.65}/\underline{52.79}}  &\multicolumn{1}{c}{\textbf{52.48}/\textbf{54.77}}\\

\multicolumn{1}{c}{IT$\to$DE}  &\multicolumn{1}{c}{46.51/51.01}  &\multicolumn{1}{c}{44.96/47.29}  &\multicolumn{1}{c}{42.58/45.53}  &\multicolumn{1}{c}{44.86/46.67}  &\multicolumn{1}{c}{\underline{49.97}/\underline{51.21}}  &\multicolumn{1}{c}{\textbf{53.90}/\textbf{53.80}}\\

\multicolumn{1}{c}{DE$\to$RU}  &\multicolumn{1}{c}{37.87/42.41}  &\multicolumn{1}{c}{31.98/34.38}  &\multicolumn{1}{c}{35.21/37.92}  &\multicolumn{1}{c}{36.57/37.09}  &\multicolumn{1}{c}{\underline{42.67}/\underline{44.29}}  &\multicolumn{1}{c}{\textbf{44.71}/\textbf{46.79}}\\

\multicolumn{1}{c}{RU$\to$DE}  &\multicolumn{1}{c}{40.54/45.78}  &\multicolumn{1}{c}{40.65/43.32}  &\multicolumn{1}{c}{36.72/40.28}  &\multicolumn{1}{c}{40.18/42.38}  &\multicolumn{1}{c}{\underline{46.05}/\underline{46.73}}  &\multicolumn{1}{c}{\textbf{48.51}/\textbf{49.71}}\\

\multicolumn{1}{c}{DE$\to$TR}  &\multicolumn{1}{c}{24.93/30.99}  &\multicolumn{1}{c}{23.84/27.18}  &\multicolumn{1}{c}{25.46/29.16}  &\multicolumn{1}{c}{23.94/27.65}  &\multicolumn{1}{c}{\underline{31.30}/\underline{34.69}}  &\multicolumn{1}{c}{\textbf{35.84}/\textbf{38.86}}\\

\multicolumn{1}{c}{TR$\to$DE}  &\multicolumn{1}{c}{27.00/31.84}  &\multicolumn{1}{c}{26.46/29.93}  &\multicolumn{1}{c}{24.92/27.85}  &\multicolumn{1}{c}{26.09/29.18}  &\multicolumn{1}{c}{\underline{33.33}/\underline{36.74}}  &\multicolumn{1}{c}{\textbf{38.50}/\textbf{40.95}}\\

\multicolumn{1}{c}{EN$\to$DE}  &\multicolumn{1}{c}{52.95/\underline{57.60}}  &\multicolumn{1}{c}{48.65/51.00}  &\multicolumn{1}{c}{45.80/47.95}  &\multicolumn{1}{c}{50.25/51.85}  &\multicolumn{1}{c}{\underline{55.50}/54.90}  &\multicolumn{1}{c}{\textbf{59.25}/\textbf{57.75}}\\

\multicolumn{1}{c}{DE$\to$EN}  &\multicolumn{1}{c}{50.97/56.55}  &\multicolumn{1}{c}{52.01/55.24}  &\multicolumn{1}{c}{46.48/50.50}  &\multicolumn{1}{c}{52.16/55.03}  &\multicolumn{1}{c}{\underline{54.77}/\underline{57.69}}  &\multicolumn{1}{c}{\textbf{56.03}/\textbf{58.95}}\\

\multicolumn{1}{c}{EN$\to$FI}  &\multicolumn{1}{c}{35.40/42.05}  &\multicolumn{1}{c}{35.25/37.75}  &\multicolumn{1}{c}{34.45/38.35}  &\multicolumn{1}{c}{34.55/39.10}  &\multicolumn{1}{c}{\underline{40.70}/\underline{44.60}}  &\multicolumn{1}{c}{\textbf{45.45}/\textbf{47.15}}\\

\multicolumn{1}{c}{FI$\to$EN}  &\multicolumn{1}{c}{34.21/41.25}  &\multicolumn{1}{c}{39.04/43.51}  &\multicolumn{1}{c}{31.69/36.26}  &\multicolumn{1}{c}{36.42/40.51}  &\multicolumn{1}{c}{\underline{41.46}/\underline{46.30}}  &\multicolumn{1}{c}{\textbf{44.82}/\textbf{50.55}}\\

\multicolumn{1}{c}{EN$\to$FR}  &\multicolumn{1}{c}{61.65/\underline{66.55}}  &\multicolumn{1}{c}{60.65/63.10}  &\multicolumn{1}{c}{57.75/62.10}  &\multicolumn{1}{c}{61.15/63.25}  &\multicolumn{1}{c}{\underline{64.35}/65.05}  &\multicolumn{1}{c}{\textbf{68.45}/\textbf{67.20}}\\

\multicolumn{1}{c}{FR$\to$EN}  &\multicolumn{1}{c}{59.23/63.11}  &\multicolumn{1}{c}{59.60/62.75}  &\multicolumn{1}{c}{54.53/58.72}  &\multicolumn{1}{c}{59.03/61.87}  &\multicolumn{1}{c}{\underline{62.23}/\underline{63.84}}  &\multicolumn{1}{c}{\textbf{64.30}/\textbf{65.49}}\\

\multicolumn{1}{c}{EN$\to$HR}  &\multicolumn{1}{c}{31.40/37.90}  &\multicolumn{1}{c}{29.70/34.05}  &\multicolumn{1}{c}{28.40/31.75}  &\multicolumn{1}{c}{28.50/31.95}  &\multicolumn{1}{c}{\underline{37.50}/\underline{40.70}}  &\multicolumn{1}{c}{\textbf{43.60}/\textbf{47.20}}\\

\multicolumn{1}{c}{HR$\to$EN}  &\multicolumn{1}{c}{28.51/35.67}  &\multicolumn{1}{c}{35.24/39.08}  &\multicolumn{1}{c}{27.83/32.61}  &\multicolumn{1}{c}{31.93/34.72}  &\multicolumn{1}{c}{\underline{38.66}/\underline{42.40}}  &\multicolumn{1}{c}{\textbf{42.61}/\textbf{49.08}}\\

\multicolumn{1}{c}{EN$\to$IT}  &\multicolumn{1}{c}{58.85/\underline{64.05}}  &\multicolumn{1}{c}{57.20/60.40}  &\multicolumn{1}{c}{55.30/59.05}  &\multicolumn{1}{c}{56.95/59.75}  &\multicolumn{1}{c}{\underline{61.55}/63.45}  &\multicolumn{1}{c}{\textbf{65.30}/\textbf{65.60}}\\

\multicolumn{1}{c}{IT$\to$EN}  &\multicolumn{1}{c}{55.09/61.50}  &\multicolumn{1}{c}{57.73/62.17}  &\multicolumn{1}{c}{52.09/56.02}  &\multicolumn{1}{c}{56.69/60.52}  &\multicolumn{1}{c}{\underline{59.90}/\underline{63.51}}  &\multicolumn{1}{c}{\textbf{62.27}/\textbf{65.27}}\\

\multicolumn{1}{c}{EN$\to$RU}  &\multicolumn{1}{c}{44.75/\underline{49.40}}  &\multicolumn{1}{c}{38.00/39.65}  &\multicolumn{1}{c}{38.90/41.10}  &\multicolumn{1}{c}{40.70/42.00}  &\multicolumn{1}{c}{\underline{48.05}/49.15}  &\multicolumn{1}{c}{\textbf{50.85}/\textbf{50.50}}\\

\multicolumn{1}{c}{RU$\to$EN}  &\multicolumn{1}{c}{42.80/48.66}  &\multicolumn{1}{c}{45.78/49.35}  &\multicolumn{1}{c}{37.51/42.64}  &\multicolumn{1}{c}{43.27/47.15}  &\multicolumn{1}{c}{\underline{48.45}/\underline{51.91}}  &\multicolumn{1}{c}{\textbf{49.24}/\textbf{54.16}}\\

\multicolumn{1}{c}{EN$\to$TR}  &\multicolumn{1}{c}{31.40/39.05}  &\multicolumn{1}{c}{30.35/32.05}  &\multicolumn{1}{c}{29.55/32.85}  &\multicolumn{1}{c}{30.80/32.40}  &\multicolumn{1}{c}{\underline{39.10}/\underline{41.35}}  &\multicolumn{1}{c}{\textbf{43.55}/\textbf{44.75}}\\

\multicolumn{1}{c}{TR$\to$EN}  &\multicolumn{1}{c}{30.78/37.43}  &\multicolumn{1}{c}{34.45/39.24}  &\multicolumn{1}{c}{28.12/33.49}  &\multicolumn{1}{c}{31.79/35.89}  &\multicolumn{1}{c}{\underline{39.03}/\underline{42.60}}  &\multicolumn{1}{c}{\textbf{39.24}/\textbf{44.78}}\\

\multicolumn{1}{c}{FI$\to$FR}  &\multicolumn{1}{c}{30.90/36.73}  &\multicolumn{1}{c}{34.68/38.26}  &\multicolumn{1}{c}{29.16/34.79}  &\multicolumn{1}{c}{33.79/37.26}  &\multicolumn{1}{c}{\underline{38.94}/\underline{42.20}}  &\multicolumn{1}{c}{\textbf{42.77}/\textbf{45.24}}\\

\multicolumn{1}{c}{FR$\to$FI}  &\multicolumn{1}{c}{29.59/34.92}  &\multicolumn{1}{c}{31.35/34.30}  &\multicolumn{1}{c}{30.42/33.26}  &\multicolumn{1}{c}{30.11/33.26}  &\multicolumn{1}{c}{\underline{36.42}/\underline{39.99}}  &\multicolumn{1}{c}{\textbf{41.18}/\textbf{43.20}}\\

\multicolumn{1}{c}{FI$\to$HR}  &\multicolumn{1}{c}{22.65/28.06}  &\multicolumn{1}{c}{27.17/31.58}  &\multicolumn{1}{c}{24.65/29.06}  &\multicolumn{1}{c}{25.54/29.06}  &\multicolumn{1}{c}{\underline{30.16}/\underline{34.89}}  &\multicolumn{1}{c}{\textbf{34.52}/\textbf{38.31}}\\

\multicolumn{1}{c}{HR$\to$FI}  &\multicolumn{1}{c}{18.20/26.35}  &\multicolumn{1}{c}{28.30/31.72}  &\multicolumn{1}{c}{26.67/31.93}  &\multicolumn{1}{c}{25.78/29.30}  &\multicolumn{1}{c}{\underline{32.51}/\underline{35.61}}  &\multicolumn{1}{c}{\textbf{37.40}/\textbf{39.56}}\\

\multicolumn{1}{c}{FI$\to$IT}  &\multicolumn{1}{c}{31.53/36.94}  &\multicolumn{1}{c}{33.89/37.99}  &\multicolumn{1}{c}{31.37/35.58}  &\multicolumn{1}{c}{33.58/36.15}  &\multicolumn{1}{c}{\underline{38.47}/\underline{42.04}}  &\multicolumn{1}{c}{\textbf{42.51}/\textbf{46.30}}\\

\multicolumn{1}{c}{IT$\to$FI}  &\multicolumn{1}{c}{29.56/34.21}  &\multicolumn{1}{c}{31.06/34.32}  &\multicolumn{1}{c}{31.47/35.09}  &\multicolumn{1}{c}{29.97/33.54}  &\multicolumn{1}{c}{\underline{35.76}/\underline{39.48}}  &\multicolumn{1}{c}{\textbf{40.78}/\textbf{43.57}}\\

\multicolumn{1}{c}{FI$\to$RU}  &\multicolumn{1}{c}{28.74/34.52}  &\multicolumn{1}{c}{31.16/34.16}  &\multicolumn{1}{c}{28.38/32.32}  &\multicolumn{1}{c}{30.37/32.79}  &\multicolumn{1}{c}{\underline{35.10}/\underline{37.73}}  &\multicolumn{1}{c}{\textbf{38.36}/\textbf{40.99}}\\

\multicolumn{1}{c}{RU$\to$FI}  &\multicolumn{1}{c}{27.29/33.11}  &\multicolumn{1}{c}{29.91/33.53}  &\multicolumn{1}{c}{28.60/33.63}  &\multicolumn{1}{c}{27.82/32.53}  &\multicolumn{1}{c}{\underline{35.57}/\underline{36.98}}  &\multicolumn{1}{c}{\textbf{38.55}/\textbf{40.91}}\\

\multicolumn{1}{c}{HR$\to$FR}  &\multicolumn{1}{c}{33.46/39.66}  &\multicolumn{1}{c}{35.35/40.24}  &\multicolumn{1}{c}{30.72/36.09}  &\multicolumn{1}{c}{35.30/38.72}  &\multicolumn{1}{c}{\underline{39.61}/\underline{44.13}}  &\multicolumn{1}{c}{\textbf{45.40}/\textbf{49.29}}\\

\multicolumn{1}{c}{FR$\to$HR}  &\multicolumn{1}{c}{30.94/35.28}  &\multicolumn{1}{c}{29.85/33.21}  &\multicolumn{1}{c}{26.90/30.88}  &\multicolumn{1}{c}{29.69/33.26}  &\multicolumn{1}{c}{\underline{36.32}/\underline{39.78}}  &\multicolumn{1}{c}{\textbf{40.71}/\textbf{44.08}}\\

\multicolumn{1}{c}{HR$\to$IT}  &\multicolumn{1}{c}{29.62/37.98}  &\multicolumn{1}{c}{36.24/40.24}  &\multicolumn{1}{c}{32.14/36.72}  &\multicolumn{1}{c}{34.19/36.98}  &\multicolumn{1}{c}{\underline{38.93}/\underline{43.77}}  &\multicolumn{1}{c}{\textbf{44.71}/\textbf{48.97}}\\

\multicolumn{1}{c}{IT$\to$HR}  &\multicolumn{1}{c}{30.34/34.06}  &\multicolumn{1}{c}{30.75/34.32}  &\multicolumn{1}{c}{27.80/32.87}  &\multicolumn{1}{c}{30.03/33.49}  &\multicolumn{1}{c}{\underline{37.26}/\underline{38.71}}  &\multicolumn{1}{c}{\textbf{41.40}/\textbf{44.75}}\\

\multicolumn{1}{c}{HR$\to$RU}  &\multicolumn{1}{c}{31.35/37.19}  &\multicolumn{1}{c}{34.19/37.98}  &\multicolumn{1}{c}{32.40/36.61}  &\multicolumn{1}{c}{33.19/36.03}  &\multicolumn{1}{c}{\underline{39.40}/\underline{41.66}}  &\multicolumn{1}{c}{\textbf{44.35}/\textbf{46.40}}\\

\multicolumn{1}{c}{RU$\to$HR}  &\multicolumn{1}{c}{31.48/35.94}  &\multicolumn{1}{c}{34.57/39.50}  &\multicolumn{1}{c}{31.48/35.78}  &\multicolumn{1}{c}{32.16/36.56}  &\multicolumn{1}{c}{\underline{37.93}/\underline{40.60}}  &\multicolumn{1}{c}{\textbf{42.17}/\textbf{45.47}}\\

\multicolumn{1}{c}{IT$\to$FR}  &\multicolumn{1}{c}{64.19/\underline{66.51}}  &\multicolumn{1}{c}{64.03/65.89}  &\multicolumn{1}{c}{62.12/64.60}  &\multicolumn{1}{c}{63.57/65.32}  &\multicolumn{1}{c}{\underline{65.37}/\underline{66.51}}  &\multicolumn{1}{c}{\textbf{66.82}/\textbf{67.86}}\\

\multicolumn{1}{c}{FR$\to$IT}  &\multicolumn{1}{c}{62.96/66.11}  &\multicolumn{1}{c}{62.70/64.72}  &\multicolumn{1}{c}{61.05/63.68}  &\multicolumn{1}{c}{62.18/64.30}  &\multicolumn{1}{c}{\underline{64.25}/\underline{66.27}}  &\multicolumn{1}{c}{\textbf{66.79}/\textbf{67.20}}\\

\multicolumn{1}{c}{RU$\to$FR}  &\multicolumn{1}{c}{44.00/47.67}  &\multicolumn{1}{c}{43.58/47.51}  &\multicolumn{1}{c}{38.82/43.64}  &\multicolumn{1}{c}{42.90/47.15}  &\multicolumn{1}{c}{\underline{48.04}/\underline{50.55}}  &\multicolumn{1}{c}{\textbf{50.13}/\textbf{52.70}}\\

\multicolumn{1}{c}{FR$\to$RU}  &\multicolumn{1}{c}{41.02/\underline{45.01}}  &\multicolumn{1}{c}{36.73/38.23}  &\multicolumn{1}{c}{36.26/37.40}  &\multicolumn{1}{c}{37.20/38.54}  &\multicolumn{1}{c}{\underline{43.35}/44.75}  &\multicolumn{1}{c}{\textbf{47.13}/\textbf{48.06}}\\

\multicolumn{1}{c}{RU$\to$IT}  &\multicolumn{1}{c}{41.49/46.57}  &\multicolumn{1}{c}{43.84/46.78}  &\multicolumn{1}{c}{39.50/43.74}  &\multicolumn{1}{c}{43.79/45.89}  &\multicolumn{1}{c}{\underline{46.52}/\underline{49.66}}  &\multicolumn{1}{c}{\textbf{48.66}/\textbf{51.96}}\\

\multicolumn{1}{c}{IT$\to$RU}  &\multicolumn{1}{c}{40.57/44.13}  &\multicolumn{1}{c}{38.35/38.71}  &\multicolumn{1}{c}{35.87/38.09}  &\multicolumn{1}{c}{38.40/39.43}  &\multicolumn{1}{c}{\underline{45.01}/\underline{45.48}}  &\multicolumn{1}{c}{\textbf{47.08}/\textbf{47.49}}\\

\multicolumn{1}{c}{TR$\to$FI}  &\multicolumn{1}{c}{21.46/26.46}  &\multicolumn{1}{c}{24.23/28.59}  &\multicolumn{1}{c}{26.14/30.67}  &\multicolumn{1}{c}{24.12/27.90}  &\multicolumn{1}{c}{\underline{31.31}/\underline{32.96}}  &\multicolumn{1}{c}{\textbf{32.85}/\textbf{34.77}}\\

\multicolumn{1}{c}{FI$\to$TR}  &\multicolumn{1}{c}{23.07/28.90}  &\multicolumn{1}{c}{24.86/29.80}  &\multicolumn{1}{c}{23.86/27.54}  &\multicolumn{1}{c}{24.01/28.64}  &\multicolumn{1}{c}{\underline{30.48}/\underline{32.95}}  &\multicolumn{1}{c}{\textbf{32.74}/\textbf{35.68}}\\

\multicolumn{1}{c}{TR$\to$FR}  &\multicolumn{1}{c}{29.13/36.10}  &\multicolumn{1}{c}{32.96/36.58}  &\multicolumn{1}{c}{30.56/34.08}  &\multicolumn{1}{c}{31.31/34.40}  &\multicolumn{1}{c}{\underline{38.13}/\underline{40.63}}  &\multicolumn{1}{c}{\textbf{41.43}/\textbf{43.88}}\\

\multicolumn{1}{c}{FR$\to$TR}  &\multicolumn{1}{c}{27.42/33.52}  &\multicolumn{1}{c}{28.87/31.76}  &\multicolumn{1}{c}{27.42/30.88}  &\multicolumn{1}{c}{26.44/29.13}  &\multicolumn{1}{c}{\underline{34.97}/\underline{37.82}}  &\multicolumn{1}{c}{\textbf{38.70}/\textbf{42.06}}\\

\multicolumn{1}{c}{TR$\to$HR}  &\multicolumn{1}{c}{20.07/24.60}  &\multicolumn{1}{c}{21.99/25.99}  &\multicolumn{1}{c}{22.42/26.68}  &\multicolumn{1}{c}{21.30/25.24}  &\multicolumn{1}{c}{\underline{29.34}/\underline{32.37}}  &\multicolumn{1}{c}{\textbf{32.43}/\textbf{36.32}}\\

\multicolumn{1}{c}{HR$\to$TR}  &\multicolumn{1}{c}{17.41/25.25}  &\multicolumn{1}{c}{24.62/27.35}  &\multicolumn{1}{c}{22.30/26.20}  &\multicolumn{1}{c}{22.09/24.62}  &\multicolumn{1}{c}{\underline{29.04}/\underline{32.61}}  &\multicolumn{1}{c}{\textbf{34.14}/\textbf{37.09}}\\

\multicolumn{1}{c}{TR$\to$IT}  &\multicolumn{1}{c}{28.91/34.56}  &\multicolumn{1}{c}{31.90/34.24}  &\multicolumn{1}{c}{29.66/32.00}  &\multicolumn{1}{c}{29.82/33.44}  &\multicolumn{1}{c}{\underline{36.32}/\underline{38.98}}  &\multicolumn{1}{c}{\textbf{38.87}/\textbf{42.17}}\\

\multicolumn{1}{c}{IT$\to$TR}  &\multicolumn{1}{c}{28.32/34.73}  &\multicolumn{1}{c}{28.11/30.70}  &\multicolumn{1}{c}{27.96/30.39}  &\multicolumn{1}{c}{27.86/29.82}  &\multicolumn{1}{c}{\underline{35.09}/\underline{37.52}}  &\multicolumn{1}{c}{\textbf{38.19}/\textbf{40.62}}\\

\multicolumn{1}{c}{TR$\to$RU}  &\multicolumn{1}{c}{23.59/28.06}  &\multicolumn{1}{c}{24.07/26.20}  &\multicolumn{1}{c}{21.99/26.20}  &\multicolumn{1}{c}{24.55/26.36}  &\multicolumn{1}{c}{\underline{31.04}/\underline{32.00}}  &\multicolumn{1}{c}{\textbf{33.60}/\textbf{36.16}}\\

\multicolumn{1}{c}{RU$\to$TR}  &\multicolumn{1}{c}{24.46/29.18}  &\multicolumn{1}{c}{23.31/27.08}  &\multicolumn{1}{c}{22.58/25.88}  &\multicolumn{1}{c}{25.04/26.35}  &\multicolumn{1}{c}{\underline{29.81}/\underline{32.74}}  &\multicolumn{1}{c}{\textbf{32.48}/\textbf{35.78}}\\

\multicolumn{1}{c}{Avg.}  &\multicolumn{1}{c}{35.78/41.22}  &\multicolumn{1}{c}{36.76/40.06}  &\multicolumn{1}{c}{34.37/38.22}  &\multicolumn{1}{c}{36.22/39.16}  &\multicolumn{1}{c}{\underline{41.95}/\underline{44.54}}  &\multicolumn{1}{c}{\textbf{45.41}/\textbf{47.88}}\\

\bottomrule

\end{tabular}
}

\caption{BLI results with $5k$ seed translation pairs. BLI prediction accuracy (P@1$\times100\%$) is reported in the NN/CSLS format (NN: Nearest Neighbor retrieval without CSLS adjustment; CSLS: CSLS retrieval). \underline{Underlined} scores denote the highest scores among purely fastText-based methods; \textbf{Bold} scores denote the highest scores in setups where both fastText and mBERT are allowed.}
\label{table:appendix-full}
\end{center}
\end{table*}

\begin{table*}[ht]
\begin{center}
\resizebox{0.79\textwidth}{!}{%
\begin{tabular}{lllllll}
\toprule 

\rowcolor{Gray}
\multicolumn{1}{c}{[1k] \bf Pairs}  &\multicolumn{1}{c}{\bf RCSLS}  &\multicolumn{1}{c}{\bf VecMap-Semi} &\multicolumn{1}{c}{\bf LNMap} &\multicolumn{1}{c}{\bf FIPP}
&\multicolumn{1}{c}{\bf C1}
&\multicolumn{1}{c}{\bf C2 (C1)}
\\ \cmidrule(lr){2-5} \cmidrule(lr){6-7}

\multicolumn{1}{c}{DE$\to$FI}  &\multicolumn{1}{c}{20.97/26.34}  &\multicolumn{1}{c}{23.68/28.33}  &\multicolumn{1}{c}{29.47/32.24}  &\multicolumn{1}{c}{25.56/30.26}  &\multicolumn{1}{c}{\underline{37.35}/\underline{40.85}}  &\multicolumn{1}{c}{\textbf{40.79}/\textbf{43.77}}\\

\multicolumn{1}{c}{FI$\to$DE}  &\multicolumn{1}{c}{21.18/27.01}  &\multicolumn{1}{c}{32.05/35.00}  &\multicolumn{1}{c}{27.64/34.47}  &\multicolumn{1}{c}{31.79/36.73}  &\multicolumn{1}{c}{\underline{37.52}/\underline{40.57}}  &\multicolumn{1}{c}{\textbf{42.83}/\textbf{44.93}}\\

\multicolumn{1}{c}{DE$\to$FR}  &\multicolumn{1}{c}{34.06/41.94}  &\multicolumn{1}{c}{46.17/49.03}  &\multicolumn{1}{c}{43.82/47.21}  &\multicolumn{1}{c}{46.48/50.18}  &\multicolumn{1}{c}{\underline{49.82}/\underline{51.75}}  &\multicolumn{1}{c}{\textbf{52.11}/\textbf{54.04}}\\

\multicolumn{1}{c}{FR$\to$DE}  &\multicolumn{1}{c}{33.89/37.92}  &\multicolumn{1}{c}{42.11/44.34}  &\multicolumn{1}{c}{39.63/42.99}  &\multicolumn{1}{c}{43.30/46.51}  &\multicolumn{1}{c}{\underline{46.09}/\underline{46.82}}  &\multicolumn{1}{c}{\textbf{48.01}/\textbf{48.16}}\\

\multicolumn{1}{c}{DE$\to$HR}  &\multicolumn{1}{c}{19.25/22.59}  &\multicolumn{1}{c}{22.64/27.39}  &\multicolumn{1}{c}{24.26/28.64}  &\multicolumn{1}{c}{21.91/27.18}  &\multicolumn{1}{c}{\underline{30.88}/\underline{35.16}}  &\multicolumn{1}{c}{\textbf{36.46}/\textbf{40.48}}\\

\multicolumn{1}{c}{HR$\to$DE}  &\multicolumn{1}{c}{19.10/23.04}  &\multicolumn{1}{c}{30.98/32.82}  &\multicolumn{1}{c}{25.25/29.46}  &\multicolumn{1}{c}{28.77/31.56}  &\multicolumn{1}{c}{\underline{35.35}/\underline{38.45}}  &\multicolumn{1}{c}{\textbf{41.19}/\textbf{44.35}}\\

\multicolumn{1}{c}{DE$\to$IT}  &\multicolumn{1}{c}{38.81/44.03}  &\multicolumn{1}{c}{46.58/48.72}  &\multicolumn{1}{c}{43.82/47.52}  &\multicolumn{1}{c}{46.01/48.98}  &\multicolumn{1}{c}{\underline{48.93}/\underline{51.28}}  &\multicolumn{1}{c}{\textbf{50.39}/\textbf{52.53}}\\

\multicolumn{1}{c}{IT$\to$DE}  &\multicolumn{1}{c}{36.64/40.83}  &\multicolumn{1}{c}{41.91/44.39}  &\multicolumn{1}{c}{39.69/42.58}  &\multicolumn{1}{c}{42.95/45.94}  &\multicolumn{1}{c}{\underline{46.56}/\underline{47.86}}  &\multicolumn{1}{c}{\textbf{49.41}/\textbf{49.66}}\\

\multicolumn{1}{c}{DE$\to$RU}  &\multicolumn{1}{c}{27.80/32.66}  &\multicolumn{1}{c}{20.97/25.46}  &\multicolumn{1}{c}{27.86/30.73}  &\multicolumn{1}{c}{26.03/30.05}  &\multicolumn{1}{c}{\underline{40.11}/\underline{40.27}}  &\multicolumn{1}{c}{\textbf{42.15}/\textbf{42.83}}\\

\multicolumn{1}{c}{RU$\to$DE}  &\multicolumn{1}{c}{27.82/32.58}  &\multicolumn{1}{c}{36.46/39.08}  &\multicolumn{1}{c}{33.84/37.30}  &\multicolumn{1}{c}{37.98/40.65}  &\multicolumn{1}{c}{\underline{42.33}/\underline{44.21}}  &\multicolumn{1}{c}{\textbf{45.00}/\textbf{46.99}}\\

\multicolumn{1}{c}{DE$\to$TR}  &\multicolumn{1}{c}{14.03/18.21}  &\multicolumn{1}{c}{20.40/23.37}  &\multicolumn{1}{c}{21.39/24.36}  &\multicolumn{1}{c}{18.94/22.85}  &\multicolumn{1}{c}{\underline{29.26}/\underline{32.03}}  &\multicolumn{1}{c}{\textbf{32.24}/\textbf{34.85}}\\

\multicolumn{1}{c}{TR$\to$DE}  &\multicolumn{1}{c}{14.43/18.10}  &\multicolumn{1}{c}{23.22/26.57}  &\multicolumn{1}{c}{20.13/24.55}  &\multicolumn{1}{c}{21.67/25.24}  &\multicolumn{1}{c}{\underline{30.83}/\underline{33.71}}  &\multicolumn{1}{c}{\textbf{34.45}/\textbf{37.11}}\\

\multicolumn{1}{c}{EN$\to$DE}  &\multicolumn{1}{c}{43.00/46.10}  &\multicolumn{1}{c}{46.40/48.20}  &\multicolumn{1}{c}{43.05/45.80}  &\multicolumn{1}{c}{47.95/49.65}  &\multicolumn{1}{c}{\underline{49.65}/\underline{50.40}}  &\multicolumn{1}{c}{\textbf{51.75}/\textbf{50.85}}\\

\multicolumn{1}{c}{DE$\to$EN}  &\multicolumn{1}{c}{43.14/48.25}  &\multicolumn{1}{c}{51.90/54.56}  &\multicolumn{1}{c}{47.16/50.23}  &\multicolumn{1}{c}{50.97/54.41}  &\multicolumn{1}{c}{\underline{53.42}/\underline{56.23}}  &\multicolumn{1}{c}{\textbf{55.24}/\textbf{57.75}}\\

\multicolumn{1}{c}{EN$\to$FI}  &\multicolumn{1}{c}{22.40/28.35}  &\multicolumn{1}{c}{24.30/27.95}  &\multicolumn{1}{c}{29.50/33.60}  &\multicolumn{1}{c}{30.40/34.50}  &\multicolumn{1}{c}{\underline{38.60}/\underline{42.15}}  &\multicolumn{1}{c}{\textbf{43.75}/\textbf{45.00}}\\

\multicolumn{1}{c}{FI$\to$EN}  &\multicolumn{1}{c}{22.70/28.38}  &\multicolumn{1}{c}{37.41/41.15}  &\multicolumn{1}{c}{29.01/35.47}  &\multicolumn{1}{c}{33.68/37.10}  &\multicolumn{1}{c}{\underline{39.73}/\underline{45.51}}  &\multicolumn{1}{c}{\textbf{42.93}/\textbf{48.77}}\\

\multicolumn{1}{c}{EN$\to$FR}  &\multicolumn{1}{c}{49.00/56.50}  &\multicolumn{1}{c}{57.90/60.00}  &\multicolumn{1}{c}{56.85/60.50}  &\multicolumn{1}{c}{59.65/61.60}  &\multicolumn{1}{c}{\underline{60.70}/\underline{61.65}}  &\multicolumn{1}{c}{\textbf{63.65}/\textbf{62.50}}\\

\multicolumn{1}{c}{FR$\to$EN}  &\multicolumn{1}{c}{49.46/55.56}  &\multicolumn{1}{c}{58.35/61.41}  &\multicolumn{1}{c}{54.32/58.41}  &\multicolumn{1}{c}{58.72/61.61}  &\multicolumn{1}{c}{\underline{60.48}/\underline{63.27}}  &\multicolumn{1}{c}{\textbf{62.65}/\textbf{64.05}}\\

\multicolumn{1}{c}{EN$\to$HR}  &\multicolumn{1}{c}{18.65/22.50}  &\multicolumn{1}{c}{21.95/24.95}  &\multicolumn{1}{c}{21.30/25.55}  &\multicolumn{1}{c}{21.70/26.65}  &\multicolumn{1}{c}{\underline{32.65}/\underline{35.65}}  &\multicolumn{1}{c}{\textbf{39.20}/\textbf{42.35}}\\

\multicolumn{1}{c}{HR$\to$EN}  &\multicolumn{1}{c}{16.57/22.88}  &\multicolumn{1}{c}{34.61/37.45}  &\multicolumn{1}{c}{26.35/30.72}  &\multicolumn{1}{c}{29.77/32.93}  &\multicolumn{1}{c}{\underline{35.30}/\underline{40.87}}  &\multicolumn{1}{c}{\textbf{40.35}/\textbf{47.55}}\\

\multicolumn{1}{c}{EN$\to$IT}  &\multicolumn{1}{c}{48.65/55.20}  &\multicolumn{1}{c}{55.15/57.55}  &\multicolumn{1}{c}{54.70/57.60}  &\multicolumn{1}{c}{56.00/58.30}  &\multicolumn{1}{c}{\underline{57.70}/\underline{59.60}}  &\multicolumn{1}{c}{\textbf{60.70}/\textbf{61.05}}\\

\multicolumn{1}{c}{IT$\to$EN}  &\multicolumn{1}{c}{48.22/53.64}  &\multicolumn{1}{c}{56.85/60.78}  &\multicolumn{1}{c}{52.61/56.69}  &\multicolumn{1}{c}{56.59/60.78}  &\multicolumn{1}{c}{\underline{59.17}/\underline{62.64}}  &\multicolumn{1}{c}{\textbf{61.40}/\textbf{63.67}}\\

\multicolumn{1}{c}{EN$\to$RU}  &\multicolumn{1}{c}{31.50/35.50}  &\multicolumn{1}{c}{21.10/25.05}  &\multicolumn{1}{c}{28.50/32.25}  &\multicolumn{1}{c}{32.75/35.15}  &\multicolumn{1}{c}{\underline{43.80}/\underline{42.50}}  &\multicolumn{1}{c}{\textbf{46.55}/\textbf{46.05}}\\

\multicolumn{1}{c}{RU$\to$EN}  &\multicolumn{1}{c}{32.37/36.62}  &\multicolumn{1}{c}{44.37/46.20}  &\multicolumn{1}{c}{36.46/41.17}  &\multicolumn{1}{c}{43.27/46.20}  &\multicolumn{1}{c}{\underline{47.25}/\underline{50.29}}  &\multicolumn{1}{c}{\textbf{48.35}/\textbf{53.17}}\\

\multicolumn{1}{c}{EN$\to$TR}  &\multicolumn{1}{c}{19.35/23.00}  &\multicolumn{1}{c}{24.45/26.70}  &\multicolumn{1}{c}{25.15/27.75}  &\multicolumn{1}{c}{26.40/29.95}  &\multicolumn{1}{c}{\underline{36.60}/\underline{38.15}}  &\multicolumn{1}{c}{\textbf{39.05}/\textbf{41.05}}\\

\multicolumn{1}{c}{TR$\to$EN}  &\multicolumn{1}{c}{19.81/24.65}  &\multicolumn{1}{c}{33.49/37.17}  &\multicolumn{1}{c}{26.94/32.59}  &\multicolumn{1}{c}{29.98/33.76}  &\multicolumn{1}{c}{\underline{36.95}/\underline{42.33}}  &\multicolumn{1}{c}{\textbf{37.86}/\textbf{43.24}}\\

\multicolumn{1}{c}{FI$\to$FR}  &\multicolumn{1}{c}{16.13/22.49}  &\multicolumn{1}{c}{31.84/34.79}  &\multicolumn{1}{c}{25.70/30.01}  &\multicolumn{1}{c}{29.58/33.74}  &\multicolumn{1}{c}{\underline{37.05}/\underline{40.36}}  &\multicolumn{1}{c}{\textbf{40.67}/\textbf{43.30}}\\

\multicolumn{1}{c}{FR$\to$FI}  &\multicolumn{1}{c}{17.69/21.73}  &\multicolumn{1}{c}{21.11/23.95}  &\multicolumn{1}{c}{25.14/28.50}  &\multicolumn{1}{c}{26.49/29.49}  &\multicolumn{1}{c}{\underline{34.30}/\underline{37.61}}  &\multicolumn{1}{c}{\textbf{37.09}/\textbf{40.56}}\\

\multicolumn{1}{c}{FI$\to$HR}  &\multicolumn{1}{c}{15.24/17.24}  &\multicolumn{1}{c}{25.22/29.90}  &\multicolumn{1}{c}{21.86/26.33}  &\multicolumn{1}{c}{23.49/26.90}  &\multicolumn{1}{c}{\underline{25.64}/\underline{30.01}}  &\multicolumn{1}{c}{\textbf{30.74}/\textbf{34.26}}\\

\multicolumn{1}{c}{HR$\to$FI}  &\multicolumn{1}{c}{14.05/18.52}  &\multicolumn{1}{c}{25.04/27.62}  &\multicolumn{1}{c}{23.57/27.83}  &\multicolumn{1}{c}{23.99/27.41}  &\multicolumn{1}{c}{\underline{28.67}/\underline{32.61}}  &\multicolumn{1}{c}{\textbf{33.46}/\textbf{36.14}}\\

\multicolumn{1}{c}{FI$\to$IT}  &\multicolumn{1}{c}{20.13/25.33}  &\multicolumn{1}{c}{32.11/34.68}  &\multicolumn{1}{c}{28.38/31.84}  &\multicolumn{1}{c}{30.27/34.21}  &\multicolumn{1}{c}{\underline{35.89}/\underline{38.99}}  &\multicolumn{1}{c}{\textbf{40.04}/\textbf{42.88}}\\

\multicolumn{1}{c}{IT$\to$FI}  &\multicolumn{1}{c}{19.07/24.60}  &\multicolumn{1}{c}{22.84/26.10}  &\multicolumn{1}{c}{27.80/30.13}  &\multicolumn{1}{c}{27.96/31.01}  &\multicolumn{1}{c}{\underline{34.94}/\underline{37.83}}  &\multicolumn{1}{c}{\textbf{38.71}/\textbf{41.65}}\\

\multicolumn{1}{c}{FI$\to$RU}  &\multicolumn{1}{c}{18.44/21.91}  &\multicolumn{1}{c}{26.69/30.27}  &\multicolumn{1}{c}{23.33/27.69}  &\multicolumn{1}{c}{26.48/30.43}  &\multicolumn{1}{c}{\underline{31.42}/\underline{33.89}}  &\multicolumn{1}{c}{\textbf{34.73}/\textbf{37.15}}\\

\multicolumn{1}{c}{RU$\to$FI}  &\multicolumn{1}{c}{15.72/20.48}  &\multicolumn{1}{c}{29.02/33.11}  &\multicolumn{1}{c}{25.93/31.01}  &\multicolumn{1}{c}{25.93/30.28}  &\multicolumn{1}{c}{\underline{32.27}/\underline{35.31}}  &\multicolumn{1}{c}{\textbf{34.94}/\textbf{37.35}}\\

\multicolumn{1}{c}{HR$\to$FR}  &\multicolumn{1}{c}{17.99/23.04}  &\multicolumn{1}{c}{35.61/39.14}  &\multicolumn{1}{c}{28.35/32.93}  &\multicolumn{1}{c}{30.19/34.67}  &\multicolumn{1}{c}{\underline{37.14}/\underline{41.14}}  &\multicolumn{1}{c}{\textbf{43.08}/\textbf{45.71}}\\

\multicolumn{1}{c}{FR$\to$HR}  &\multicolumn{1}{c}{16.76/20.54}  &\multicolumn{1}{c}{23.80/27.52}  &\multicolumn{1}{c}{24.00/28.45}  &\multicolumn{1}{c}{25.50/28.56}  &\multicolumn{1}{c}{\underline{32.70}/\underline{35.33}}  &\multicolumn{1}{c}{\textbf{36.26}/\textbf{39.68}}\\

\multicolumn{1}{c}{HR$\to$IT}  &\multicolumn{1}{c}{20.52/26.20}  &\multicolumn{1}{c}{36.40/38.77}  &\multicolumn{1}{c}{29.46/33.09}  &\multicolumn{1}{c}{31.93/35.03}  &\multicolumn{1}{c}{\underline{37.40}/\underline{40.24}}  &\multicolumn{1}{c}{\textbf{42.40}/\textbf{46.19}}\\

\multicolumn{1}{c}{IT$\to$HR}  &\multicolumn{1}{c}{18.81/23.72}  &\multicolumn{1}{c}{23.88/28.68}  &\multicolumn{1}{c}{24.81/28.63}  &\multicolumn{1}{c}{26.10/30.44}  &\multicolumn{1}{c}{\underline{33.02}/\underline{35.92}}  &\multicolumn{1}{c}{\textbf{37.62}/\textbf{41.29}}\\

\multicolumn{1}{c}{HR$\to$RU}  &\multicolumn{1}{c}{20.99/24.72}  &\multicolumn{1}{c}{32.40/36.09}  &\multicolumn{1}{c}{29.35/34.30}  &\multicolumn{1}{c}{30.30/34.09}  &\multicolumn{1}{c}{\underline{37.30}/\underline{39.40}}  &\multicolumn{1}{c}{\textbf{40.72}/\textbf{42.14}}\\

\multicolumn{1}{c}{RU$\to$HR}  &\multicolumn{1}{c}{20.32/25.67}  &\multicolumn{1}{c}{34.10/38.08}  &\multicolumn{1}{c}{29.70/33.94}  &\multicolumn{1}{c}{30.91/36.14}  &\multicolumn{1}{c}{\underline{34.68}/\underline{38.92}}  &\multicolumn{1}{c}{\textbf{38.03}/\textbf{41.17}}\\

\multicolumn{1}{c}{IT$\to$FR}  &\multicolumn{1}{c}{55.25/59.95}  &\multicolumn{1}{c}{\underline{63.41}/65.06}  &\multicolumn{1}{c}{60.93/63.93}  &\multicolumn{1}{c}{63.05/65.22}  &\multicolumn{1}{c}{\underline{63.41}/\underline{65.63}}  &\multicolumn{1}{c}{\textbf{65.27}/\textbf{66.77}}\\

\multicolumn{1}{c}{FR$\to$IT}  &\multicolumn{1}{c}{55.25/59.91}  &\multicolumn{1}{c}{62.13/63.58}  &\multicolumn{1}{c}{60.37/62.80}  &\multicolumn{1}{c}{61.98/64.15}  &\multicolumn{1}{c}{\underline{63.11}/\underline{64.56}}  &\multicolumn{1}{c}{\textbf{64.46}/\textbf{65.49}}\\

\multicolumn{1}{c}{RU$\to$FR}  &\multicolumn{1}{c}{26.72/33.68}  &\multicolumn{1}{c}{42.33/45.42}  &\multicolumn{1}{c}{36.04/40.54}  &\multicolumn{1}{c}{41.91/46.57}  &\multicolumn{1}{c}{\underline{46.52}/\underline{48.87}}  &\multicolumn{1}{c}{\textbf{48.87}/\textbf{51.28}}\\

\multicolumn{1}{c}{FR$\to$RU}  &\multicolumn{1}{c}{27.06/30.83}  &\multicolumn{1}{c}{20.33/24.57}  &\multicolumn{1}{c}{27.57/31.92}  &\multicolumn{1}{c}{29.69/32.90}  &\multicolumn{1}{c}{\underline{40.71}/\underline{40.46}}  &\multicolumn{1}{c}{\textbf{43.66}/\textbf{43.61}}\\

\multicolumn{1}{c}{RU$\to$IT}  &\multicolumn{1}{c}{30.59/35.36}  &\multicolumn{1}{c}{41.91/43.74}  &\multicolumn{1}{c}{38.92/41.80}  &\multicolumn{1}{c}{42.54/44.94}  &\multicolumn{1}{c}{\underline{45.10}/\underline{48.35}}  &\multicolumn{1}{c}{\textbf{46.46}/\textbf{49.24}}\\

\multicolumn{1}{c}{IT$\to$RU}  &\multicolumn{1}{c}{29.82/32.97}  &\multicolumn{1}{c}{22.89/26.10}  &\multicolumn{1}{c}{29.20/31.47}  &\multicolumn{1}{c}{33.49/35.76}  &\multicolumn{1}{c}{\underline{41.34}/\underline{41.50}}  &\multicolumn{1}{c}{\textbf{43.41}/\textbf{43.57}}\\

\multicolumn{1}{c}{TR$\to$FI}  &\multicolumn{1}{c}{13.31/16.03}  &\multicolumn{1}{c}{19.81/24.76}  &\multicolumn{1}{c}{21.73/26.36}  &\multicolumn{1}{c}{21.73/26.20}  &\multicolumn{1}{c}{\underline{26.94}/\underline{29.93}}  &\multicolumn{1}{c}{\textbf{30.35}/\textbf{32.96}}\\

\multicolumn{1}{c}{FI$\to$TR}  &\multicolumn{1}{c}{11.77/15.08}  &\multicolumn{1}{c}{21.97/25.80}  &\multicolumn{1}{c}{19.71/24.17}  &\multicolumn{1}{c}{21.49/25.64}  &\multicolumn{1}{c}{\underline{24.96}/\underline{28.32}}  &\multicolumn{1}{c}{\textbf{27.80}/\textbf{30.64}}\\

\multicolumn{1}{c}{TR$\to$FR}  &\multicolumn{1}{c}{16.67/20.23}  &\multicolumn{1}{c}{30.46/32.85}  &\multicolumn{1}{c}{26.57/31.52}  &\multicolumn{1}{c}{28.27/31.84}  &\multicolumn{1}{c}{\underline{35.46}/\underline{38.82}}  &\multicolumn{1}{c}{\textbf{38.92}/\textbf{41.59}}\\

\multicolumn{1}{c}{FR$\to$TR}  &\multicolumn{1}{c}{14.43/18.37}  &\multicolumn{1}{c}{22.19/25.19}  &\multicolumn{1}{c}{23.02/25.30}  &\multicolumn{1}{c}{21.83/24.37}  &\multicolumn{1}{c}{\underline{32.02}/\underline{35.59}}  &\multicolumn{1}{c}{\textbf{35.70}/\textbf{38.44}}\\

\multicolumn{1}{c}{TR$\to$HR}  &\multicolumn{1}{c}{11.66/13.84}  &\multicolumn{1}{c}{16.19/20.50}  &\multicolumn{1}{c}{19.01/22.15}  &\multicolumn{1}{c}{17.15/21.19}  &\multicolumn{1}{c}{\underline{22.74}/\underline{27.00}}  &\multicolumn{1}{c}{\textbf{27.85}/\textbf{32.16}}\\

\multicolumn{1}{c}{HR$\to$TR}  &\multicolumn{1}{c}{10.10/12.73}  &\multicolumn{1}{c}{19.57/20.67}  &\multicolumn{1}{c}{18.57/21.99}  &\multicolumn{1}{c}{18.36/20.83}  &\multicolumn{1}{c}{\underline{22.51}/\underline{28.25}}  &\multicolumn{1}{c}{\textbf{28.88}/\textbf{33.04}}\\

\multicolumn{1}{c}{TR$\to$IT}  &\multicolumn{1}{c}{17.15/22.31}  &\multicolumn{1}{c}{29.29/31.42}  &\multicolumn{1}{c}{26.94/29.66}  &\multicolumn{1}{c}{26.62/30.56}  &\multicolumn{1}{c}{\underline{33.65}/\underline{36.47}}  &\multicolumn{1}{c}{\textbf{36.42}/\textbf{39.19}}\\

\multicolumn{1}{c}{IT$\to$TR}  &\multicolumn{1}{c}{16.12/20.98}  &\multicolumn{1}{c}{22.22/25.06}  &\multicolumn{1}{c}{23.93/26.10}  &\multicolumn{1}{c}{23.62/26.25}  &\multicolumn{1}{c}{\underline{32.66}/\underline{34.47}}  &\multicolumn{1}{c}{\textbf{35.50}/\textbf{37.93}}\\

\multicolumn{1}{c}{TR$\to$RU}  &\multicolumn{1}{c}{12.94/15.87}  &\multicolumn{1}{c}{13.05/15.55}  &\multicolumn{1}{c}{15.87/19.60}  &\multicolumn{1}{c}{17.04/20.55}  &\multicolumn{1}{c}{\underline{25.35}/\underline{28.12}}  &\multicolumn{1}{c}{\textbf{29.82}/\textbf{31.95}}\\

\multicolumn{1}{c}{RU$\to$TR}  &\multicolumn{1}{c}{11.42/14.77}  &\multicolumn{1}{c}{16.61/18.60}  &\multicolumn{1}{c}{17.02/20.12}  &\multicolumn{1}{c}{20.90/22.89}  &\multicolumn{1}{c}{\underline{26.40}/\underline{29.54}}  &\multicolumn{1}{c}{\textbf{30.07}/\textbf{33.05}}\\

\multicolumn{1}{c}{Avg.}  &\multicolumn{1}{c}{24.73/29.31}  &\multicolumn{1}{c}{32.50/35.56}  &\multicolumn{1}{c}{31.10/34.90}  &\multicolumn{1}{c}{33.00/36.45}  &\multicolumn{1}{c}{\underline{38.97}/\underline{41.74}}  &\multicolumn{1}{c}{\textbf{42.33}/\textbf{44.77}}\\

\bottomrule

\end{tabular}
}

\caption{BLI results with $1k$ seed translation pairs. BLI prediction accuracy (P@1$\times100\%$) is reported in the NN/CSLS format (NN: Nearest Neighbor retrieval without CSLS adjustment; CSLS: CSLS retrieval). \underline{Underlined} scores denote the highest scores among purely fastText-based methods; \textbf{Bold} scores denote the highest scores in setups where both fastText and mBERT are allowed.}
\label{table:appendix2}
\end{center}
\end{table*}

\section{Full Ablation Study}
\label{appendix:ablation}
Complete results of the ablation study, over all languages in the evaluation set of \newcite{glavas-etal-2019-properly}, are available in Table~\ref{table:appendix-ablation}, and can be seen as additional evidence which supports the claims from the main paper (see \S\ref{s:further})

\begin{table*}[!t]
\begin{center}
\resizebox{1.0\textwidth}{!}{%
\begin{tabular}{llllllll}
\toprule 

\rowcolor{Gray}
\multicolumn{1}{c}{[5k] \bf Pairs}  &\multicolumn{1}{c}{\bf C1 w/o CL}  &\multicolumn{1}{c}{\bf C1 w/o SL} &\multicolumn{1}{c}{\bf C1} &\multicolumn{1}{c}{\bf mBERT}
&\multicolumn{1}{c}{\bf mBERT(tuned)}
&\multicolumn{1}{c}{\bf C1+mBERT}
&\multicolumn{1}{c}{\bf C2 (C1)}
\\ \cmidrule(lr){2-4} \cmidrule(lr){5-8}

\multicolumn{1}{c}{DE$\to$$*$}  &\multicolumn{1}{c}{35.16/39.30}  &\multicolumn{1}{c}{41.70/45.07}  &\multicolumn{1}{c}{\underline{43.43}/\underline{46.14}}  &\multicolumn{1}{c}{8.90/9.39}  &\multicolumn{1}{c}{17.70/18.66}  &\multicolumn{1}{c}{43.13/46.25}  &\multicolumn{1}{c}{\textbf{46.24}/\textbf{48.86}}\\

\multicolumn{1}{c}{$*$$\to$DE}  &\multicolumn{1}{c}{37.24/41.23}  &\multicolumn{1}{c}{43.46/45.85}  &\multicolumn{1}{c}{\underline{44.85}/\underline{46.39}}  &\multicolumn{1}{c}{8.86/9.51}  &\multicolumn{1}{c}{18.10/19.21}  &\multicolumn{1}{c}{44.61/46.47}  &\multicolumn{1}{c}{\textbf{48.96}/\textbf{50.12}}\\

\multicolumn{1}{c}{EN$\to$$*$}  &\multicolumn{1}{c}{37.99/41.58}  &\multicolumn{1}{c}{48.41/50.99}  &\multicolumn{1}{c}{\underline{49.54}/\underline{51.31}}  &\multicolumn{1}{c}{9.29/9.55}  &\multicolumn{1}{c}{15.08/15.87}  &\multicolumn{1}{c}{49.44/51.55}  &\multicolumn{1}{c}{\textbf{53.78}/\textbf{54.31}}\\

\multicolumn{1}{c}{$*$$\to$EN}  &\multicolumn{1}{c}{46.36/50.16}  &\multicolumn{1}{c}{47.36/51.18}  &\multicolumn{1}{c}{\underline{49.21}/\underline{52.61}}  &\multicolumn{1}{c}{10.42/10.71}  &\multicolumn{1}{c}{21.34/22.58}  &\multicolumn{1}{c}{48.96/52.77}  &\multicolumn{1}{c}{\textbf{51.22}/\textbf{55.47}}\\

\multicolumn{1}{c}{FI$\to$$*$}  &\multicolumn{1}{c}{31.92/36.78}  &\multicolumn{1}{c}{33.62/38.21}  &\multicolumn{1}{c}{\underline{36.35}/\underline{39.80}}  &\multicolumn{1}{c}{5.73/5.93}  &\multicolumn{1}{c}{12.23/13.23}  &\multicolumn{1}{c}{35.97/40.00}  &\multicolumn{1}{c}{\textbf{40.00}/\textbf{43.44}}\\

\multicolumn{1}{c}{$*$$\to$FI}  &\multicolumn{1}{c}{26.16/31.13}  &\multicolumn{1}{c}{33.07/37.26}  &\multicolumn{1}{c}{\underline{35.89}/\underline{38.82}}  &\multicolumn{1}{c}{5.57/5.89}  &\multicolumn{1}{c}{11.99/12.95}  &\multicolumn{1}{c}{35.48/39.05}  &\multicolumn{1}{c}{\textbf{39.67}/\textbf{41.97}}\\

\multicolumn{1}{c}{FR$\to$$*$}  &\multicolumn{1}{c}{38.60/42.41}  &\multicolumn{1}{c}{45.27/48.40}  &\multicolumn{1}{c}{\underline{46.81}/\underline{49.12}}  &\multicolumn{1}{c}{9.65/10.18}  &\multicolumn{1}{c}{18.37/19.70}  &\multicolumn{1}{c}{46.65/49.29}  &\multicolumn{1}{c}{\textbf{50.29}/\textbf{51.91}}\\

\multicolumn{1}{c}{$*$$\to$FR}  &\multicolumn{1}{c}{45.30/48.85}  &\multicolumn{1}{c}{47.35/50.82}  &\multicolumn{1}{c}{\underline{49.42}/\underline{51.84}}  &\multicolumn{1}{c}{9.86/10.38}  &\multicolumn{1}{c}{20.01/21.10}  &\multicolumn{1}{c}{49.07/51.92}  &\multicolumn{1}{c}{\textbf{52.73}/\textbf{54.53}}\\

\multicolumn{1}{c}{HR$\to$$*$}  &\multicolumn{1}{c}{30.88/35.52}  &\multicolumn{1}{c}{33.95/38.51}  &\multicolumn{1}{c}{\underline{36.76}/\underline{40.22}}  &\multicolumn{1}{c}{7.11/7.72}  &\multicolumn{1}{c}{17.52/18.57}  &\multicolumn{1}{c}{36.13/40.40}  &\multicolumn{1}{c}{\textbf{41.95}/\textbf{45.53}}\\

\multicolumn{1}{c}{$*$$\to$HR}  &\multicolumn{1}{c}{26.94/32.19}  &\multicolumn{1}{c}{32.24/36.42}  &\multicolumn{1}{c}{\underline{34.67}/\underline{37.82}}  &\multicolumn{1}{c}{7.09/7.54}  &\multicolumn{1}{c}{16.83/17.81}  &\multicolumn{1}{c}{34.23/38.08}  &\multicolumn{1}{c}{\textbf{39.13}/\textbf{42.65}}\\

\multicolumn{1}{c}{IT$\to$$*$}  &\multicolumn{1}{c}{39.06/42.67}  &\multicolumn{1}{c}{45.55/48.39}  &\multicolumn{1}{c}{\underline{46.91}/\underline{48.92}}  &\multicolumn{1}{c}{7.47/8.13}  &\multicolumn{1}{c}{18.64/20.18}  &\multicolumn{1}{c}{46.35/48.91}  &\multicolumn{1}{c}{\textbf{50.06}/\textbf{51.91}}\\

\multicolumn{1}{c}{$*$$\to$IT}  &\multicolumn{1}{c}{44.48/47.60}  &\multicolumn{1}{c}{46.35/49.93}  &\multicolumn{1}{c}{\underline{48.10}/\underline{50.99}}  &\multicolumn{1}{c}{7.03/7.46}  &\multicolumn{1}{c}{16.24/17.12}  &\multicolumn{1}{c}{47.66/51.07}  &\multicolumn{1}{c}{\textbf{51.33}/\textbf{53.85}}\\

\multicolumn{1}{c}{RU$\to$$*$}  &\multicolumn{1}{c}{37.46/40.84}  &\multicolumn{1}{c}{39.30/42.81}  &\multicolumn{1}{c}{\underline{41.77}/\underline{44.17}}  &\multicolumn{1}{c}{1.95/2.29}  &\multicolumn{1}{c}{14.50/15.74}  &\multicolumn{1}{c}{41.56/44.38}  &\multicolumn{1}{c}{\textbf{44.25}/\textbf{47.24}}\\

\multicolumn{1}{c}{$*$$\to$RU}  &\multicolumn{1}{c}{27.85/32.12}  &\multicolumn{1}{c}{39.04/41.46}  &\multicolumn{1}{c}{\underline{40.66}/\underline{42.15}}  &\multicolumn{1}{c}{1.38/1.94}  &\multicolumn{1}{c}{11.47/13.25}  &\multicolumn{1}{c}{40.53/42.39}  &\multicolumn{1}{c}{\textbf{43.73}/\textbf{45.20}}\\

\multicolumn{1}{c}{TR$\to$$*$}  &\multicolumn{1}{c}{26.14/30.92}  &\multicolumn{1}{c}{31.12/35.08}  &\multicolumn{1}{c}{\underline{34.07}/\underline{36.61}}  &\multicolumn{1}{c}{6.18/6.53}  &\multicolumn{1}{c}{12.10/12.87}  &\multicolumn{1}{c}{33.41/36.81}  &\multicolumn{1}{c}{\textbf{36.70}/\textbf{39.86}}\\

\multicolumn{1}{c}{$*$$\to$TR}  &\multicolumn{1}{c}{22.88/26.74}  &\multicolumn{1}{c}{30.08/34.55}  &\multicolumn{1}{c}{\underline{32.83}/\underline{35.67}}  &\multicolumn{1}{c}{6.07/6.28}  &\multicolumn{1}{c}{10.14/10.79}  &\multicolumn{1}{c}{32.09/35.85}  &\multicolumn{1}{c}{\textbf{36.52}/\textbf{39.26}}\\

\multicolumn{1}{c}{Avg.}  &\multicolumn{1}{c}{34.65/38.75}  &\multicolumn{1}{c}{39.87/43.43}  &\multicolumn{1}{c}{\underline{41.95}/\underline{44.54}}  &\multicolumn{1}{c}{7.04/7.46}  &\multicolumn{1}{c}{15.77/16.85}  &\multicolumn{1}{c}{41.58/44.70}  &\multicolumn{1}{c}{\textbf{45.41}/\textbf{47.88}}\\


\toprule 

\rowcolor{Gray}
\multicolumn{1}{c}{[1k] \bf Pairs}  &\multicolumn{1}{c}{\bf C1 w/o CL}  &\multicolumn{1}{c}{\bf C1 w/o SL} &\multicolumn{1}{c}{\bf C1} &\multicolumn{1}{c}{\bf mBERT}
&\multicolumn{1}{c}{\bf mBERT(tuned)}
&\multicolumn{1}{c}{\bf C1+mBERT}
&\multicolumn{1}{c}{\bf C2 (C1)}
\\ \cmidrule(lr){2-4} \cmidrule(lr){5-8}

\multicolumn{1}{c}{DE$\to$$*$}  &\multicolumn{1}{c}{33.39/37.54}  &\multicolumn{1}{c}{24.74/32.59}  &\multicolumn{1}{c}{\underline{41.40}/\underline{43.94}}  &\multicolumn{1}{c}{8.90/9.39}  &\multicolumn{1}{c}{20.26/20.92}  &\multicolumn{1}{c}{41.46/44.08}  &\multicolumn{1}{c}{\textbf{44.20}/\textbf{46.61}}\\

\multicolumn{1}{c}{$*$$\to$DE}  &\multicolumn{1}{c}{35.21/38.73}  &\multicolumn{1}{c}{24.01/32.08}  &\multicolumn{1}{c}{\underline{41.19}/\underline{43.15}}  &\multicolumn{1}{c}{8.86/9.51}  &\multicolumn{1}{c}{20.78/21.10}  &\multicolumn{1}{c}{41.48/43.37}  &\multicolumn{1}{c}{\textbf{44.66}/\textbf{46.01}}\\

\multicolumn{1}{c}{EN$\to$$*$}  &\multicolumn{1}{c}{35.65/39.46}  &\multicolumn{1}{c}{33.21/39.31}  &\multicolumn{1}{c}{\underline{45.67}/\underline{47.16}}  &\multicolumn{1}{c}{9.29/9.55}  &\multicolumn{1}{c}{16.92/17.29}  &\multicolumn{1}{c}{46.05/47.56}  &\multicolumn{1}{c}{\textbf{49.24}/\textbf{49.84}}\\

\multicolumn{1}{c}{$*$$\to$EN}  &\multicolumn{1}{c}{44.95/49.02}  &\multicolumn{1}{c}{28.26/39.19}  &\multicolumn{1}{c}{\underline{47.47}/\underline{51.59}}  &\multicolumn{1}{c}{10.42/10.71}  &\multicolumn{1}{c}{26.11/26.82}  &\multicolumn{1}{c}{47.08/51.63}  &\multicolumn{1}{c}{\textbf{49.83}/\textbf{54.03}}\\

\multicolumn{1}{c}{FI$\to$$*$}  &\multicolumn{1}{c}{29.34/33.91}  &\multicolumn{1}{c}{13.17/21.10}  &\multicolumn{1}{c}{\underline{33.17}/\underline{36.81}}  &\multicolumn{1}{c}{5.73/5.93}  &\multicolumn{1}{c}{15.66/16.13}  &\multicolumn{1}{c}{33.15/36.90}  &\multicolumn{1}{c}{\textbf{37.11}/\textbf{40.28}}\\

\multicolumn{1}{c}{$*$$\to$FI}  &\multicolumn{1}{c}{23.35/28.38}  &\multicolumn{1}{c}{14.12/20.73}  &\multicolumn{1}{c}{\underline{33.30}/\underline{36.61}}  &\multicolumn{1}{c}{5.57/5.89}  &\multicolumn{1}{c}{14.80/15.35}  &\multicolumn{1}{c}{33.27/36.83}  &\multicolumn{1}{c}{\textbf{37.01}/\textbf{39.63}}\\

\multicolumn{1}{c}{FR$\to$$*$}  &\multicolumn{1}{c}{36.34/39.49}  &\multicolumn{1}{c}{27.86/34.51}  &\multicolumn{1}{c}{\underline{44.20}/\underline{46.23}}  &\multicolumn{1}{c}{9.65/10.18}  &\multicolumn{1}{c}{20.74/21.59}  &\multicolumn{1}{c}{44.15/46.52}  &\multicolumn{1}{c}{\textbf{46.83}/\textbf{48.57}}\\

\multicolumn{1}{c}{$*$$\to$FR}  &\multicolumn{1}{c}{44.06/47.64}  &\multicolumn{1}{c}{28.73/36.32}  &\multicolumn{1}{c}{\underline{47.16}/\underline{49.75}}  &\multicolumn{1}{c}{9.86/10.38}  &\multicolumn{1}{c}{23.03/23.59}  &\multicolumn{1}{c}{47.24/49.88}  &\multicolumn{1}{c}{\textbf{50.37}/\textbf{52.17}}\\

\multicolumn{1}{c}{HR$\to$$*$}  &\multicolumn{1}{c}{28.42/33.07}  &\multicolumn{1}{c}{12.40/20.76}  &\multicolumn{1}{c}{\underline{33.38}/\underline{37.28}}  &\multicolumn{1}{c}{7.11/7.72}  &\multicolumn{1}{c}{20.41/20.97}  &\multicolumn{1}{c}{33.01/37.38}  &\multicolumn{1}{c}{\textbf{38.58}/\textbf{42.16}}\\

\multicolumn{1}{c}{$*$$\to$HR}  &\multicolumn{1}{c}{24.15/28.84}  &\multicolumn{1}{c}{14.61/20.67}  &\multicolumn{1}{c}{\underline{30.33}/\underline{34.00}}  &\multicolumn{1}{c}{7.09/7.54}  &\multicolumn{1}{c}{19.18/19.74}  &\multicolumn{1}{c}{30.49/34.30}  &\multicolumn{1}{c}{\textbf{35.17}/\textbf{38.77}}\\

\multicolumn{1}{c}{IT$\to$$*$}  &\multicolumn{1}{c}{36.71/40.37}  &\multicolumn{1}{c}{29.04/36.45}  &\multicolumn{1}{c}{\underline{44.44}/\underline{46.55}}  &\multicolumn{1}{c}{7.47/8.13}  &\multicolumn{1}{c}{22.25/23.29}  &\multicolumn{1}{c}{44.42/46.74}  &\multicolumn{1}{c}{\textbf{47.33}/\textbf{49.22}}\\

\multicolumn{1}{c}{$*$$\to$IT}  &\multicolumn{1}{c}{43.02/46.05}  &\multicolumn{1}{c}{29.42/37.68}  &\multicolumn{1}{c}{\underline{45.97}/\underline{48.50}}  &\multicolumn{1}{c}{7.03/7.46}  &\multicolumn{1}{c}{19.27/19.86}  &\multicolumn{1}{c}{45.75/48.54}  &\multicolumn{1}{c}{\textbf{48.70}/\textbf{50.94}}\\

\multicolumn{1}{c}{RU$\to$$*$}  &\multicolumn{1}{c}{35.36/38.69}  &\multicolumn{1}{c}{18.95/27.72}  &\multicolumn{1}{c}{\underline{39.22}/\underline{42.21}}  &\multicolumn{1}{c}{1.95/2.29}  &\multicolumn{1}{c}{18.86/19.12}  &\multicolumn{1}{c}{39.09/42.27}  &\multicolumn{1}{c}{\textbf{41.67}/\textbf{44.61}}\\

\multicolumn{1}{c}{$*$$\to$RU}  &\multicolumn{1}{c}{24.33/28.77}  &\multicolumn{1}{c}{20.82/26.62}  &\multicolumn{1}{c}{\underline{37.15}/\underline{38.02}}  &\multicolumn{1}{c}{1.38/1.94}  &\multicolumn{1}{c}{14.57/15.74}  &\multicolumn{1}{c}{37.41/38.37}  &\multicolumn{1}{c}{\textbf{40.15}/\textbf{41.04}}\\

\multicolumn{1}{c}{TR$\to$$*$}  &\multicolumn{1}{c}{24.06/28.62}  &\multicolumn{1}{c}{11.39/18.22}  &\multicolumn{1}{c}{\underline{30.27}/\underline{33.77}}  &\multicolumn{1}{c}{6.18/6.53}  &\multicolumn{1}{c}{14.80/15.28}  &\multicolumn{1}{c}{30.07/33.92}  &\multicolumn{1}{c}{\textbf{33.67}/\textbf{36.89}}\\

\multicolumn{1}{c}{$*$$\to$TR}  &\multicolumn{1}{c}{20.19/23.73}  &\multicolumn{1}{c}{10.80/17.36}  &\multicolumn{1}{c}{\underline{29.20}/\underline{32.34}}  &\multicolumn{1}{c}{6.07/6.28}  &\multicolumn{1}{c}{12.14/12.40}  &\multicolumn{1}{c}{28.69/32.44}  &\multicolumn{1}{c}{\textbf{32.75}/\textbf{35.57}}\\

\multicolumn{1}{c}{Avg.}  &\multicolumn{1}{c}{32.41/36.39}  &\multicolumn{1}{c}{21.35/28.83}  &\multicolumn{1}{c}{\underline{38.97}/\underline{41.74}}  &\multicolumn{1}{c}{7.04/7.46}  &\multicolumn{1}{c}{18.74/19.32}  &\multicolumn{1}{c}{38.93/41.92}  &\multicolumn{1}{c}{\textbf{42.33}/\textbf{44.77}}\\

\bottomrule
\end{tabular}
}

\caption{Full ablation study on $8$ languages, $28$ language pairs in both directions with $5k$ and $1k$ seed translation pairs respectively, that is, $112$ BLI setups for each method. $L\to$$*$ and $*\to$$L$ denote the average BLI scores of BLI setups where $L$ is the source and the target language, respectively. BLI prediction accuracy (P@1$\times100\%$) is reported in the NN/CSLS format (NN: Nearest Neighbor retrieval without CSLS adjustment; CSLS: CSLS retrieval). \underline{Underlined} scores denote the highest scores among purely fastText-based methods; \textbf{Bold} scores denote the highest scores in setups where both fastText and mBERT are allowed.}
\label{table:appendix-ablation}
\end{center}
\end{table*}

\section{Translation Examples}
\label{appendix:transexamples}

We showcase some translation examples of both C1 alignment (see \S\ref{s:c1}) and C2 alignment (see \S\ref{s:c2}) in HR$\to$EN and IT$\to$EN word translation directions. In order to gain insight into the effectiveness of contrastive learning, we adopt C1 w/o CL as a baseline (also used in Table~\ref{table:ablationmain}). All three models (i.e., C1 w/o CL, C1 and C2) are learned with $5k$ seed training word pairs, and we report top five predictions via Nearest Neighbor (NN) retrieval (for simplicity) on the BLI test sets. 

We consider both SUCCESS and FAIL examples in terms of BLI-oriented contrastive fine-tuning, where `SUCCESS' represents the cases where at least one of C1 and C2 predicts the correct answer when the baseline fails, and `FAIL' denotes the scenario where the baseline succeeds but both C1 and C2 make wrong predictions. Here, we show some statistics for each language pair: (1) HR$\to$EN sees $280$ SUCCESS samples and $77$ FAIL ones; (2) IT$\to$EN has $162$ SUCCESS data points, but only $26$ FAIL ones. Table~\ref{table:translationexamples} provides $5$ SUCCESS examples and $5$ FAIL ones for each of the two language pairs.\footnote{In some cases, a source word may have two or more translations in the test set $\mathcal{D}_T$. A prediction is then considered correct if it is any one of the ground-truth translations in $\mathcal{D}_T$. The P@1 scores throughout our study are also calculated in this fashion, derived by evaluating the translation of each unique source word.} 

\begin{table*}[ht!]
\begin{center}
\resizebox{0.955\textwidth}{!}{%
\begin{tabular}{llll}
\toprule 
\rowcolor{Gray}


\multicolumn{1}{c}{\bf Ground Truth Translation Pair}&\multicolumn{1}{c}{\bf Method} &\multicolumn{1}{c}{\bf Top Five Predictions} &\multicolumn{1}{c}{\bf Effectiveness of CL}

\\ \cmidrule(lr){1-4}
\multicolumn{1}{l}{prepoznaje (HR) $\to$ \textbf{recognizes} (EN)}&\multicolumn{1}{l}{Baseline}&\multicolumn{1}{l}{\hspace{1.168cm}explains identifies reveals perceives \textbf{recognizes}}  &\multicolumn{1}{c}{SUCCESS} \\

\multicolumn{1}{l}{}&\multicolumn{1}{l}{C1}&\multicolumn{1}{l}{\hspace{1.168cm}\textbf{recognizes} identifies expresses interprets reveals}  &\multicolumn{1}{c}{} \\

\multicolumn{1}{l}{}&\multicolumn{1}{l}{C2 (C1)}&\multicolumn{1}{l}{\hspace{1.168cm}identifies \textbf{recognizes} recognises reveals interprets}  &\multicolumn{1}{c}{} \\

\cmidrule(lr){1-4}
\multicolumn{1}{l}{majmuni (HR) $\to$ \textbf{monkeys} (EN)}&\multicolumn{1}{l}{Baseline}&\multicolumn{1}{l}{\hspace{1.168cm}sloths lemurs \textbf{monkeys} tarsiers apes}  &\multicolumn{1}{c}{SUCCESS} \\

\multicolumn{1}{l}{}&\multicolumn{1}{l}{C1}&\multicolumn{1}{l}{\hspace{1.168cm}\textbf{monkeys} apes gorillas anteaters chimps}  &\multicolumn{1}{c}{} \\

\multicolumn{1}{l}{}&\multicolumn{1}{l}{C2 (C1)}&\multicolumn{1}{l}{\hspace{1.168cm}\textbf{monkeys} apes gorillas dinosaurs animals}  &\multicolumn{1}{c}{} \\

\cmidrule(lr){1-4}
\multicolumn{1}{l}{enzimi (HR) $\to$ \textbf{enzymes} (EN)}&\multicolumn{1}{l}{Baseline}&\multicolumn{1}{l}{\hspace{1.168cm}proteins proteases enzymatic \textbf{enzymes} enzymatically}  &\multicolumn{1}{c}{SUCCESS} \\

\multicolumn{1}{l}{}&\multicolumn{1}{l}{C1}&\multicolumn{1}{l}{\hspace{1.168cm}proteins \textbf{enzymes} acids peptides polypeptides}  &\multicolumn{1}{c}{} \\

\multicolumn{1}{l}{}&\multicolumn{1}{l}{C2 (C1)}&\multicolumn{1}{l}{\hspace{1.168cm}\textbf{enzymes} proteins acids molecules peptides}  &\multicolumn{1}{c}{} \\

\cmidrule(lr){1-4}
\multicolumn{1}{l}{breskva (HR) $\to$ \textbf{peach} (EN)}&\multicolumn{1}{l}{Baseline}&\multicolumn{1}{l}{\hspace{1.168cm}strawberries plums cherries persimmons peaches}  &\multicolumn{1}{c}{SUCCESS} \\

\multicolumn{1}{l}{}&\multicolumn{1}{l}{C1}&\multicolumn{1}{l}{\hspace{1.168cm}peaches \textbf{peach} mango damson honey}  &\multicolumn{1}{c}{} \\

\multicolumn{1}{l}{}&\multicolumn{1}{l}{C2 (C1)}&\multicolumn{1}{l}{\hspace{1.168cm}\textbf{peach} berry plum mango vine}  &\multicolumn{1}{c}{} \\

\cmidrule(lr){1-4}
\multicolumn{1}{l}{brada (HR) $\to$ \textbf{beard} or \textbf{chin} (EN)}&\multicolumn{1}{l}{Baseline}&\multicolumn{1}{l}{\hspace{1.168cm}cheekbones cheeks whiskers hair cheek}  &\multicolumn{1}{c}{SUCCESS} \\

\multicolumn{1}{l}{}&\multicolumn{1}{l}{C1}&\multicolumn{1}{l}{\hspace{1.168cm}hair cheek collar \textbf{beard} rooney}  &\multicolumn{1}{c}{} \\

\multicolumn{1}{l}{}&\multicolumn{1}{l}{C2 (C1)}&\multicolumn{1}{l}{\hspace{1.168cm}\textbf{beard} hair collar belly neck}  &\multicolumn{1}{c}{} \\

\cmidrule(lr){1-4}
\multicolumn{1}{l}{algebarski (HR) $\to$ \textbf{algebraic} (EN)}&\multicolumn{1}{l}{Baseline}&\multicolumn{1}{l}{\hspace{1.168cm}\textbf{algebraic} equational algebraically quaternionic integrals}  &\multicolumn{1}{c}{FAIL} \\

\multicolumn{1}{l}{}&\multicolumn{1}{l}{C1}&\multicolumn{1}{l}{\hspace{1.168cm}mathematical mathematic equational \textbf{algebraic} combinatorics}  &\multicolumn{1}{c}{} \\

\multicolumn{1}{l}{}&\multicolumn{1}{l}{C2 (C1)}&\multicolumn{1}{l}{\hspace{1.168cm}mathematical geometry \textbf{algebraic} equational mathematic}  &\multicolumn{1}{c}{} \\

\cmidrule(lr){1-4}
\multicolumn{1}{l}{biseri (HR) $\to$ \textbf{pearls} (EN)}&\multicolumn{1}{l}{Baseline}&\multicolumn{1}{l}{\hspace{1.168cm}\textbf{pearls} sapphires rubies carnations jades}  &\multicolumn{1}{c}{FAIL} \\

\multicolumn{1}{l}{}&\multicolumn{1}{l}{C1}&\multicolumn{1}{l}{\hspace{1.168cm}gems \textbf{pearls} sapphires gem treasures}  &\multicolumn{1}{c}{} \\

\multicolumn{1}{l}{}&\multicolumn{1}{l}{C2 (C1)}&\multicolumn{1}{l}{\hspace{1.168cm}gems jewels \textbf{pearls} diamonds arks}  &\multicolumn{1}{c}{} \\

\cmidrule(lr){1-4}
\multicolumn{1}{l}{tiho (HR) $\to$ \textbf{quietly} (EN)}&\multicolumn{1}{l}{Baseline}&\multicolumn{1}{l}{\hspace{1.168cm}\textbf{quietly} quiet sobbing joyously crying}  &\multicolumn{1}{c}{FAIL} \\

\multicolumn{1}{l}{}&\multicolumn{1}{l}{C1}&\multicolumn{1}{l}{\hspace{1.168cm}hums sunshine crying tink tablo}  &\multicolumn{1}{c}{} \\

\multicolumn{1}{l}{}&\multicolumn{1}{l}{C2 (C1)}&\multicolumn{1}{l}{\hspace{1.168cm}quiet crying loud hums tink}  &\multicolumn{1}{c}{} \\

\cmidrule(lr){1-4}
\multicolumn{1}{l}{kanu (HR) $\to$ \textbf{canoe} (EN)}&\multicolumn{1}{l}{Baseline}&\multicolumn{1}{l}{\hspace{1.168cm}\textbf{canoe} canoes archery kabaddi outrigger}  &\multicolumn{1}{c}{FAIL} \\

\multicolumn{1}{l}{}&\multicolumn{1}{l}{C1}&\multicolumn{1}{l}{\hspace{1.168cm}sport \textbf{canoe} taekwondo archery sports}  &\multicolumn{1}{c}{} \\

\multicolumn{1}{l}{}&\multicolumn{1}{l}{C2 (C1)}&\multicolumn{1}{l}{\hspace{1.168cm}sport \textbf{canoe} budo sports sambo}  &\multicolumn{1}{c}{} \\

\cmidrule(lr){1-4}
\multicolumn{1}{l}{oluje (HR) $\to$ \textbf{storms} (EN)}&\multicolumn{1}{l}{Baseline}&\multicolumn{1}{l}{\hspace{1.168cm}\textbf{storms} thunderstorm storm windstorms thunderstorms}  &\multicolumn{1}{c}{FAIL} \\

\multicolumn{1}{l}{}&\multicolumn{1}{l}{C1}&\multicolumn{1}{l}{\hspace{1.168cm}storm \textbf{storms} winds blizzards tsunami}  &\multicolumn{1}{c}{} \\

\multicolumn{1}{l}{}&\multicolumn{1}{l}{C2 (C1)}&\multicolumn{1}{l}{\hspace{1.168cm}winds \textbf{storms} storm fires rain}  &\multicolumn{1}{c}{} \\

\cmidrule(lr){1-4}
\multicolumn{1}{l}{bombardiere (IT) $\to$ \textbf{bomber} (EN)}&\multicolumn{1}{l}{Baseline}&\multicolumn{1}{l}{\hspace{1.168cm}aircraft \textbf{bomber} floatplane biplane pilotless}  &\multicolumn{1}{c}{SUCCESS} \\

\multicolumn{1}{l}{}&\multicolumn{1}{l}{C1}&\multicolumn{1}{l}{\hspace{1.168cm}\textbf{bomber} aircraft floatplane biplane superfortress}  &\multicolumn{1}{c}{} \\

\multicolumn{1}{l}{}&\multicolumn{1}{l}{C2 (C1)}&\multicolumn{1}{l}{\hspace{1.168cm}\textbf{bomber} aircraft airliner biplane arado}  &\multicolumn{1}{c}{} \\

\cmidrule(lr){1-4}
\multicolumn{1}{l}{spinaci (IT) $\to$ \textbf{spinach} (EN)}&\multicolumn{1}{l}{Baseline}&\multicolumn{1}{l}{\hspace{1.168cm}carrots \textbf{spinach} onions vegetables garlic}  &\multicolumn{1}{c}{SUCCESS} \\

\multicolumn{1}{l}{}&\multicolumn{1}{l}{C1}&\multicolumn{1}{l}{\hspace{1.168cm}\textbf{spinach} carrots onions tomato beans}  &\multicolumn{1}{c}{} \\

\multicolumn{1}{l}{}&\multicolumn{1}{l}{C2 (C1)}&\multicolumn{1}{l}{\hspace{1.168cm}\textbf{spinach} carrots beans chilies tomato}  &\multicolumn{1}{c}{} \\

\cmidrule(lr){1-4}
\multicolumn{1}{l}{passero (IT) $\to$ \textbf{sparrow} (EN)}&\multicolumn{1}{l}{Baseline}&\multicolumn{1}{l}{\hspace{1.168cm}chaffinch sparrowhawk strepera whimbrel chiffchaff}  &\multicolumn{1}{c}{SUCCESS} \\

\multicolumn{1}{l}{}&\multicolumn{1}{l}{C1}&\multicolumn{1}{l}{\hspace{1.168cm}bird \textbf{sparrow} partridge dove sparrowhawk}  &\multicolumn{1}{c}{} \\

\multicolumn{1}{l}{}&\multicolumn{1}{l}{C2 (C1)}&\multicolumn{1}{l}{\hspace{1.168cm}\textbf{sparrow} bird dove pigeon crow}  &\multicolumn{1}{c}{} \\

\cmidrule(lr){1-4}
\multicolumn{1}{l}{aspettativa (IT) $\to$ \textbf{expectation} or \textbf{expectancy} (EN)}&\multicolumn{1}{l}{Baseline}&\multicolumn{1}{l}{\hspace{1.168cm}expectancies \textbf{expectancy} \textbf{expectation} expectations maturity}  &\multicolumn{1}{c}{SUCCESS} \\

\multicolumn{1}{l}{}&\multicolumn{1}{l}{C1}&\multicolumn{1}{l}{\hspace{1.168cm}\textbf{expectancy} \textbf{expectation} expectations expectancies chance}  &\multicolumn{1}{c}{} \\

\multicolumn{1}{l}{}&\multicolumn{1}{l}{C2 (C1)}&\multicolumn{1}{l}{\hspace{1.168cm}\textbf{expectation} \textbf{expectancy} chance expectations experience}  &\multicolumn{1}{c}{} \\

\cmidrule(lr){1-4}
\multicolumn{1}{l}{cereale (IT) $\to$ \textbf{cereal} (EN)}&\multicolumn{1}{l}{Baseline}&\multicolumn{1}{l}{\hspace{1.168cm}sorghum cereals barley wheat corn}  &\multicolumn{1}{c}{SUCCESS} \\

\multicolumn{1}{l}{}&\multicolumn{1}{l}{C1}&\multicolumn{1}{l}{\hspace{1.168cm}barley wheat \textbf{cereal} sorghum corn}  &\multicolumn{1}{c}{} \\

\multicolumn{1}{l}{}&\multicolumn{1}{l}{C2 (C1)}&\multicolumn{1}{l}{\hspace{1.168cm}\textbf{cereal} wheat grain barley sorghum}  &\multicolumn{1}{c}{} \\

\cmidrule(lr){1-4}
\multicolumn{1}{l}{cifre (IT) $\to$ \textbf{digits} (EN)}&\multicolumn{1}{l}{Baseline}&\multicolumn{1}{l}{\hspace{1.168cm}\textbf{digits} digit numbers decimals numeric}  &\multicolumn{1}{c}{FAIL} \\

\multicolumn{1}{l}{}&\multicolumn{1}{l}{C1}&\multicolumn{1}{l}{\hspace{1.168cm}numbers \textbf{digits} digit decimals numeric}  &\multicolumn{1}{c}{} \\

\multicolumn{1}{l}{}&\multicolumn{1}{l}{C2 (C1)}&\multicolumn{1}{l}{\hspace{1.168cm}numbers \textbf{digits} digit number decimals}  &\multicolumn{1}{c}{} \\

\cmidrule(lr){1-4}
\multicolumn{1}{l}{obbligatorio (IT) $\to$ \textbf{compulsory} (EN)}&\multicolumn{1}{l}{Baseline}&\multicolumn{1}{l}{\hspace{1.168cm}\textbf{compulsory} mandatory obligatory requirement mandating}  &\multicolumn{1}{c}{FAIL} \\

\multicolumn{1}{l}{}&\multicolumn{1}{l}{C1}&\multicolumn{1}{l}{\hspace{1.168cm}mandatory \textbf{compulsory} obligatory required permitted}  &\multicolumn{1}{c}{} \\

\multicolumn{1}{l}{}&\multicolumn{1}{l}{C2 (C1)}&\multicolumn{1}{l}{\hspace{1.168cm}mandatory obligatory \textbf{compulsory} permitted required}  &\multicolumn{1}{c}{} \\

\cmidrule(lr){1-4}
\multicolumn{1}{l}{violoncello (IT) $\to$ \textbf{cello} (EN)}&\multicolumn{1}{l}{Baseline}&\multicolumn{1}{l}{\hspace{1.168cm}\textbf{cello} violin clarinet piano violoncello}  &\multicolumn{1}{c}{FAIL} \\

\multicolumn{1}{l}{}&\multicolumn{1}{l}{C1}&\multicolumn{1}{l}{\hspace{1.168cm}violin \textbf{cello} piano violoncello clarinet}  &\multicolumn{1}{c}{} \\

\multicolumn{1}{l}{}&\multicolumn{1}{l}{C2 (C1)}&\multicolumn{1}{l}{\hspace{1.168cm}violin \textbf{cello} violoncello piano clarinet}  &\multicolumn{1}{c}{} \\

\cmidrule(lr){1-4}
\multicolumn{1}{l}{pavone (IT) $\to$ \textbf{peacock} (EN)}&\multicolumn{1}{l}{Baseline}&\multicolumn{1}{l}{\hspace{1.168cm}\textbf{peacock} partridge dove doves pheasant}  &\multicolumn{1}{c}{FAIL} \\

\multicolumn{1}{l}{}&\multicolumn{1}{l}{C1}&\multicolumn{1}{l}{\hspace{1.168cm}dove red \textbf{peacock} blue garland}  &\multicolumn{1}{c}{} \\

\multicolumn{1}{l}{}&\multicolumn{1}{l}{C2 (C1)}&\multicolumn{1}{l}{\hspace{1.168cm}garland dove \textbf{peacock} bull red}  &\multicolumn{1}{c}{} \\

\cmidrule(lr){1-4}
\multicolumn{1}{l}{sanzione (IT) $\to$ \textbf{sanction} (EN)}&\multicolumn{1}{l}{Baseline}&\multicolumn{1}{l}{\hspace{1.168cm}\textbf{sanction} infraction offence sanctionable discretionary}  &\multicolumn{1}{c}{FAIL} \\

\multicolumn{1}{l}{}&\multicolumn{1}{l}{C1}&\multicolumn{1}{l}{\hspace{1.168cm}infraction offence \textbf{sanction} discretionary penalty}  &\multicolumn{1}{c}{} \\

\multicolumn{1}{l}{}&\multicolumn{1}{l}{C2 (C1)}&\multicolumn{1}{l}{\hspace{1.168cm}infraction \textbf{sanction} offence penalty probation}  &\multicolumn{1}{c}{} \\

\bottomrule

\end{tabular}
}

\caption{Translation examples on HR$\to$EN and IT$\to$EN. We include here ground truth translation pairs and show top five predictions (in the "Top Five Predictions" column above, left $\to$ right: number one item in the ranked list $\to $ number five item in the ranked list) via NN retrieval for each of the three methods, that is, C1 w/o CL (Baseline), C1 and C2 (C1).}
\label{table:translationexamples}
\end{center}
\end{table*}